\pdfoutput=1

\documentclass[11pt]{article}

\usepackage[]{ACL2023}

\usepackage{inconsolata}

\usepackage{times}
\usepackage{latexsym}

\usepackage[T1]{fontenc}

\usepackage[utf8]{inputenc}

\usepackage{microtype}
\usepackage{multirow}
\usepackage{graphicx}
\usepackage{algorithm}
\usepackage{algorithmic}
\usepackage[mathscr]{euscript}
\usepackage{subcaption}
\usepackage{times}
\usepackage{latexsym}
\usepackage{tabularx}
\usepackage{booktabs}
\usepackage{array}
\usepackage{footnote}
\usepackage{graphicx}
\usepackage{dingbat}
\usepackage{svg}
\usepackage{amsmath}
\usepackage{amsfonts}
\usepackage{mathtools}
\usepackage[normalem]{ulem}
\usepackage{bchart}
\usepackage{pgfplots}
\usepackage{hyperref}
\usepackage{array}
\usepackage{amssymb}
\usepackage{pifont}
\usepackage{adjustbox}
\usepackage{xr}
\usepackage{collcell}
\usepackage{colortbl}
\usepackage{todonotes}
\usepackage{appendix}
\usepackage{lipsum}                     
\usepackage{xargs}                      
\usepackage{mathrsfs}
\usepackage{nccmath}
\usepackage[inline]{enumitem}
\usepackage{xcolor}
\pgfplotsset{compat=1.18}

\definecolor{bblue}{HTML}{4F81BD}
\definecolor{rred}{HTML}{C0504D}
\definecolor{ggreen}{HTML}{9BBB59}
\definecolor{ppurple}{HTML}{9F4C7C}

\newcommandx{\unsure}[2][1=]{\todo[linecolor=red,backgroundcolor=red!25,bordercolor=red,#1]{#2}}
\newcommandx{\change}[2][1=]{\todo[linecolor=blue,backgroundcolor=blue!25,bordercolor=blue,#1]{#2}}
\newcommandx{\info}[2][1=]{\todo[linecolor=OliveGreen,backgroundcolor=OliveGreen!25,bordercolor=OliveGreen,#1]{#2}}
\newcommandx{\improvement}[2][1=]{\todo[linecolor=Plum,backgroundcolor=Plum!25,bordercolor=Plum,#1]{#2}}
\newcommandx{\thiswillnotshow}[2][1=]{\todo[disable,#1]{#2}}

\newcommand{\setset}{\mathcal{D}}

\newcommand{\mbert}{M-BERT}
\newcommand{\xlmrbase}{XLM-R}
\newcommand{\lng}{\boldsymbol\ell}
\newcommand{\task}{K}
\newcommand{\datalangs}[1]{\setset_{1 \cdots {#1}}}

\newcommand{\datalang}[1]{\setset_{#1}}

\newcommand{\alllang}{\mathscr{L}}
\newcommand{\permute}{\mathfrak{S}}
\newcommand{\langallsetelongated}{\{\lng_1, \lng_2 \cdots \lng_N\}}

\newcommand{\naiveft}{\emph{Naive Seq FT}}
\newcommand{\langspec}{\emph{Lang-Spec FT}}
\newcommand{\jointinc}{\emph{Inc Joint}}
\newcommand{\multi}{\emph{Multilingual}}

\newcommand{\spectrans}{\emph{Lang-Spec Trans}}
\newcommand{\spechead}{\emph{Lang-Spec Task}}
\newcommand{\ada}{\emph{Lang-Spec Ada}}
\newcommand{\adatuned}{\emph{Lang-Spec Ada(T)}}
\newcommand{\adafrozen}{\emph{Lang-Spec Ada(F)}}

\newcommand{\htol}{\emph{H2L}}
\newcommand{\ltoh}{\emph{L2H}}

\newcommand{\specemb}{\emph{Lang-Spec Embed}}
\newcommand{\specenca}{\emph{Lang-Spec Enc[1-9]}}
\newcommand{\specencb}{\emph{Lang-Spec Enc[10-12]}}
\newcommand{\specencaa}{\emph{Lang-Spec Enc[1-3]}}
\newcommand{\specencbb}{\emph{Lang-Spec Enc[4-6]}}
\newcommand{\specenccc}{\emph{Lang-Spec Enc[7-9]}}
\newcommand{\specencaaa}{\emph{Lang-Spec Enc[1-6]}}
\newcommand{\specencbbb}{\emph{Lang-Spec Enc[7-12]}}
\newcommand{\specencall}{\emph{Lang-Spec Enc[1-12]}}

\newcommand{\langspecadatuned}{\langspec{} + Ada(T)}
\newcommand{\langspecadafrozen}{\langspec{} + Ada(F)}

\newcommand{\ewc}{\emph{EWC}}
\newcommand{\ewconline}{\emph{EWC-Online}}

\newcommand{\er}{\emph{ER}}
\newcommand{\kdlogit}{\emph{KD-Logit}}
\newcommand{\kdrep}{\emph{KD-Rep}}

\makeatletter
\newcommand*{\inlineequation}[2][]{%
  \begingroup
    \refstepcounter{equation}%
    \ifx\\#1\\%
    \else
      \label{#1}%
    \fi
    \relpenalty=10000 %
    \binoppenalty=10000 %
    \ensuremath{%
      #2%
    }%
    ~\@eqnnum
  \endgroup
}
\makeatother

\title{Cross-lingual Continual Learning}

\author{Meryem M'hamdi \qquad Xiang Ren \qquad Jonathan May \\
Information Sciences Institute \\ 
University of Southern California \\ 
\texttt{\{meryem, xiangren, jonmay\}@isi.edu}}

\begin{document}

\maketitle
\begin{abstract}
The longstanding goal of multi-lingual learning has been to develop a universal cross-lingual model that can withstand the changes in multi-lingual data distributions. There has been a large amount of work to adapt such multi-lingual models to unseen target languages. However, the majority of work in this direction focuses on the standard one-hop transfer learning pipeline from source to target languages, whereas in realistic scenarios, new languages can be incorporated at any time in a sequential manner. In this paper, we present a principled \textbf{C}ross-lingual \textbf{C}ontinual \textbf{L}earning (CCL) evaluation paradigm, where we analyze different categories of approaches used to continually adapt to emerging data from different languages. We provide insights into what makes multilingual sequential learning particularly challenging. To surmount such challenges, we benchmark a representative set of cross-lingual continual learning algorithms and analyze their knowledge preservation, accumulation, and generalization capabilities compared to baselines on carefully curated datastreams. The implications of this analysis include a recipe for how to measure and balance different cross-lingual continual learning desiderata, which go beyond conventional transfer learning. 
\end{abstract}

\section{Introduction}
\label{sec:intro}

\begin{figure}[t]
 \centering
\includegraphics[width=0.48\textwidth]{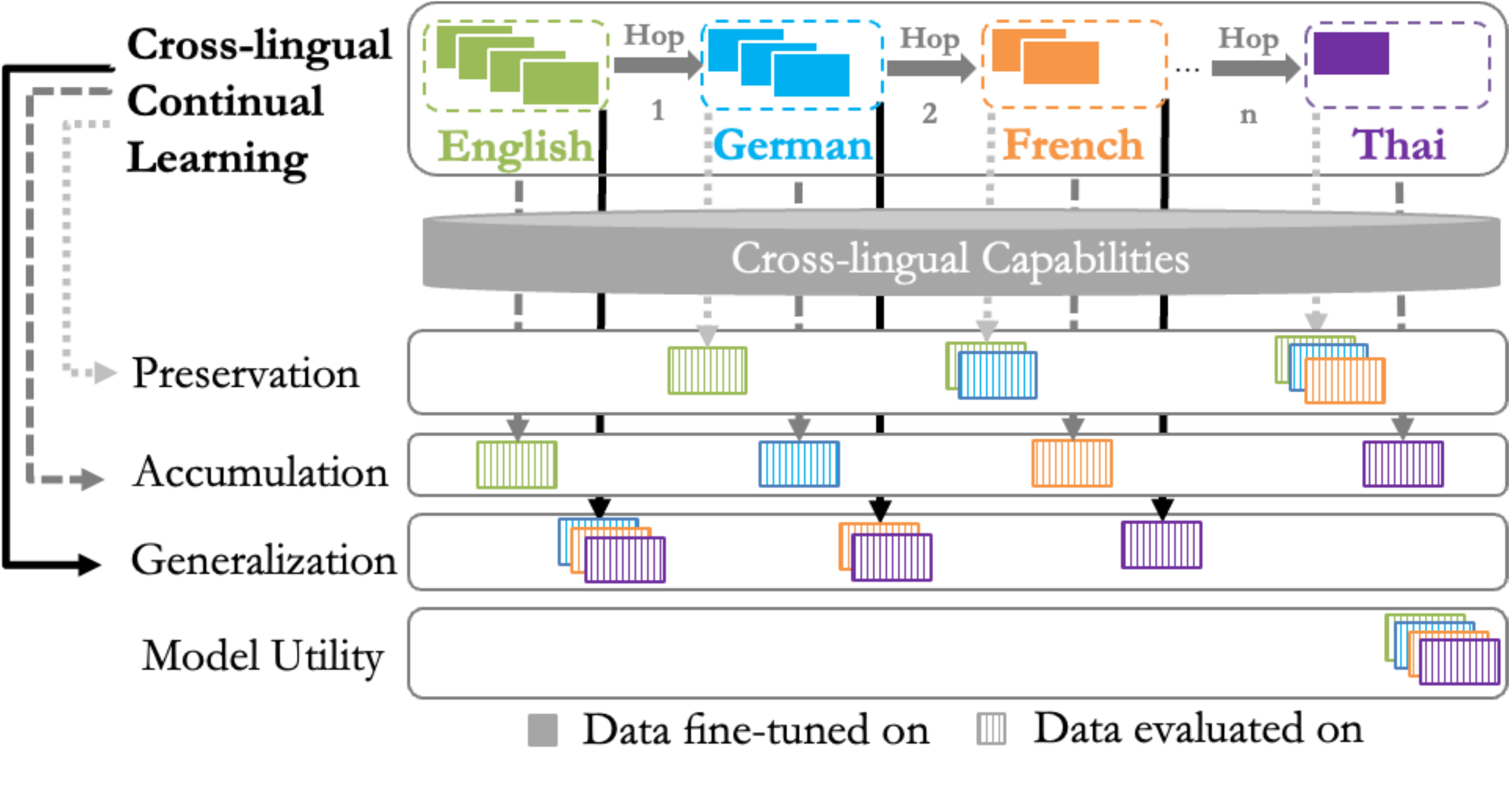}
 \caption{An overview of CCL: We use an example of a non-stationary datastream moving from high to low resource languages. Each bold and dashed box represents either a training or test data instance being fine-tuned or evaluated on, respectively. To support this problem setup, we evaluate the cross-lingual capabilities of \emph{continual approaches}. Those capabilities include knowledge \textbf{preservation} on old languages, \textbf{accumulation} to the current language, and \textbf{generalization} to unseen languages at each point of the training. In addition to that, we evaluate \textbf{model utility} at the end of continual learning.} 
 \label{fig:x-cont-learn-framework}
 \vspace{-0.4cm}
\end{figure}

With more than 7,000 languages spoken around the globe, downstream applications still lack proper linguistic resources across languages~\cite{statenlp-joshi-acl20}, necessitating the use of \textit{transfer learning} techniques that take advantage of data that is mismatched to the application. 
In an effort to simplify architecture complexity and energy consumption, it is desirable to unify multi-lingual performance into a single, parameter- and memory-constrained model, and to allow this model to evolve, learning on multi-lingual training data as it becomes available without having to pre-train or fine-tune from scratch. Such is the longstanding goal of language representation learning. Existing multi-lingual representations such as \mbert{}~\cite{bert-devlin-naacl19} and \xlmrbase{}~\cite{xlm-r-conneau} are strong pillars in cross-lingual transfer learning, but if care is not taken when choosing how to fine-tune them, they can neglect to maximize \textit{transfer}~\cite{ruder-etal-2019-transfer} to new tasks or languages and are subject to \textit{forgetting}~\cite{MCCLOSKEY1989109}, where performance decreases after exposure to new task or language. 
\vspace{-0.06cm}

Most previous work that attempts to deal with the challenge of transfer exploitation and forgetting mitigation focuses on the problem of sequentially learning over different NLP downstream tasks or domains~\cite{sun-lamol-iclr20,relextr-han-acl20,madotto-contTOD-emnlp21}, rather than on language shifts. Indeed, the current literature for learning over sequences of languages is rather scarce and is mostly reduced to cross-lingual transfer learning between a pair of languages~\cite{xcontlearn_ner_pos_gem-liu-repl4nlp21,cont_multinmt-garcia-naacl21,muller-etal-2021-unseen,pfeiffer-etal-2021-unks,minixhofer-etal-2022-wechsel}. \citeauthor{xcontlearn_ner_pos_gem-liu-repl4nlp21} pre-train a (parent) language model and then fine-tune it on a downstream task in one of several different (child) languages. This conflates task and language transfer and confuses analysis -- the interference between the pre-trained language model `task' and the fine-tuned task, along with the parent and child languages, cannot be disentangled. \citeauthor{cont_multinmt-garcia-naacl21} propose an adaptation scheme to each new language pair independently while retaining the translation quality on the parent language pairs. Similarly, \citet{muller-etal-2021-unseen} and~\citet{pfeiffer-etal-2021-unks} propose lexical and semantic level techniques to adapt to target languages. However, all these mentioned works still focus on the `one-hop' case, consisting of two steps: (1) training on initial parent language(s) (pairs), then (2) adapting to new children language(s) (pairs); the effect of multiple shifts in the datastream is not trivially generalizable to more than one hop. More recently,~\citet{pfeiffer-etal-2022-lifting} propose an approach for language-specific modules based on adapters and evaluate that on sequential streams of languages. However, they only focus on adapters and two desiderata of continual learning: interference mitigation and transfer maximization. We need a more robust and comprehensive fine-grained evaluation that balances the dynamics between different cross-lingual continual learning desiderata. 
\vspace{-0.05cm}

In this paper, we pave the way for a more comprehensive multi-hop continual learning evaluation that simulates the sequential learning of a single task over a stream of input from different languages. This evaluation paradigm requires experimentation over \textit{balanced streams} of $n$ data scenarios for $n > 2$. Unlike previous work, this paper concretely defines the following comprehensive goals along with their evaluation metrics as guidelines for analyzing the cross-lingual capabilities of multilingual sequential training: knowledge preservation, accumulation, generalization, and model utility as shown in Figure~\ref{fig:x-cont-learn-framework}. We apply our test bed to a six-language task-oriented dialogue benchmark and comprehensively analyze a wide variety of successful continual learning algorithms. These algorithms are derived from previous literature investigated in continual learning contexts different from the cross-lingual context, including (a) model-expansion~\cite{madx-pfeiffer-emnlp20}, (b) regularization~\cite{ewc-kirkpatrick-nas17}, (c) memory replay~\cite{er-chaudhry-arxiv19}, and (d) distillation-based approaches~\cite{hinton-kdlogit-arxiv15,aguilar-kdrep-aaai20}. Our findings confirm the need for a multi-hop analysis and the effectiveness of continual learning algorithms in enhancing knowledge preservation and accumulation of our multilingual language model. We additionally demonstrate the robustness of different continual learning approaches to variations in individual data setup choices that would be misleading if presented in a traditional manner.


Our \textbf{main contributions} are:
\begin{enumerate*}[label=(\arabic*)]
\item {We are the first to explore and analyze cross-lingual continual fine-tuning\footnote{Our code is available at \url{https://github.com/meryemmhamdi1/x-continuous-learning}.} across multiple hops and show the importance of this multi-hop analysis in reaching clearer conclusions with greater confidence compared to conventional cross-lingual transfer learning ~(\S\ref{sec:q6-multi-step-cross-cont-learn}).}
\item {We demonstrate the aggregated effectiveness of a range of different continual learning approaches (Figure~\ref{fig:x-cont-learn-framework}) at reducing forgetting and improving transfer~(\S\ref{sec:q3-effectiveness-cross-cont-learn}) compared to multilingual sequential baselines~(\S\ref{sec:q1-catastrophic-forgetting-generalization}).}\item {We make concrete recommendations on model design to balance transfer and final model performance with forgetting~(\S\ref{sec:q3-effectiveness-cross-cont-learn}).}
\item{We show that the order of languages and data set size impacts the knowledge preservation and accumulation of multi-lingual sequential fine-tuning and identify the continual learning approaches that are most robust to this variation~(\S\ref{sec:q2-analysis-across-lang-perm}).}
\item {We analyze zero-shot generalization trends and their correlation with forgetting and show that current continual learning approaches do not substantially improve the generalization~(\S\ref{sec:q5-zero-shot-generalization}).}
\end{enumerate*}

\section{Cross-lingual Continual Learning}
\label{sec:x-cont-fine-tuning}

In this section, we formally define cross-lingual continual learning, describe its goals and challenges, and introduce the downstream tasks, datastreams, and evaluation protocols used. Although we are not the first to define or investigate continual learning for languages, we are, to the best of our knowledge, the first to define and study cross-lingual continual learning where continual learning is focused on languages only. Thus, we formally define cross-lingual continual learning as learning over a set of languages seen sequentially in multiple hops, which is truer to the term of cross-lingual and continual learning, respectively. We distinguish that from `cross-lingual cross-task cross-stage continual learning', which continually learns over a set of pretraining and downstream tasks sampled from different languages~\cite{ xcontlearn_ner_pos_gem-liu-repl4nlp21} and `cross-lingual one-hop transfer learning'~\cite{cont_multinmt-garcia-naacl21}.
\vspace{-0.3cm}
\subsection{Problem Formulation}
\label{sec:problem-formulation}
We define cross-lingual continual learning as the problem of sequentially fine-tuning a model $\theta$ for a particular downstream task \task{} over a cross-lingual datastream. In this case, a cross-lingual data \textit{stream} is made of $N$ labeled and distinct datasets $\datalangs{N}$, each one sampled from a distinct language and consisting of separate train and test portions. Let \textit{$hop_i$} be the stage in cross-lingual continual learning where $\theta_i$ is optimized to $\theta_{i+1}$ via exposure to $\datalang{i}$. Let $\alllang = \langallsetelongated$ be a set of labeled \textit{languages}, let $\permute(\alllang)$ be the set of all \textit{permutations} of $\alllang$, and without loss of generality let $p \in \permute(\alllang)$ be one such permutation and $p[i] \in \alllang$ be the $i$th language in $p$. The language of $\datalang{i}$ is $p[i]$. Therefore, by default, the number of languages used is equal to the number of datasets. Let $\datalang{<i}$ and $\datalang{>i}$ refer to a sequence of datasets (train or test portions, depending on context) used in hops from 1 to $i-2$ and $i$ to $N-1$, respectively; we generalize these terms to $\datalang{{\leq}i}$ and $\datalang{{\geq}i}$ by including hop $i-1$ as well at the end or, respectively, beginning of the sequence. 
\subsection{Goals}
\label{sec:goals}
We define the goals,\footnote{To the best of our knowledge, those goals were never synthesized for the context of cross-lingual continual learning.} necessarily dependent on each other, for our study of cross-lingual continual learning as follows (also depicted in Figure~\ref{fig:x-cont-learn-framework}):

\begin{itemize}[leftmargin=*]
    \itemsep0em 
    \vspace{-0.3cm}
    \item{\emph{Cross-lingual preservation.} This is the ability to retain previous knowledge of seen languages.}
    \item{\emph{Cross-lingual accumulation.} This is the ability to accumulate knowledge learned from previous languages to benefit from learning on the current language.}
    \item{\emph{Cross-lingual generalization.} This is the ability to generalize uniformly well to unseen languages, which goes beyond accumulating knowledge up to the current languages.}
    \item{\emph{Model utility.} This is the ability of the fully trained model to perform equally well in all languages.}
    \vspace{-0.3cm}
\end{itemize}

In this paper, we wish to understand the relationships between these goals. Our aim is to come up with a recipe for a more systematic cross-lingual continual learning. Thus, we need to understand if the goals are aligned with each other or if maximizing some goals leads to minimizing other goals.
\vspace{-0.6cm}
\subsection{Challenges}
\label{sec:challenges}
Learning sequentially from a non-stationary data distribution (i.e., task datasets coming from different languages) can impose considerable challenges on the goals defined earlier: 

\begin{itemize}[leftmargin=*]
    \itemsep0em 
    \vspace{-0.2cm}
    \item{\emph{Catastrophic forgetting.} This happens when fine-tuning a model on $\datalang{\geq i}$ leads to a decrease in the performance on $\datalang{<i}$.}
    \item{\emph{Negative transfer.} This happens when fine-tuning a model up to $\datalang{\leq i}$ leads to a lower performance on $\datalang{i}$ than training on it alone.}
    \item{\emph{Low zero-shot transfer.} This happens when fine-tuning on $\datalang{\leq i}$ gives a lower performance than random on unseen $\datalang{>i}$.}
    \item{\emph{Low final performance.} This is when fine-tuning on all $\datalang{\leq N}$ gives an uneven performance between languages when tested on $\datalang{\leq N}$ at the end of training.}
    \vspace{-0.1cm}
\end{itemize}
    
\vspace{-0.5cm}
\subsection{Downstream Tasks and Datastreams}
\label{sec:downstream-task-datastreams}

Here, we describe the downstream tasks and multi-lingual sequential datastreams used.  
\vspace{-0.2cm}
\paragraph{Downstream Tasks.}
\label{sec:dataset}
We choose task-oriented dialogue parsing as a use case and consider the multi-lingual task-oriented parsing (MTOP) benchmark~\cite{mtop-li-eacl21}. Task-oriented dialogue parsing provides a rich testbed for analysis, as it encompasses two subtasks: \textit{intent classification} and \textit{slot filling}, thus allowing us to test different task capabilities in cross-lingual continual learning.

\vspace{-0.2cm}
\paragraph{Datastream Construction.}
\label{sec:permutations}
For a set of $N$ languages $\alllang$, our study considers a permutation subset $P \subset \permute(\alllang)$ with the following properties:\footnote{Details of the different language permutations used for the datastreams can be found in Appendix~\ref{app:datastream}.}
\vspace{-0.7cm}
\begin{itemize}[leftmargin=*]
    \itemsep0em 
    \item $|P| = |\alllang| = N$, i.e. $P$ consists of $N$ permutations, each of which is a sequence of $N$ datasets in each of the $N$ languages in $\alllang$.
    \item $\forall \lng \in \alllang$, $\forall j \in 1 \ldots N$, there exists some $p \in P$ such that $p[j]=\lng$.
    \item \htol{} $\in P$, the permutation from most high-resource to most low-resource fine-tuning data sets, based on the training split dataset size.
    \item \ltoh{} $\in P$, the reverse of \htol{}.
\end{itemize}
\vspace{-0.2cm}
In our experiments, we use MTOP~\cite{mtop-li-eacl21}, which is a multi-lingual task-oriented dialogue dataset that covers six typologically diverse languages and spans over 11 domains and 117 intents. We chose MTOP since it is the largest scale dataset available for task-oriented dialogue, and because it covers languages that have varying amounts of data resources available. We use only the flat representation of slots (without nesting) to simplify our evaluation. We use the original data for most experiments. Table~\ref{tab:data-mtop} shows a summary of the number of sentences (dialogue utterances) per language and split.

\begin{table}[ht]
\small
\centering
    \begin{tabular}{l|l|l|l|l}
    \toprule
    \textbf{Lang} & \textbf{ISO} & \textbf{Train} & \textbf{Dev} & \textbf{Test} \\ \toprule
    English       & EN &   15,667 & 2,235 & 4,386 \\ 
    German       & DE &    13,424 & 1,815 & 3,549  \\
    French         & FR &  11,814 & 1,577 & 3,193  \\
    Hindi         &  HI &  11,330 & 2,012 & 2,789  \\
    Spanish         & ES &  10,934 & 1,527 & 2,998 \\    
    Thai        & TH &  10,759 & 1,671 & 2,765 \\ \bottomrule
    \end{tabular}
\caption{Number of sentences in MTOP per language and split.}
\label{tab:data-mtop}

\vspace{-0.5cm}
\end{table}
\vspace{-0.2cm}
\subsection{Evaluation Protocols}
\label{sec:metrics}
For each language permutation, we train on each dataset in sequence but continually evaluate on all languages. Let $R$ be some success metric for evaluating a downstream task $\task{}$ and $R_{i,\leq{j}}$ be the evaluation on the test set for language $\lng{}_i$ fine-tuning $\task{}$ on $\datalang{\leq j}$. We define the following \emph{meta-metrics} (which are inspired, but slightly different from the metrics in~\citet{gem-lopezpaz-nips17} and~\citet{agem-chaudhry-iclr19}):
\vspace{-0.2cm}
\begin{itemize}[leftmargin=*] 
 \itemsep0em 
    \item{\textbf{Forgetting (F $\downarrow$).} 
    This is the average forgetting \emph{over all datasets} (excluding the first dataset) computed as:
    \vspace{-0.4cm}
    \begin{ceqn}
        \begin{align}
        \label{eqn:forgetting}
            \begin{split}
                F=\frac{1}{N-1} \sum_{j=2}^{N}{F_{\leq{j}}},
                \\
                F_{\leq{j}} =\frac{1}{j-1}\sum_{i=1}^{j-1}{F_{i,\leq{j}}},
            \end{split}
        \end{align}
    \end{ceqn}
     where $F_{\leq{j}}$ is the average forgetting that occurred at the point of training $\datalang{j}$. We compute $F_{i, \leq{j}} = \max_{k \in \left[1,j-1\right]}R_{i,\leq{k}}-R_{i,\leq{j}}\label{eqn:forgetting_ij}$. $F_{i,\leq{j}}$ is the degree to which performance on $\datalang{i}$ has suffered by continuing to train on $\datalang{\leq j}$ instead of stopping before covering $\datalang{j}$.}
    \vspace{-0.1cm}
    \item{\textbf{Transfer (T $\uparrow$).}
    This is the average forward transfer computed as: 
    \vspace{-0.3cm}
    \begin{ceqn}
        \begin{align}
        \label{eqn:fwt}
            \begin{split}
             T = \frac{1}{N-1} \sum_{i=2}^N {T_i},
             \\
             T_i = R_{i,\leq{i}} - R_i,
            \end{split}
        \end{align}
    \end{ceqn}
    
    where $R_i$ denotes evaluation of a model fine-tuned \textit{only} on $\datalang{i}$. Then, $T_i$ is the incremental impact of sequential training on datasets prior to seeing $\datalang{i}$. To measure \emph{generalization to new languages}, we add a \textbf{zero-shot transfer ($\textbf{T}^\textbf{0}$ $\uparrow$)} metric\label{sec:zero-shot-transfer-eq} measured as:
    \vspace{-0.3cm}
    \begin{ceqn}
        \begin{align}
        \label{eqn:fwt0}
            \begin{split}
             T^0 = \frac{1}{N-1} \sum_{i=2}^N {T^0_i},
             \\
             T^0_i=\frac{1}{i-1}\sum_{j=1}^{i-1} R_{i,\leq{j}} - R^0_i,
            \end{split}
        \end{align}
    \end{ceqn}
    where $T^0_i$ is the average performance of a model on the forward transfer to a language $\lng{}_i$ after training on $\datalang{<i}$ compared to the random performance $R^{0}_{i}$ before even fine-tuning on any language (i.e. using fixed pre-trained \mbert{} weights and randomly initialized weights for the output layer).}
    \vspace{-0.1cm}
     \item{\textbf{Final performance (FP $\uparrow$).}  \label{sec:final-perf-eq} This is the average performance after training on all datasets in the studied stream, computed as:
     \vspace{-0.2cm}
     \begin{equation}
         \label{eqn:final-perf}
         FP = \frac{1}{N} \sum_{i=1}^N {R_{i,\leq{N}}}.
     \end{equation}
    }
\end{itemize}
\vspace{-0.5cm}

\section{Methods}
\label{sec:method}

\begin{figure*}[t]
\vspace{-0.4cm}
 \centering
\includegraphics[width=1\textwidth]{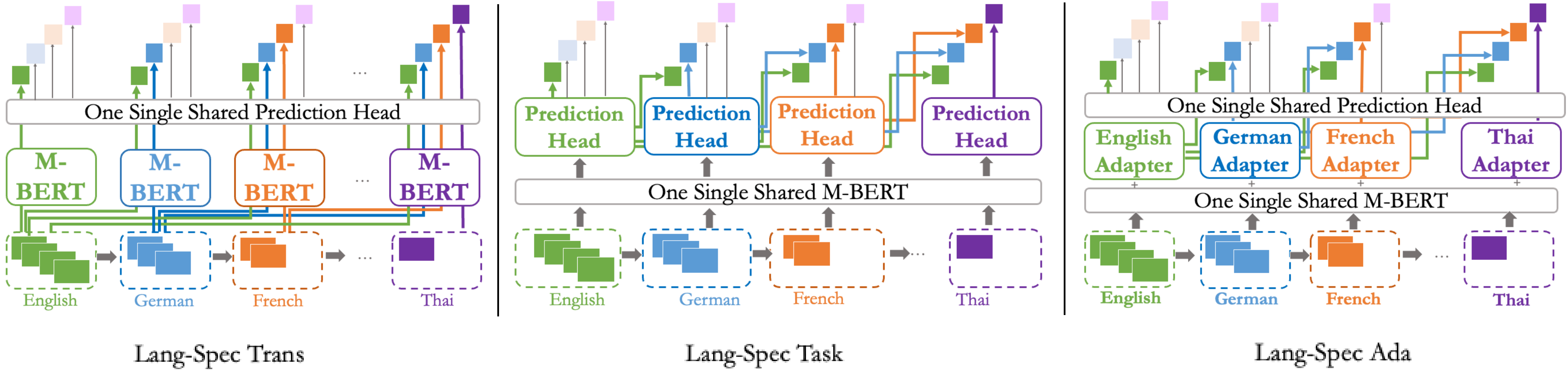}
 \caption{A comparison between different variants of model expansion for this problem setting: either at the side of the input (\spectrans{}), the output (\spechead{}), or using adapters (\ada{}).}
 \label{fig:model-expansion-approaches}
 \vspace{-0.3cm}
\end{figure*}

For our base model, we use the same \mbert{}-based architecture as was used in~\citet{jointBERT-giuseppe-19} and \citet{xmetraada-mhamdi-naacl21} to jointly learn the intent classification and slot filling subtasks of MTOP.\footnote{More details about the base model can be found in Appendix~\ref{app:base-model}.}
On top of that, we define baselines, non-continual learning reference models, and continual learning algorithms. 
\vspace{-0.2cm}
\subsection{Baseline \& Reference Models}
\label{sec:baselines}

Before delving into continual learning approaches, we consider a simple lower-bound baseline. In addition to that, we design reference models either trained from scratch for each new language, in a joint manner, or in a sequential multi-hop manner. Those are upper-bound non-continual learning models that are used to assess the performance of different models trained with continual learning methodologies. Those reference models can be in general superior to continual learning models but can also be less efficient and not feasible. For a fair comparison, all models use the same base model architecture and its loss with no further additions or special optimizations to the architecture. 
\vspace{-0.25cm}
\paragraph{Lower-bound Baseline.} This consists of \textit{naive sequential fine-tuning} (\naiveft{}), which sequentially fine-tunes with no continual learning.
\vspace{-0.7cm}
\paragraph{Non-continual Learning Upper-bound Models.} These are stronger upper-bound models used as reference points of performance. However, they are either not efficient or prohibitive in the context of cross-lingual continual learning. Some of them require training from scratch for each language which is not efficient. Others require having access to all languages either at the same time or incrementally. Having such access can be restrictive due to privacy or storage efficiency concerns.  
\begin{itemize}[leftmargin=*]
\itemsep0em
\item{\textit{Language-specific fine-tuning} (\langspec{}). This trains independent models on the data set for each language $\lng{}_i$ using only $\datalang{i}$.}
\vspace{-0.15cm}
\item{\textit{Multi-lingual learning} (\multi{}). This trains one single model jointly across all data sets $\datalang{\leq N}$.}
\vspace{-0.6cm}
\item{\textit{Incremental joint learning} (\jointinc{}). This incrementally trains adding the data set for each language in the stream. This consists of the following hops: 1) $\datalang{\leq 1}$, 2) $\datalang{\leq 2}$, $\cdots$, and N-1) $\datalang{\leq N-1}$. This is the only sequential reference model.}
\end{itemize}
\vspace{-0.4cm}
\subsection{Continual Learning Approaches}
\label{sec:cont-learn-approaches}

To continually fine-tune on different languages, we establish a representative set of strong approaches\footnote{More details about the approaches and their hyper-parameters can be found in Appendix~\ref{app:details-approaches} and~\ref{app:hyperparameters}, respectively.} spanning the following categories inspired by previous evaluation paradigms such as~\citet{jin-etal-2022-lifelong-pretraining} lifelong language model domain-incremental pertaining. To the best of our knowledge, we are the first to exhaustively investigate such approaches for the context of cross-lingual continual learning, whereas different approaches were investigated separately for different problem definitions. 
\vspace{-0.2cm}
\paragraph{Model Expansion.}
We consider the following approaches that add hop-specific parameters, as shown in Figure~\ref{fig:model-expansion-approaches}. We expand on either the input (i.e. \mbert{} representations) or the output side (i.e. task-specific prediction heads). For the former (\spectrans{}), the transformer layers are replicated for each hop while sharing the prediction heads. To expand on the output side (\spechead{}), we use different prediction heads across hops but share the M-BERT layers. We additionally consider \specenca{}, which trains \mbert{} encoder layers $\in 1 \ldots 9$ in a language-specific manner while sharing the rest. We also separately add MAD-X adapters~\cite{madx-pfeiffer-emnlp20}. We either fine-tune the adapter layers and freeze the rest of \mbert{} (\adafrozen{}) or tune them both (\adatuned{}).\footnote{More details on adapters and how zero-shot evaluation works for model expansion approaches are in Appendix~\ref{app:model-expansion}.}
\vspace{-0.2cm}
\paragraph{Regularization.}
We focus on elastic weight consolidation (\ewc{})~\cite{ewc-kirkpatrick-nas17}, which mitigates catastrophic forgetting by reducing the changes in parameters that are deemed critical to previously seen languages. We use the online version of EWC (\ewconline{}) for efficiency.
\vspace{-0.2cm}
\paragraph{Memory Replay.}
We use experience replay (\er{})~\cite{er-chaudhry-arxiv19}, which alleviates forgetting by maintaining a fixed-size memory equally balanced between the different languages and regularly drawing examples from the memory to replay.
\vspace{-0.6cm}
\paragraph{Distillation-based.}
On top of ER, we distill dark knowledge~\cite{dark-knowledge} from previous model checkpoints. We explore two variants: logit distillation (\kdlogit{})~\cite{hinton-kdlogit-arxiv15} and representation distillation (\kdrep{})~\cite{aguilar-kdrep-aaai20}, which optimize the minimum squared error loss on either the output logits or \mbert{} representations between the current and previous models.  

\vspace{-0.2cm}
\section{Results \& Analysis}
\label{sec:results-analysis}
\vspace{-0.2cm}
In this section, we provide an extensive analysis in the form of different ablation studies. We ask critical analysis questions that revolve around the continual learning goals described in~\S\ref{sec:goals}. For~\S\ref{sec:q1-catastrophic-forgetting-generalization}, scores are reported using accuracy (Acc) and F1-score (F1) for intent classification and slot filling, respectively. For the remaining sections, all results are reported for intent classification only, slot-filling results, for which the same trends are observed, can be found in Appendix~\ref{app:more-results}. Bootstrap sampling (over test data shuffling) is used to compute the average and 95\% confidence intervals (averaged over all language permutations except for~\S\ref{sec:q2-analysis-across-lang-perm}). More details can be found in Appendix~\ref{app:boostrap}. We also separately repeat key experiments over 3 different seeds and obtain similar findings, which can be found in Appendix~\ref{app:seeds-experiments}. We decided to report the results using bootstrap sampling since they have tighter confidence intervals. 

\vspace{-0.2cm}

\subsection{How is a Multi-Hop Analysis Different from its One-Hop Counterpart?}
\label{sec:q6-multi-step-cross-cont-learn}

\begin{figure}[ht]
 \centering
\includegraphics[width=0.48\textwidth]{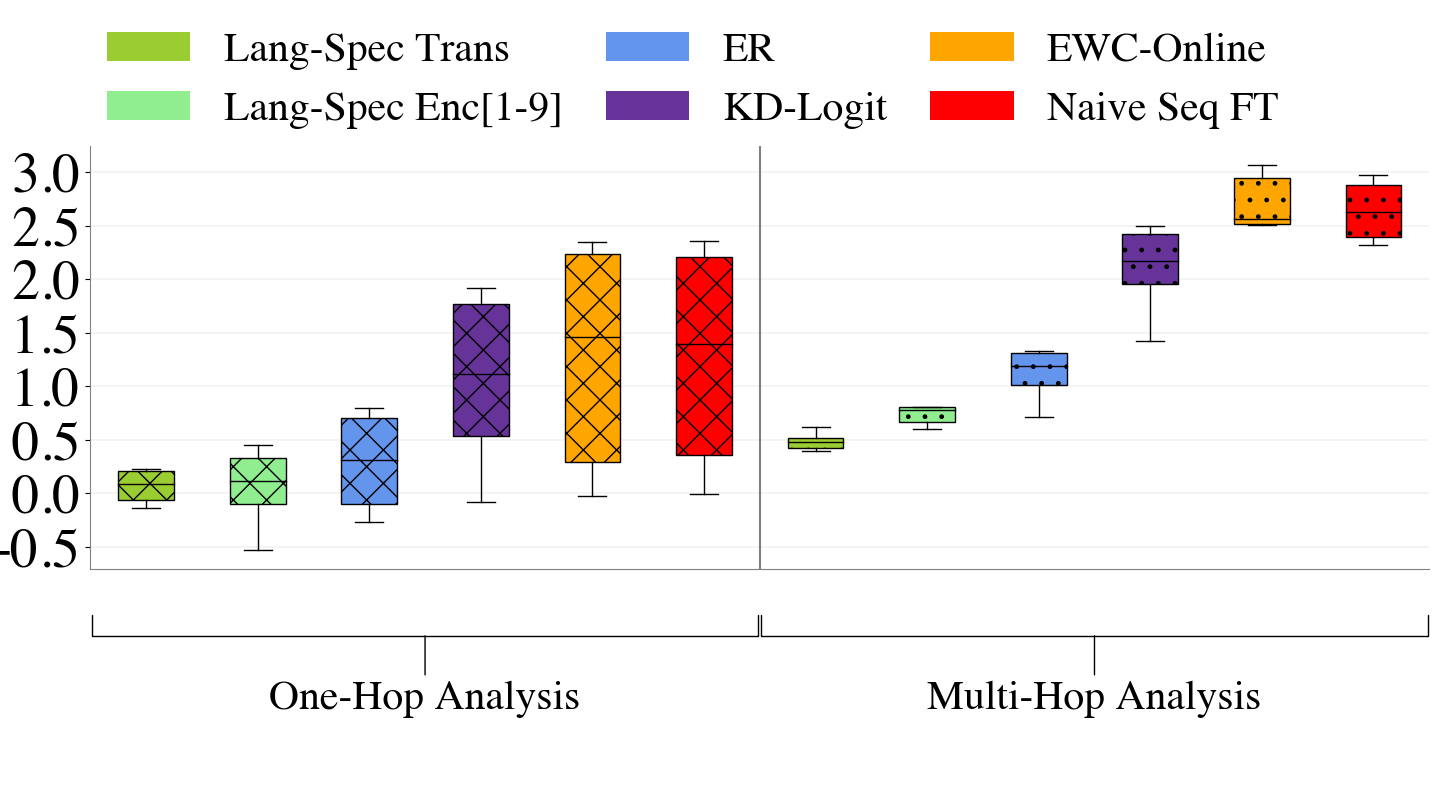}
\vspace{-0.6cm}
\caption{\label{fig:forg-steps-merged} Comparison between forgetting trends for intent classification using one-hop (crossed boxplots on the left) and multi-hop analysis (dotted boxplots on the right), showing the variance over different language permutations. One-hop analysis exhibits higher variance than its multi-hop counterpart.}
\vspace{-0.4cm}
\end{figure}

To motivate our cross-lingual continual learning evaluation paradigm, we start by investigating how a multi-hop analysis is different from a conventional one-hop transfer learning analysis. Figure~\ref{fig:forg-steps-merged} shows a comparison between the two in terms of forgetting (Eq.~\ref{eqn:forgetting}) for different approaches aggregated over different language permutations. More results for slot filling and other metrics can be found in Figure~\ref{fig:two-multi-steps-more} in Appendix~\ref{app:more-analysis}.   

\spectrans{} tends to have the least forgetting and \naiveft{} the most, but importantly \textbf{the variance for the multi-hop analysis is much smaller than that for the one-hop analysis}. Having larger confidence intervals, the one-hop analysis also tends to be misleading in the sense that certain models are depicted as having a good performance, while it is not truly the case. For example, \naiveft{}, according to the one-hop analysis, shows a range of forgetting from very little (0.5) to a lot (2.0). So in some circumstances, it has little forgetting, thus a good performance under the one-hop analysis. But according to the multi-hop analysis, it clearly has a lot of forgetting with more confidence. Therefore, the multi-hop analysis leads to a more conclusive analysis. We conjecture that averaging over more hops and balanced diversified datastreams is what leads to narrower confidence intervals. This agrees with the well-known fact that larger sample sizes lead to narrower confidence intervals~\cite{hazra_2017}.

\subsection{Can a Multi-lingual Language Model Learn to Preserve and Accumulate Knowledge across Different Languages?}
\label{sec:q1-catastrophic-forgetting-generalization}

\vspace{-0.1cm}
Given the conclusiveness of the multi-hop analysis in \S\ref{sec:q6-multi-step-cross-cont-learn}, we follow that type of analysis thereafter. In this section, we investigate how well the baseline and different non-continual learning reference models learn to preserve and accumulate knowledge across different languages by looking at the average over language permutations. Since not all reference models are sequential, we start by comparing them to the baseline using their final performances (Eq.~\ref{eqn:final-perf}). The final performance is indicative of how well a single final model can encapsulate the knowledge across all languages at the end of training. From Table~\ref{tab:cll-metrics-average-fp}, we notice that \naiveft{} and \multi{} have the worst and best final performances, respectively. This suggests that \textbf{a multilingual joint model is more beneficial than sequential models}. In practical scenarios, however, we may not have access to all languages at the same time. Among non-continual learning reference models, \jointinc{} is closest to \multi{} if all data may be preserved. However, this may also not be the case. In that case, \jointinc{} is nearly as good. Training incrementally and sequentially (\jointinc{}) is also more beneficial than fine-tuning on just the language of interest (\langspec{}), as the former exploits cross-lingual transfer capabilities.
\begin{table}[ht] 
\centering
\scalebox{0.8}{
\begin{tabular}{l|ll} \toprule  
\multicolumn{1}{c|}{\textbf{Model}} & \multicolumn{1}{c}{\textbf{Intent Class (Acc)}} & \multicolumn{1}{c}{\textbf{Slot Filling (F1)}} \\
\midrule
\naiveft{} & 91.06 $\pm 1.08$ & 69.37 $\pm 1.06$ \\
\langspec{} & 93.40 $\pm 0.08$ & 73.90 $\pm 0.83$ \\
\jointinc{} & \underline{94.16 $\pm 0.18$} & \underline{74.88 $\pm 0.38$} \\
\multi{} & \textbf{94.25 $\pm 0.07$} & \textbf{76.34 $\pm 0.82$} \\

\bottomrule
\end{tabular}
}
\caption{\label{tab:cll-metrics-average-fp} The average final performance across different language permutations for the baseline compared to reference models. We highlight the best scores in \textbf{bold} and \underline{underline} the second best across models. 
}
\vspace{-0.3cm}
\end{table}

We focus, thereafter, on \jointinc{}\footnote{We do not use \multi{} since it is non-sequential. Metrics like forgetting are thus always zero, which makes this model not comparable with other continual learning approaches and sequential reference models.} and compare its forgetting (Eq.~\ref{eqn:forgetting}) and transfer (Eq.~\ref{eqn:fwt}) trends to the baseline \naiveft{}, as shown in Table~\ref{tab:cll-metrics-average}. \jointinc{} exhibits significantly less forgetting, which also causes its final performance to be higher than \naiveft{}. This suggests that recalling previously used training data is helpful in knowledge preservation. However, \naiveft{} seems to slightly outperform \jointinc{} in terms of transfer. This difference is not statistically significant.\footnote{We report the p-values from pairwise Tukey's HSD analysis to gain a reliable unified view that individual t-tests may fail to convey. More explanation can be found in Appendix~\ref{app:boostrap}.} We hypothesize that this could be due to exposing \jointinc{} to all resources from previously seen languages, so it is likely that the data distribution between all these languages may distract the model from learning on the new one.

\begin{table}[ht!] 
\centering
\scalebox{0.68}{
\begin{tabular}{l|ll|ll} \toprule  
\multirow{2}{*}{\textbf{Model}}  
& \multicolumn{2}{c|}{Intent Class (Acc)} &  \multicolumn{2}{c}{Slot Filling(F1)}  \\ & F $\downarrow$  & T $\uparrow$ & F $\downarrow$ & T $\uparrow$  \\
\midrule
\naiveft{} & 2.93 $\pm 1.24$ & \textbf{0.68 $\pm 0.14$} & 5.67 $\pm 0.93$ & \textbf{1.37 $\pm 0.53$} \\
\jointinc{} & \textbf{0.11 $\pm 0.10$} & 0.52 $\pm 0.19$ & \textbf{0.91 $\pm 0.34$} & 0.83 $\pm 0.77$ \\
\bottomrule
\end{tabular}
}
\caption{\label{tab:cll-metrics-average} Forgetting (F) and transfer (T) performance averaged across different language permutations for \emph{sequential baseline and reference models}. We highlight the best models in \textbf{bold} for each subtask and metric.}
\vspace{-0.2cm}
\end{table}

\vspace{-0.2cm}

\begin{figure*}[ht!]
 \centering
\includegraphics[width=1\textwidth]{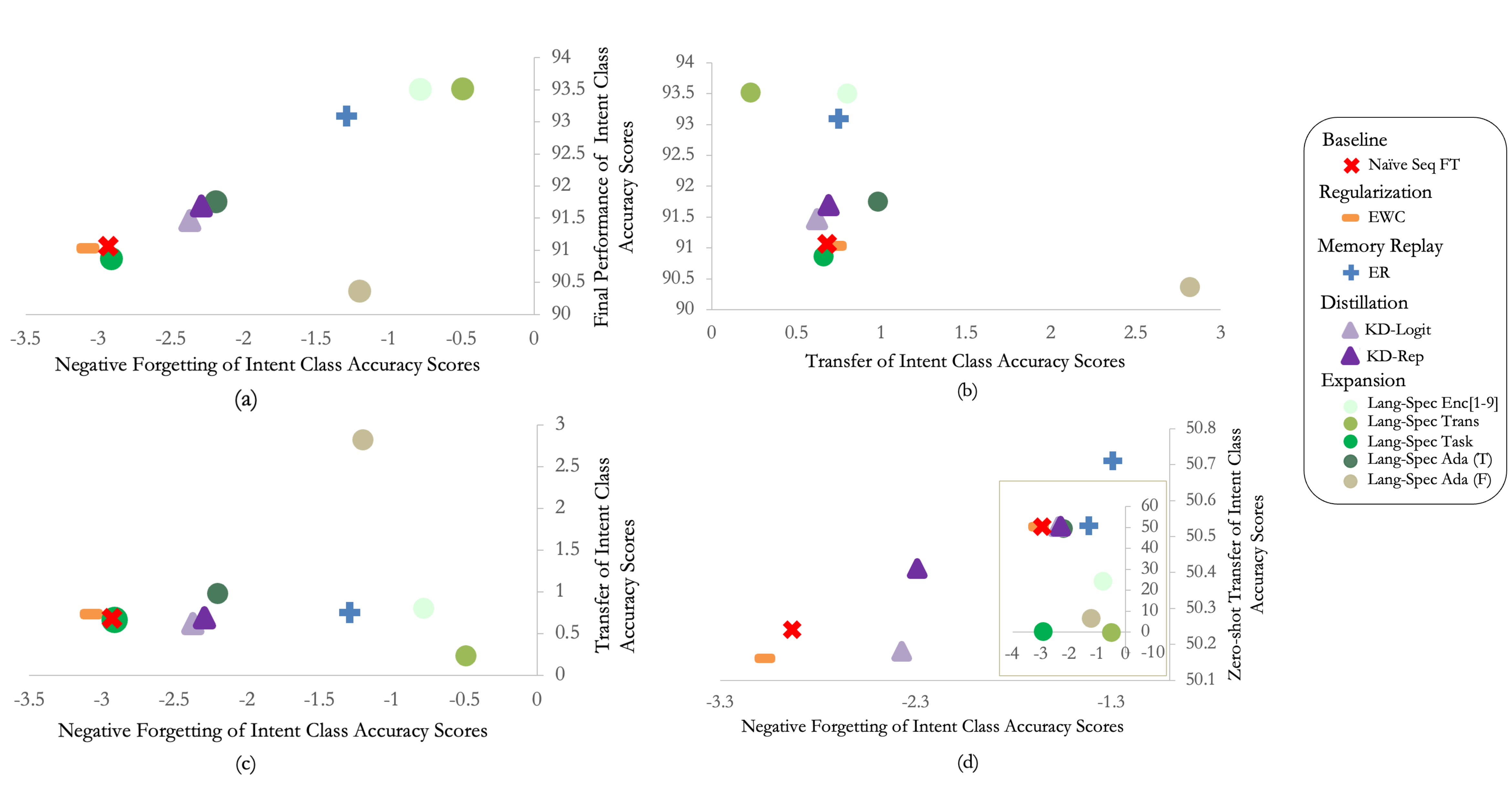}
 \caption{Correlations between different pairs of metrics: (a) Final performance versus negative forgetting for the task of intent classification. The lower the forgetting, the higher the final performance. (b) Final performance versus transfer for the task of intent classification. As hypothesized, there is no direct correlation between final performance and transfer. (c) Transfer versus negative forgetting for the intent classification task. In general, there is no direct correlation between transfer and forgetting. (d) Zero-shot generalization versus negative forgetting for intent classification. Model expansion approaches are highlighted in shades of green. We zoom over the rest of the models in the main graph and show an overview of all approaches in the lower right corner subplot. Mitigating forgetting leads to higher generalization, with the exception of multi-headed models highlighted in green.}
 \label{fig:all-correlation-intents}
 \vspace{-0.5cm}
\end{figure*}

\subsection{Is Continual Learning Effective in Boosting Knowledge Preservation, Accumulation, and Model Utility?}
\label{sec:q3-effectiveness-cross-cont-learn}

To study the effectiveness of continual learning approaches, we compare them to the baseline using the average over language permutations. We show, in Figures~\ref{fig:all-correlation-intents}(a) and~\ref{fig:all-correlation-intents}(c), the final performances (Eq.~\ref{eqn:final-perf}) and transfer (Eq.~\ref{eqn:fwt}) of different approaches, respectively, versus their negative forgetting (Eq.~\ref{eqn:forgetting}). In general, we observe that continual learning approaches mitigate forgetting and improve final performance. They also improve transfer, to some degree, though gains are mostly not significant compared to \naiveft{}(Appendix~\ref{app:boostrap}).

From Figure~\ref{fig:all-correlation-intents}(a), we notice that model expansion approaches\footnote{We include a full analysis of the expansion over several subsets of \mbert{} components in Appendix~\ref{app:comp-analysis}.}(\spectrans{} and \specenca{} described previously) are good at mitigating forgetting and improving the final performance while \spechead{} is not. \mbert{}, when trained in a language-specific manner, is responsible for encapsulating the cross-lingual representations necessary for enabling knowledge preservation, whereas changes to the downstream task-specific layers do not make much of a difference. This implies that in cross-lingual continual learning more attention should be paid to how to train those representations in a language-specific manner efficiently. \adatuned{} is one way to do this more efficiently, but its performance still lags behind other model expansion approaches. \er{} achieves a performance close to \spectrans{} and \specenca{} and this suggests that \textbf{using a portion of the memory is beneficial}.\footnote{An ablation study using different sizes of the memory is shown in Appendix~\ref{app:er-ablation-studies}. It shows that even smaller sizes up to 5\% are still beneficial. We report here the highest memory size as it leads to the best results.}



In the baseline approach, which suffers from the highest forgetting, we also notice the lowest final performance and transfer in Figures~\ref{fig:all-correlation-intents}(a) and~\ref{fig:all-correlation-intents}(c). As continual learning approaches reduce forgetting, they also improve the final performance, and some of them also improve transfer but not to the same degree. This suggests that \textbf{the lower the forgetting a model can achieve, the easier it gets for it to learn a stronger final model}. However, there is no direct correlation between forgetting and transfer. For example, \spectrans{} is the best model in reducing forgetting but also the worst in terms of transfer. This could be due to the fact that \spectrans{} exhibits similar behavior to \langspec{} thus, the transfer of a model, which is the difference between the performance of that model and that of \langspec{}, is almost null. On the other hand, although \adafrozen{} has the highest transfer, it has the lowest final performance and is close to average forgetting. Although the adapter will not be updated anymore after the model has been fine-tuned on, we think that the forgetting could be due to the shared task-specific layer leading to a forgetting closer to Lang-Spec Trans more than Lang-Spec Ada(T), which also shares M-BERT and tunes it. We show in Figure~\ref{fig:all-correlation-intents}(b) that there is no direct correlation between final performance and transfer. This posits that all three metrics need to be studied independently for a more insightful analysis.
\vspace{-0.3cm}


\subsection{Which Permutations Impose More Challenges on Knowledge Preservation, Accumulation, and Model Utility?}
\label{sec:q2-analysis-across-lang-perm}

So far, our analysis has focused on the average over language permutations, but are the same patterns observed for different language permutations? To shed light on this, we analyze the performance of different continual learning algorithms and the baseline in terms of their forgetting (Eq.~\ref{eqn:forgetting}), transfer (Eq.~\ref{eqn:fwt}), and final performance (Eq.~\ref{eqn:final-perf}) over \htol{} and \ltoh{} permutations, in Table~\ref{tab:short-metrics-perm}.\footnote{Full results for slot filling, more language permutations, and a balanced version of data can be found in Appendix~\ref{app:full-lang-perm}.}
In general, we observe that it is \textbf{more challenging to learn from low to high resource languages}. However, model expansion and memory replay approaches reduce forgetting and final performance gaps between language permutations. We hypothesize that \ltoh{} being more challenging than \htol{} could be due to the fine-tuning data size that is different between languages.

\begin{table}[t!]
\centering
\scalebox{0.7}{
\begin{tabular}{l|ll||ll||ll} \toprule  
 \multirow{2}{*}{\textbf{Model}} & \multicolumn{2}{c||}{F $\downarrow$} &  \multicolumn{2}{c||}{T $\uparrow$}  &  \multicolumn{2}{c}{FP $\uparrow$}          \\ 
 & \htol{} & \ltoh{} & \htol{} & \ltoh{} & \htol{} & \ltoh{} \\
 \toprule
\naiveft{} & \textbf{1.52} & 5.52 & \textbf{0.93} & 0.57 & \textbf{92.06} & 88.80 \\
\spectrans{} & \underline{\textbf{0.40}} & \underline{0.62} & \textbf{0.59} & 0.03 & \underline{\textbf{93.86}} & \underline{93.37} \\
\specenca{} & \textit{\textbf{0.60}} & \textit{1.05} & \textbf{1.00} & 0.63 & \textit{\textbf{93.75}} & \textit{93.15} \\
\spechead{} & \textbf{1.53} & 5.53 & \textbf{0.84} & 0.38 & \textbf{91.93} & 87.68 \\
\adatuned{} & \textbf{1.18} & 4.43 & \textit{\textbf{1.29}} & \textit{0.79} & \textbf{92.36} & 88.66 \\
\adafrozen{} & \textbf{0.84} & 1.87 & \underline{\textbf{3.41}} & \underline{2.43} & \textbf{91.08} & 89.92 \\
\ewc{} & \textbf{1.82} & 5.90 & \textbf{0.74} & 0.48 & \textbf{91.16} & 88.28 \\
\er{} & \textbf{0.71} & 2.35 & \textbf{0.95} & 0.78 & \textbf{93.51} & 92.58 \\
\kdlogit{} & \textbf{1.42} & 4.07 & \textbf{0.77} & 0.51 & \textbf{91.60} & 89.65 \\
\kdrep{} & \textbf{1.49} & 4.00 & \textbf{0.96} & 0.53 & \textbf{91.64} & 90.17 \\
 \bottomrule
\end{tabular}
}
\caption{\label{tab:short-metrics-perm} Comparison of intent classification for two language permutations. We highlight in \textbf{bold} the best forgetting (F), highest transfer (T), and final performance (FP) of accuracy scores among \htol{} and \ltoh{}, whereas the best and second best scores across approaches for \htol{} and \ltoh{} separately are \underline{underlined} and \textit{italicized}, respectively. We report mean performances for each metric and language order. All 95\% confidence intervals range from $\pm$ 0.01 to $\pm$ 0.04.}
\vspace{-0.4cm}
\end{table}

To verify this hypothesis, we dig deeper to check if the differences among fine-tuning data sizes between languages are the main factor by performing an ablation study on that. Therefore, we use the same amount of fine-tuning and evaluation resources for each language (9,219 for train, 1,285 for Dev, and 2,299 for test splits) and report the results on \naiveft{} in Table~\ref{tab:cll-equal}. We notice that there is still a gap between these two language permutations for forgetting and final performance. This suggests that the difference in fine-tuning data size is not what accounts for the differences between the two language permutations. There are perhaps biases in the pre-training or other linguistic artifacts that need to be studied in future work.
\vspace{-0.2cm}

\begin{table}[ht!] 
\small
\centering
\scalebox{0.88}{
\begin{tabular}{l|ll||ll||ll} \toprule  
\multirow{2}{*}{\textbf{Model}}  
& \multicolumn{2}{c||}{F $\downarrow$} &  \multicolumn{2}{c||}{T $\uparrow$} &  \multicolumn{2}{c}{FP $\uparrow$}  
\\
 & \htol{} & \ltoh{} & \htol{} & \ltoh{} & \htol{} & \ltoh{} \\
\toprule
Original Data & \textbf{1.52} & 5.52 & \textbf{0.93} & 0.57 & \textbf{92.06} & 88.80 \\    
Balanced Data & \textbf{1.25} & 5.81  & \textbf{0.89} & 0.75 & \textbf{89.33}  &  85.81 \\

\bottomrule
\end{tabular}
} \caption{\label{tab:cll-equal} Performance on intent classification comparison between two versions of the data: original data version and balanced data for \naiveft{} across the same permutations as Table~\ref{tab:short-metrics-perm}. We \textbf{bold} the best among \htol{} and \ltoh{} for each metric.}
\vspace{-0.2cm}
\end{table}

\subsection{How do Continual Learning Models Generalize to Unseen Languages?}
\label{sec:q5-zero-shot-generalization}
\vspace{-0.1cm}

To analyze the zero-shot transfer to languages unseen during fine-tuning, we plot the performance of zero-shot transfer (Eq.~\ref{eqn:fwt0}) as a function of negative forgetting over the average of different language permutations to investigate any relationships between generalization and preservation. In Figure~\ref{fig:all-correlation-intents}(d), we infer that \textbf{most continual learning approaches do not substantially improve generalization compared to \naiveft{}}. In particular, model expansion approaches (in red) hurt generalization even if they significantly reduce forgetting. This \textbf{zero-shot transfer versus interference trade-off} is referred to as the stability-plasticity dilemma~\cite{mermillod-stabilityplasticity-psychology13}, where the weights responsible for improving on new tasks are often responsible for forgetting previous tasks. Except for model expansion approaches, we notice that approaches that reduce forgetting also improve generalization compared to \naiveft{}. Better approaches to balance between the two can be investigated in future work.


\section{Related Work}
\label{sec:related-work}
\vspace{-0.2cm}
Continual learning for cross-lingual NLP work is under-explored, either focusing on proposing cross-lingual approaches that indirectly support continual learning, such as~\citet{artexte-monotransf-acl20}, of the transfer-ability of monolingual models. Other approaches derive a cross-lingual continual learning problem directly from cross-lingual transfer learning, such as~\citet{cont_multinmt-garcia-naacl21, pfeiffer-etal-2021-unks, minixhofer-etal-2022-wechsel}, who propose different lexical and semantic approaches to adapt to new low-resource languages for different downstream tasks. Similarly,~\citet{xcontlearn_ner_pos_gem-liu-repl4nlp21} explore continual techniques to fine-tune on downstream applications for new languages while preserving the original cross-lingual ability of the pre-trained model. ~\citet{muller-etal-2021-unseen} analyze the adaptability and usability of large language models to unseen and under-studied low-resource languages. However, they all focus on a one-hop analysis from high to low-resource language pairs or pre-training to fine-tuning tasks, unlike our work, which analyzes across multiple hops. 
More recently,~\citet{pfeiffer-etal-2022-lifting} propose a new methodology based on adapters and show that their approach mitigates negative interference between languages while enabling positive transfer. They use a multi-hop evaluation paradigm closer to our setup, but they only evaluate with respect to adapters using interference and transfer and do not analyze other aspects of cross-lingual continual learning capabilities.


\section{Conclusion}

We formulate the cross-lingual continual learning problem setup. We show that naive sequential fine-tuning is prone to catastrophic forgetting and has poor accumulation capabilities sensitive to different language permutations. We provide the first benchmark to compare the effectiveness of different continual learning algorithms for the cross-lingual case. We show that continual learning models improve cross-lingual knowledge preservation, which also contributes to improving final model performance and, to a lesser degree, accumulation and generalization. We also discuss the challenges of sequentially training for certain language permutations. We hope that this study will encourage more analyses in the same spirit to cover more benchmarks and datastream setups to gain more insights that go beyond conventional cross-lingual transfer learning.

\section*{Limitations}
\paragraph{Application to Other Benchmarks}
A central limitation of our work is that the main experiments are based on a single task-oriented dialogue benchmark. While there are multiple other natural language understanding benchmarks like XNLI, XQUAD, MLQA, and PAWS-X~\cite{conneau-xnli-emnlp18, artexte-monotransf-acl20,lewis-mlqa-acl20,yang-pawsx-acl19} that can also be used to back up our claims, we argue that this is outside the scope of this paper. The main objectives of this paper are to first come up with a new definition of a cross-lingual continual learning challenge and then to give an example using a comprehensive and realistic benchmark like task-oriented dialogue to catalyze more research in that direction. 
\paragraph{Choice of Realistic Permutations}
For more realistic setups of continual learning, we need to come up with an approach to define continual learning annotation scenarios of languages. Rather than using brute force with all possible ways the languages could be annotated at different stages, a principled way would be more desired. Since it is hard to tell if there is any logic or pattern in the annotation process itself and given the sheer amount of realistic scenarios, we chose one scenario experienced by some of the users: a model is built for a user, then the user reveals that more languages are desired. We test in our work the plausibility of continual learning approaches where the sequence moves from one language to another without repetition of the same language. Working on scenarios where the data from different languages are integrated as soon as they are annotated, implying different languages for different hops, is out of the scope of this paper. 
\paragraph{Data and Model Size Analysis}
In this paper, we pick certain model expansion approach variations to analyze the effect of model components (one aspect of model size) and two data distribution scenarios. However, analyzing extensively the effect of the scale of data and model size is beyond the scope of our work. We agree that different data sizes can be used, and it is interesting to analyze different supervision levels, such as using different proportions of the data for each language and simulating few-shot scenarios. We believe that for low-resource scenarios, we need to investigate specific approaches to continual learning, like meta-learning. We plan to investigate that in future work.
\paragraph{Application to Other Transformers}
Another possible limitation of our work is the restriction of the evaluation to a base model on top of \mbert{} Transformers. With the advent of Transformer-based encoders as strong pillars for transfer-learning, several Transformers such as \xlmrbase{} have been proposed more recently. Although those models have been shown to outperform \mbert{} on numerous downstream applications, especially on low-resource languages~\cite{xlm-r-conneau}, \mbert{} is still largely used due to its reduced number of parameters. In our specific continual learning challenge, efficiency is a top concern as we are training in multiple hops and benchmarking on different models. So, \mbert{} has been feasible in our use case. We leave experimenting with other Transformer-based encoders to future work. 

\section*{Acknowledgements}
This material is partially based upon work supported in part by the Office of the Director of National Intelligence (ODNI), Intelligence Advanced Research Projects Activity (IARPA), via Contract No. 2019-19051600007. The views and conclusions contained herein are those of the authors and should not be interpreted as necessarily representing the official policies, either expressed or implied, of ODNI, IARPA, or the U.S. Government. The U.S. Government is authorized to reproduce and distribute reprints for governmental purposes, notwithstanding any copyright annotation therein. We also would like to extend our thanks to Xisen Yin for insightful discussions on continual learning and all anonymous reviewers and meta-reviewers for their valuable feedback. 
\bibliographystyle{acl_natbib}
\bibliography{acl2023}

\appendix
\section{More Related Work}
\label{app:more-related-work}
Continual learning approaches have found favor, especially among the computer vision community, including regularization-based ~\cite{ewc-kirkpatrick-nas17,zenke-si-icml17,li_hoiem-lwf-eccv16,ritter-laplaceonline-neurips18} and memory-based approaches~\cite{shin-deepreplay-neurips17, er-chaudhry-arxiv19, agem-chaudhry-iclr19}. Only recently, continual learning has started gaining more interest in the NLP community. Most efforts on continual learning for NLP have focused on classification tasks and fall into the category of domain or class incremental continual learning~\cite{relextr-han-acl20}. Current approaches often fail to effectively retain previous knowledge and adapt to new information simultaneously~\cite{biesialska-survey-coling20,mbpa-demasson-nips19}. New challenges are formulated to study the problem of continual learning from different perspectives. \citet{jin-etal-2022-lifelong-pretraining} formulate the lifelong learning pretraining challenge, where pertaining language models continually adapt to emerging data from new corpora. 

Continual learning for cross-lingual NLP is under-explored, either focusing on proposing cross-lingual approaches that indirectly support continual learning, such as~\citet{artexte-monotransf-acl20}, on the transfer-ability of monolingual models. Other approaches derive a cross-lingual continual learning problem directly from cross-lingual transfer learning, such as~\citet{cont_multinmt-garcia-naacl21}, who propose a lexical approach to adapt to new low-resource languages for machine translation. Similarly, ~\citet{pfeiffer-etal-2021-unks} propose lexical-level adaptation schemes that can be applied to models relying on subword-based tokenization to adapt them to low-resource languages not covered or whose scripts are unseen during pre-training.~\citet{minixhofer-etal-2022-wechsel} also propose adaptations that go beyond the lexical level. Their approach facilitates the creation of monolingual language models that are transferable to new languages. ~\citet{xcontlearn_ner_pos_gem-liu-repl4nlp21} explore continual techniques to fine-tune on downstream applications for new languages while preserving the original cross-lingual ability of the pre-trained model. However, they all focus on a one-hop analysis from high to low-resource language pairs or pre-training to fine-tuning tasks, unlike our work, which analyzes across multiple hops. ~\citet{muller-etal-2021-unseen} analyze the adaptability and usability of large language models to unseen and under-studied low-resource languages. Based on that and depending on the degree of additional pre-training and fine-tuning required, they categorize the low-resource languages into easy, intermediate, and hard. Although this work paves the way for a better understanding of the mechanics of transferability to low-resource scenarios, they do not study the scenario where the transferability needs to be performed in multiple hops following a sequential stream of data. 
More recently,~\citet{pfeiffer-etal-2022-lifting} propose a new methodology for language-specific modules to add additional capacity and deal with the curse of multilinguality and show that their approach mitigates negative interference between languages while enabling positive transfer. They use a continual learning multi-hop evaluation paradigm which is closer to our setup, but they only evaluate using interference and transfer and only using one approach based on adapters and do not analyze other aspects of cross-lingual continual learning capabilities using a holistic approach like our work. 
\section{More Details about Approaches}
\label{app:details-approaches}

In this Section, we provide more details on the base model architecture used (Section ~\ref{app:base-model}) and different continual learning approaches in Sections~\ref{app:model-expansion},~\ref{app:ewc},~\ref{app:er}, and~\ref{app:kd}.

\subsection{Base Model Architecture}
\label{app:base-model}

\begin{figure}[ht]
\centering
\includegraphics[width=0.35\textwidth]{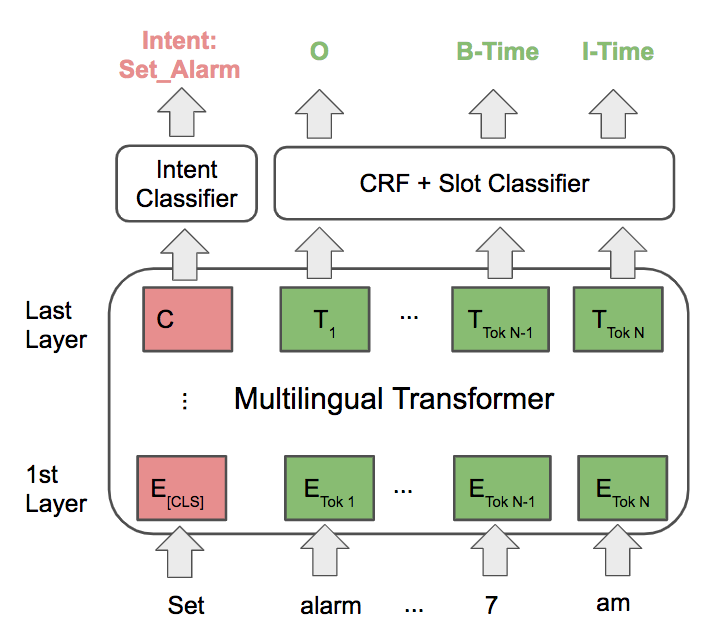}
\caption{\label{mtod-base-model} Architecture of task-oriented dialogue base model.}
\end{figure}

We use the same architecture as in~\citet{jointBERT-giuseppe-19} and~\citet{xmetraada-mhamdi-naacl21} to jointly learn intent classification and slot filling subtasks. As shown in Figure~\ref{mtod-base-model}, we leverage features from Transformer \cite{transformers-vaswani-nips17} encoder and add classification prediction heads on top of it. More specifically, a multi-lingual pre-trained model is used to encode the input. Then, to predict the intent and slot spans, we add task-specific prediction heads. For intent prediction, this takes the form of a linear layer plus softmax on top of the $[CLS]$ token representation. For slot filling, we use a sequence labeling layer in the form of a linear layer plus CRF, respectively. We use the sum of both intent and CRF-based slot losses to optimize the model parameters.

\subsection{Model Expansion}
\label{app:model-expansion}
Model expansion methods, such as \spectrans{} and \specenca{}, are fine-tuned for each language with either an entirely or partially language-specific M-BERT (whole 12 layers in addition to the embeddings or just the top 8 layers in the case of \spectrans{} and \specenca{} respectively). When fine-tuning them on a new language, the previously tuned parameters on the old languages are retained unchanged while the rest of the parameters that are not language-specific are fine-tuned. During the evaluation on a particular language, the tuned parameters for that language are restored and used if the language has been seen in training. Otherwise, the parameters initialized from M-BERT (before fine-tuning on any language) are used for zero-shot evaluation.

Adapters consist of downsampling layers followed by upsampling layers inserted between layers of our Transformer encoder in addition to their invertible components. We do not add task-specific adapters, as according to our ablation studies, they didn't prove beneficial. We add adapter components to every encoder layer following MAD-X configuration and using their pre-trained weights.\footnote{obtained from AdapterHub~\cite{adapterhub-pfeiffer-emnlp20} \url{https://adapterhub.ml/explore/text_lang/}} We either fine-tune the weights for the languages available in AdapterHub or train from scratch for languages for which there are no pre-training adapter weights. At inference time, we use adapter layers fine-tuned independently for each language in the datastream.

\subsection{Online Elastic Weight Consolidation (EWC-Online)}
\label{app:ewc}
To penalize changes in the parameters crucial to previous languages, we use EWC, which adds a regularization term to the loss applied only after the first data set $\datalang{i}$ in the language stream is seen. $\forall i \in 2 \ldots N$, we compute the total loss as follows:
\begin{equation}
    \mathcal{L}^i_{total} = \mathcal{L}^i_{cur} + \lambda \mathcal{L}^i_{reg},
\end{equation}

where $\mathcal{L}_{cur}$ is the usual loss of the downstream task on the current data $\datalang{i}$ and $\mathcal{L}_{reg}$ is the regularization term, and $\lambda$ is a hyperparameter to control the regularization strength (which is fixed to 20). For efficiency purposes, we use the online version of EWC (\ewconline{}). Following that, our regularization term is computed as, based on the formulation in~\citet{gido-threetypes-nature2022}: 
\begin{equation}
    \mathcal{L}^i_{reg} = \sum^{N_p}_{j=1}{\tilde{F}^{(i-1)}_{jj}(\theta_j - \theta^k_j)^2},
\end{equation}

where $\theta$ are the parameters of the Transformer model in addition to the downstream prediction heads, $N_p$ is the total number of parameters, and $\tilde{F}^{(i-1)}_{jj}$ is the Fisher information matrix on the last language just before training on $\datalang{i}$. This is computed as the running sum of the $i^{th}$ diagonal elements of the Fisher Information matrices of $\datalang{j}$, for all $j \in 1 \ldots (i-1)$. $\tilde{F}^{(i)}_{jj} = \gamma \tilde{F}^{(i-1)}_{jj} + F^{i}_{jj}$ and $\tilde{F}^{1}_{jj} = F^{1}_{jj}$. In practice, $F^{i}$ is simply the gradients of all parameters flattened into one single matrix. 

\subsection{Experience Replay (ER)}
\label{app:er}
After training for each $\datalang{i}$ for all $i \in 1 \ldots N$, we populate the memory with randomly sampled examples from $\datalang{i}$. For each $\datalang{i}$ for all $i \in 2 \ldots N$, after training for every $k=100$ mini-batches and optimizing for the current loss separately, the model randomly samples an equal batch from the memory for each $\datalang{j}$ such that $j \in 1 \ldots (i-1)$ and replays them using the current model checkpoint used for training on $\datalang{i}$. We retrieve an equal amount of memory items from each language and at each step and hop. The loss from the current $\datalang{i}$ and the loss on the memory on the $\datalang{j}$ are interleaved as the replay on the memory only happens every $k$ steps. This prioritization of the current language helps make the training more stable without over-fitting on the small memory from previous languages.  
\subsection{Knowledge Distillation (KD-Logit \& KD-Rep)}
\label{app:kd}
We use the same strategy explained in~\S\ref{app:er} to select the memory to be replayed using a knowledge distillation loss. For each $\datalang{i}$ for all $i \in 2 \ldots N$, after training for every $k=100$ mini-batches, we randomly sample an equal batch from the memory for each $\datalang{j}$ such that $j \in 1 \ldots (i-1)$. We also load the model checkpoints for each $hop_j$ and use that model and the memory for $\datalang{j}$ to compute either the intent and slot logits in the case of \kdlogit{} or the multilingual representations of M-BERT in the case of \kdrep{}. We do the same thing using the current model checkpoint this time. Then, we use the minimum square error loss to minimize the distance between the intent logits obtained using the previous and current model checkpoints and do the same thing for slot logits for \kdlogit{}. Then, we take the same over intent and slot distillation losses across different languages retrieved from the memory. The same is done for computing the distillation loss over the multilingual representations in \kdrep{}.

\section{Experimental Setup Details}
\label{app:exp-setup}

In this Section, we describe the datastreams used (Section~\ref{app:datastream}) and provide more implementation details (Section~\ref{app:hyperparameters}), in addition to bootstrap sampling and statistical significance specifics (Section~\ref{app:boostrap}). 

\subsection{Datastreams}
\label{app:datastream}


We use the following datastreams for all our experiments as summarized in Table~\ref{tab:simulated-perm}. The MTOP dataset has been released by Facebook~\cite{mtop-li-eacl21} under Creative Commons Attribution-ShareAlike 4.0 International Public License. 

\begin{table}[ht!]
\centering
\scalebox{0.70}{
\begin{tabular}{llllll}
\toprule
\textbf{Order 1} & \textbf{Order 2} & \textbf{Order 3} & \textbf{Order 4} & \textbf{Order 5} & \textbf{Order 6} \\ \toprule
English &  Thai    & Spanish  & French  & Hindi   &  German\\ 
German  &  Spanish & Hindi    & Thai    & English &  French\\
French  &  Hindi   & English  & German  & Spanish &  Thai \\ 
Hindi   &  French  & German   & English & Thai    &  Spanish\\ 
Spanish &  German  & Thai     & Hindi   & French  &  English \\ 
Thai    &  English & French   & Spanish &  German &  Hindi\\  \bottomrule
\end{tabular}
}
\caption{Simulated language permutations.}
\label{tab:simulated-perm}
\end{table}

\subsection{Implementation Details}
\label{app:hyperparameters}
For all experiments, we use M-BERT(bert-base-multilingual-cased)\footnote{\url{github.com/huggingface/transformers} version 3.4.0 pre-trained on 104 languages, including all languages evaluated on in this paper.} with 12 layers as our pre-trained Transformer model.
We use the dev set to pick the hyperparameters of the optimizer to be used. We perform a manual search for the most optimal learning rate over a range $[1e-4$, $3e-4$, $1e-5$, $3e-5]$ for Adam optimizer~\cite{adam-kingma-iclr15} and finally based on the dev performance we fix the learning rate to $3e-5$ for all experiments for a fair comparison. We use $\epsilon = 1e-8$, $\beta_1 = 0.9$, $\beta_2 = 0.99$, batch size of 16, $\gamma = 0.1$ for EWC Online, $6000$ memory size for ER and knowledge distillation. For all experiments, we run for $10$ epochs maximum and pick the best model based on dev data. We also fix a seed of 42 for the random initialization of Numpy, random, and torch over all bootstrap experiments. For additional experiments using multiple seeds, we fix three seeds. All experiments are run using the same computing infrastructure Pytorch version 1.7.1, using \emph{one} Tesla P100 GPU of $16280$ MiB of memory CUDA version 11.2.

The runtime and the number of parameters depend on the approach and the mode of training used as shown in Table~\ref{tab:runtime}. With the exception of model expansion and language-specific approaches, all approaches have the same number of parameters coming from the sum of \mbert{} and prediction head parameters. \spectrans{} has the highest number of parameters, which is six times more than \naiveft{} but only requires two times more runtime as only one $\frac{1}{6}$ part of language-specific \mbert{} is updated at each hop for each whereas the rest is used in evaluation mode only. \adafrozen{} has the smallest number of parameters which is around 24\% and takes 2 times less than the usual runtime of \naiveft{} (while exhibiting lower forgetting and higher transfer than \naiveft{}, as shown in Table~\ref{tab:full-cll-metrics-average}). Memory replay and knowledge distillation approaches have more runtime (slightly more than \spectrans{}) as they store and handle memory and compute the replay or distillation losses interleaved with the main loss, which makes them time-consuming. What impacts the runtime of ER is much more than just iterating over a small sampled memory. Its runtime does not only depend on the size of the memory as much as it depends on the frequency of interleaving happening at the fine-tuning schedule. After each k minibatch steps, we sample a minibatch from the memory and fine-tune on it interleaved with the fine-tuning on the main minibatch. So, that makes the runtime depend on k and not only the size of the memory. This make its training more time-consuming than if we had to sample only after each epoch with the same memory size.

\begin{table}[ht!]
\centering
\scalebox{0.7}{
\begin{tabular}{llll}
\toprule
\textbf{Model} & \textbf{Runtime} & \textbf{\# Param} \\ \toprule
\naiveft{} &  3h16min   &     178,081,402   \\
\langspec{} &  5h02min   &  1,068,488,412   \\
\jointinc{} &   1d22h51min   &   178,081,402      \\
\multi{} &   16h45min  & 178,081,402   \\
\specemb{} & 7h46min    &    639,123,322      \\
\specencaa{} & 7h52min   &  284,399,482      \\
\specencbb{} & 7h12min    &  284,399,482   \\
\specenccc{} & 7h8min    &  284,399,482      \\
\specencb{} & 7h20min   &   284,399,482     \\
\specenca{} & 8h1min    &    497,035,642      \\
\spectrans{} & 7h15min    &  1,067,348,602      \\
\specencall{} & 7h53min     &   603,353,722      \\
\specencaaa{} & 7h16min    &   390,717,562      \\
\specencbbb{} & 7h10min    &  390,717,562      \\
\spechead{} &  6h18min   &     179,221,212    \\
\adatuned{} & 4h34min   &  222,301,402         \\
\adafrozen{} & 1h57min   &  44,447,962        \\
\ewconline{} & 1d3h17min    &   178,081,402     \\
\er{} & 8h55min    &     178,081,402      \\
\kdlogit{} &  7h23min    &   178,081,402      \\
\kdrep{} &  8h   &    178,081,402     \\
  \bottomrule
\end{tabular}
}
\caption{Runtime and parameters statistics.}
\label{tab:runtime}
\end{table}

\subsection{Bootstrap Sampling \& Statistical Significance}
\label{app:boostrap}
We run all experiments over one fixed seed of 42. We then use bootstrap sampling~\cite{statistical-koehn-emnlp04} to compute the mean and confidence intervals for each of the metrics described in~\S\ref{sec:metrics} over a single approach. For each language permutation, and for each $R_{i,\leq{j}}$, representing some performance metric on language $\lng{}_i$ after training on $\datalang{\leq j}$, we sample with replacement 600 sentences from the testing data over 600 iterations. By using this number of iterations and sampling sentences, we ensure and also double-check that all sentences in the test set are covered in the evaluation, ensuring a uniform evaluation across approaches. Let $x$ be the list of results we get for each iteration independently. Then, we compute the mean and standard deviation $\bar{x}$ and $std(x)$ respectively and the 95\% confidence interval size $CI$ using the following equation:
\begin{ceqn}
    \begin{align}
    \label{eqn:boostrap}
    \begin{split}
        CI = \frac{1.9639 \times std(x) }{\sqrt{600}},
        \\
        std(x) = \sqrt{\frac{\sum{(x-\bar{x})^2}}{600}}.
    \end{split}
    \end{align}
\end{ceqn}

This computes $x$ and $CI$ for each language permutation separately. To aggregate this across different language permutations, we simply take the average and the standard deviation.

To compute the statistical significance between different approaches, we use ANOVA and perform a multiple pairwise comparisons analysis using Tukey's honestly significant difference (HSD) test\footnote{We use bioinfokit library \url{https://github.com/reneshbedre/bioinfokit}} over different language permutations for each metric.


\section{More Results \& Analysis using Boostrap Sampling}
\label{app:more-results}

In this Section, we present more results averaged over language orders in addition to fine-grained results and ablation studies. 

\subsection{Full Average Results}
\label{app:full-avg-results}

Table~\ref{tab:full-cll-metrics-average} shows the full results and confidence intervals for different continual learning approaches. Compared to intent classification, we observe a higher forgetting and slightly higher transfer but a lower zero-shot transfer and final performance in the case of slot filling. This could be due to the nature of the task of slot filling, which is more challenging to learn. In general, we can observe the same forgetting, transfer, zero-shot transfer, and final performance trends between intent classification and slot filling. In other words, if a model $a$ has higher forgetting of intent classification than model $b$ then the same thing applies to slot filling. This could be due to the transfer between intent classification and slot filling that is maximized when training them jointly. The best model for transfer is \adafrozen{}, which we hypothesize is due to its lightweight adaptation to the current language, which makes it overfit on that at the cost of a lower average and final performance overall.

\begin{table*}[ht] 
\centering
\scalebox{0.71}{
\begin{tabular}{l|ll|ll|ll|ll} \toprule  
\multirow{2}{*}{\textbf{Model}}  
& \multicolumn{2}{c|}{F $\downarrow$} &  \multicolumn{2}{c|}{T $\uparrow$} &  \multicolumn{2}{c|}{$T^0$ $\uparrow$} &  \multicolumn{2}{c}{FP $\uparrow$}  \\ & Acc & F1 & Acc & F1 & Acc & F1 & Acc & F1 \\
\midrule \rowcolor{lightgray} \multicolumn{9}{c}{Shared \{Trans, Task\} Baselines}   \\ \midrule 
\naiveft{} & 2.93 $\pm 1.24$ & 5.67 $\pm 0.93$ & 0.68 $\pm 0.14$ & 1.37 $\pm 0.53$ & 50.24 $\pm 3.43$ & 36.32 $\pm 1.91$ & 91.06 $\pm 1.08$ & 69.37 $\pm 1.06$ \\
\langspec{} &  & &  &  &  & & 93.40 $\pm 0.08$ & 73.90 $\pm 0.83$  \\
\langspecadatuned{} &  & &  &  &  & & 93.04 $\pm 0.09$ & 72.90 $\pm 0.80$  \\
\langspecadafrozen{} &  & &  &  &  & & 88.79 $\pm 0.13$ & 67.46 $\pm 0.89$  \\
\jointinc{} & \textbf{0.11 $\pm 0.10$} & \textbf{0.91 $\pm 0.34$} & 0.52 $\pm 0.19$ & 0.83 $\pm 0.77$ & 50.07 $\pm 2.48$ & \underline{36.39 $\pm 2.60$} & \underline{94.16 $\pm 0.18$} & \underline{74.88 $\pm 0.38$} \\
\multi{} &  & &  &  &  & & \textbf{94.25 $\pm 0.07$} & \textbf{76.34 $\pm 0.82$}  \\
\rowcolor{lightgray} \multicolumn{9}{c}{Model Expansion Baselines}\\\midrule
\textbf{\spectrans{}} & \underline{0.49 $\pm 0.08$} & \underline{1.32 $\pm 0.23$} & 0.23 $\pm 0.21$ & 0.95 $\pm 0.21$ & -0.43 $\pm 0.16$ & 0.42 $\pm 0.06$ & 93.51 $\pm 0.18$ & 74.74 $\pm 0.20$ \\
\textbf{\specenca{}} & 0.78 $\pm 0.15$ & 1.95 $\pm 0.51$ & 0.80 $\pm 0.19$ & 1.44 $\pm 0.71$ & 24.23 $\pm 1.73$ & 12.32 $\pm 1.24$ & 93.50 $\pm 0.21$ & 74.19 $\pm 0.92$ \\
\textbf{\spechead{}} & 2.91 $\pm 1.26$ & 5.26 $\pm 1.01$ & 0.66 $\pm 0.18$ & 1.15 $\pm 1.15$ & 0.10 $\pm 0.25$ & 0.07 $\pm 0.02$ & 90.86 $\pm 1.46$ & 69.41 $\pm 1.57$ \\
\textbf{\adatuned{}} & 2.19 $\pm 1.12$ & 4.23 $\pm 1.26$ & \underline{0.98 $\pm 0.18$} & \underline{2.04 $\pm 0.92$} & 49.35 $\pm 3.64$ & 33.60 $\pm 2.98$ & 91.75 $\pm 1.39$ & 71.13 $\pm 1.68$ \\
\textbf{\adafrozen{}} & 1.20 $\pm 0.35$ & 3.35 $\pm 0.85$ & \textbf{2.82 $\pm 0.33$} & \textbf{3.93 $\pm 0.68$} & 6.52 $\pm 2.16$ & 2.80 $\pm 0.59$ & 90.36 $\pm 0.37$ & 68.55 $\pm 1.10$ \\
\midrule \rowcolor{lightgray}  \multicolumn{9}{c}{Other Continuous Learning Algorithms }\\\midrule
\textbf{\ewc{}} & 3.07 $\pm 1.32$ & 5.78 $\pm 1.00$ & 0.73 $\pm 0.12$ & 1.46 $\pm 0.65$ & 50.16 $\pm 3.48$ & 36.31 $\pm 1.94$ & 91.03 $\pm 1.26$ & 69.63 $\pm 1.52$ \\ \midrule 
\textbf{\er{}} & 1.29 $\pm 0.51$ & 3.06 $\pm 0.59$ & 0.75 $\pm 0.17$ & 1.47 $\pm 0.85$ & \textbf{50.71 $\pm 3.55$} & \textbf{36.91 $\pm 2.14$} & 93.09 $\pm 0.29$ & 73.00 $\pm 0.52$ \\ \midrule 
\textbf{\kdlogit{}} & 2.37 $\pm 0.83$ & 5.53 $\pm 0.96$ & 0.62 $\pm 0.15$ & 1.40 $\pm 0.68$ & 50.18 $\pm 3.14$ & 36.25 $\pm 1.91$ & 91.46 $\pm 0.87$ & 69.64 $\pm 1.58$ \\
\textbf{\kdrep{}} & 2.29 $\pm 0.80$ & 5.35 $\pm 0.69$ & 0.69 $\pm 0.20$ & 1.43 $\pm 0.59$ & \underline{50.41 $\pm 2.92$} & 36.26 $\pm 1.96$ & 91.69 $\pm 0.71$ & 70.03 $\pm 1.09$ \\
\bottomrule
\end{tabular}
}
\caption{\label{tab:full-cll-metrics-average} A summary of results for different continual learning approaches over the average across language order. For each metric and score, we highlight the best score in \textbf{bold} and \underline{underline} the second best score.}
\end{table*}

\begin{table*}[ht] 
\small
\centering
\scalebox{0.83}{
\begin{tabular}{l|ll|ll|ll|ll} \toprule  
\multirow{2}{*}{\textbf{Model}}  
& \multicolumn{2}{c|}{F $\downarrow$} &  \multicolumn{2}{c|}{T $\uparrow$} & \multicolumn{2}{c|}{$T^0$ $\uparrow$} &  \multicolumn{2}{c}{FP $\uparrow$}  \\ & Acc & F1 & Acc & F1 & Acc & F1 & Acc & F1 \\
\midrule 
\naiveft{} & 2.93 $\pm 1.24$ & 5.67 $\pm 0.93$ & 0.68 $\pm 0.14$ & 1.37 $\pm 0.53$ & \textit{50.24 $\pm 3.43$} & \textit{36.32 $\pm 1.91$} & 91.06 $\pm 1.08$ & 69.37 $\pm 1.06$ \\ \midrule
\langspec{} &  & &  &  &  & & 93.40 $\pm 0.08$ & 73.90 $\pm 0.83$ \\
\spectrans{} & \underline{\textbf{0.49 $\pm 0.08$}} & \underline{1.32 $\pm 0.23$} & 0.23 $\pm 0.21$ & 0.95 $\pm 0.21$ & -0.43 $\pm 0.16$ & 0.42 $\pm 0.06$ & \underline{93.51 $\pm 0.18$} & \textbf{74.74 $\pm 0.20$} \\ \midrule
\specencall{} & \underline{\textbf{0.49 $\pm 0.08$}} & \textbf{1.30 $\pm 0.16$} & 0.23 $\pm 0.21$ & 0.77 $\pm 0.31$ & -0.31 $\pm 0.18$ & 0.57 $\pm 0.09$ & \textbf{93.52 $\pm 0.12$} & \underline{74.51 $\pm 0.25$} \\
\specemb{} & 3.13 $\pm 1.35$ & 5.88 $\pm 0.95$ & 0.74 $\pm 0.20$ & 1.24 $\pm 0.79$ & \underline{50.67 $\pm 2.98$} & \underline{36.62 $\pm 1.89$} & 90.69 $\pm 1.28$ & 69.59 $\pm 1.23$ \\
\midrule
\specencaa{} & 1.88 $\pm 0.77$ & 4.32 $\pm 0.69$ & 0.77 $\pm 0.19$ & 1.37 $\pm 0.64$ & \textbf{52.20 $\pm 3.23$} & \textbf{37.42 $\pm 1.99$} & 92.25 $\pm 0.76$ & 71.59 $\pm 1.52$ \\
\specencbb{} & 1.47 $\pm 0.65$ & 2.87 $\pm 0.36$ & \textit{0.78 $\pm 0.23$} & \underline{1.61 $\pm 0.45$} & 47.83 $\pm 3.00$ & 34.66 $\pm 1.79$ & 92.71 $\pm 0.65$ & 73.06 $\pm 0.97$ \\
\specenccc{} & 1.45 $\pm 0.56$ & 3.02 $\pm 0.52$ & 0.70 $\pm 0.16$ & 1.32 $\pm 0.52$ & 38.33 $\pm 3.00$ & 23.68 $\pm 2.36$ & 92.43 $\pm 0.78$ & 72.28 $\pm 1.05$ \\
\specencb{} & 2.21 $\pm 0.86$ & 4.14 $\pm 0.84$ & 0.47 $\pm 0.24$ & 1.35 $\pm 0.56$ & 41.38 $\pm 2.13$ & 20.04 $\pm 1.89$ & 91.41 $\pm 1.08$ & 71.14 $\pm 1.13$ \\
\midrule
\specencaaa{} & 1.27 $\pm 0.67$ & 2.99 $\pm 0.62$ & \textbf{0.87 $\pm 0.17$} & \textbf{1.64 $\pm 0.65$} & 45.23 $\pm 2.56$ & 31.21 $\pm 2.17$ & 92.92 $\pm 0.52$ & 73.33 $\pm 1.09$ \\
\specencbbb{} & 1.66 $\pm 0.36$ & 3.37 $\pm 0.69$ & 0.31 $\pm 0.33$ & 0.65 $\pm 0.73$ & 6.04 $\pm 1.13$ & 4.53 $\pm 0.96$ & 91.97 $\pm 0.38$ & 71.63 $\pm 1.15$ \\
\midrule
\specenca{} & \textit{0.78 $\pm 0.15$} & \textit{1.95 $\pm 0.51$} & \underline{0.80 $\pm 0.19$} & \textit{1.44 $\pm 0.71$} & 24.23 $\pm 1.73$ & 12.32 $\pm 1.24$ & \textit{93.50 $\pm 0.21$} & \textit{74.19 $\pm 0.92$} \\
\specencb{} & 2.21 $\pm 0.86$ & 4.14 $\pm 0.84$ & 0.47 $\pm 0.24$ & 1.35 $\pm 0.56$ & 41.38 $\pm 2.13$ & 20.04 $\pm 1.89$ & 91.41 $\pm 1.08$ & 71.14 $\pm 1.13$ \\
\bottomrule
\end{tabular}
}
\caption{\label{tab:full-cll-components} Per group layer analysis: ablation studies of different \mbert{}'s components. Best, second best, and third best scores for each metric are in \textbf{bold}, underlined, and \textit{italicized} respectively.}
\end{table*}

\subsection{Per M-BERT Components Analysis}
\label{app:comp-analysis}

Table~\ref{tab:full-cll-components} shows ablation studies for the analysis of \mbert{} components following four different categories: groups of 12 layers with or without embeddings, groups of 3 layers, 6 layers, and 9 layers at a time trained in a language-specific manner and the rest shared between languages. We notice that training the full \spectrans{} and \specencall{} have the best in terms of forgetting and final performance. Training only the first 8 encoder layers \specenca{}, excluding embeddings, in a language-specific manner comes next in terms of a low forgetting and a comparable final performance, with a relatively better transfer and zero-shot transfer performance. Other good models reaching a good compromise between transfer, zero-shot transfer, and forgetting with less language-specific layers are \specencaa{} and \specencaaa{}. \naiveft{} is comparable to those model-expansion approaches in terms of zero-shot performance but has lower final performance and significantly higher forgetting. We also notice the same trend for language-specific embeddings \specemb{}, which reaches the second-best zero-shot transfer performance but with also a high forgetting. This suggests that language-specific knowledge is less likely to be encoded in the embeddings and more at the encoder layers. This shows that there is a real plasticity-stability tradeoff between zero-shot transfer and knowledge preservation (which we explain in more detail in~\S\ref{sec:q5-zero-shot-generalization}). 

\subsection{Full Results on Language Permutations}
\label{app:full-lang-perm}

Full results for all language permutations can be found in Tables~\ref{tab:full-cll-metrics-perm},~\ref{tab:full-cll-metrics-perm-1}, and~\ref{tab:full-cll-metrics-perm-2}. By looking at additional language permutations, \ltoh{} (Thai $\rightarrow$ Spanish $\rightarrow$ Hindi $\rightarrow$ French $\rightarrow$ German $\rightarrow$ English) is still the most challenging one in terms of knowledge preservation, accumulation, generalization, and model utility. \htol{} (English $\rightarrow$ German $\rightarrow$ French $\rightarrow$ Hindi $\rightarrow$ Spanish $\rightarrow$ Thai) is still the easiest to learn. Order 5(Hindi $\rightarrow$ English $\rightarrow$ Spanish $\rightarrow$ Thai $\rightarrow$ French $\rightarrow$ German) is the second most challenging language permutation to train. In general, the same trends regarding the more challenging nature of training for certain language permutations are observed for both intent classification and slot filling uniformly. Table~\ref{tab:full-cll-equal} includes the results for more language permutations for the balanced data.

\begin{table*}[ht!] 
\small
\centering
\scalebox{0.82}{
\begin{tabular}{l|llll|llll} \toprule  
\multirow{3}{*}{\textbf{Model}}  
& \multicolumn{4}{c|}{\htol{}} & \multicolumn{4}{c}{\ltoh{}} \\ 
&
 \multicolumn{8}{c}{\textbf{Test Intent Accuracy On}} \\
& F $\downarrow$ &  T $\uparrow$ & $T^0$ $\uparrow$ &  FP $\uparrow$ &  F $\downarrow$ & T $\uparrow$ & $T^0$ $\uparrow$  & FP $\uparrow$  \\ 
\midrule \rowcolor{lightgray} 

\rowcolor{lightgray} \multicolumn{9}{c}{Shared \{Trans, Task\} Baselines }   \\ \midrule 
\naiveft{} & \textbf{1.52 $\pm 0.02$} & \textbf{0.93 $\pm 0.02$} & \textbf{50.68 $\pm 0.03$} & \textbf{92.06 $\pm 0.02$} & 5.52 $\pm 0.04$ & 0.57 $\pm 0.01$ & 44.66 $\pm 0.02$ & 88.80 $\pm 0.02$ \\
\langspec{} &  & & & 93.40 $\pm 0.08$  &  & & & 93.40 $\pm 0.08$\\
\langspecadatuned{} &  & & & 93.04 $\pm 0.09$ &  & & & 93.04 $\pm 0.09$\\
\langspecadafrozen{} &  & & & 88.79 $\pm 0.13$  &  & & & 88.79 $\pm 0.13$\\
\jointinc{} & \underline{\textbf{-0.01 $\pm 0.01$}} & 0.15 $\pm 0.02$ & \textbf{50.32 $\pm 0.03$} & \underline{93.91 $\pm 0.01$} & \underline{0.12 $\pm 0.01$} & \textbf{0.63 $\pm 0.01$} & \underline{45.87 $\pm 0.03$} & \underline{\textbf{94.30 $\pm 0.01$}} \\
\multi{} &  & & & 94.25 $\pm 0.07$  &  & & & 94.25 $\pm 0.07$\\
\rowcolor{lightgray} \multicolumn{9}{c}{Model Expansion Baselines}\\ \midrule
\textbf{\spectrans{}} & \textbf{0.40 $\pm 0.01$} & \textbf{0.59 $\pm 0.02$} & \textbf{-0.48 $\pm 0.00$} & \textbf{93.86 $\pm 0.01$} & 0.62 $\pm 0.02$ & 0.03 $\pm 0.01$ & -0.54 $\pm 0.00$ & 93.37 $\pm 0.01$ \\
\textbf{\specenca{}} & \textbf{0.60 $\pm 0.01$} & \textbf{1.00 $\pm 0.01$} & 22.02 $\pm 0.02$ & \textbf{93.75 $\pm 0.02$} & 1.05 $\pm 0.02$ & 0.63 $\pm 0.01$ & \textbf{22.50 $\pm 0.01$} & 93.15 $\pm 0.01$ \\
\textbf{\spechead{}} & \textbf{1.53 $\pm 0.02$} & \textbf{0.84 $\pm 0.01$} & \textbf{0.17 $\pm 0.00$} & \textbf{91.93 $\pm 0.01$} & 5.53 $\pm 0.04$ & 0.38 $\pm 0.02$ & -0.11 $\pm 0.00$ & 87.68 $\pm 0.02$ \\
\textbf{\adatuned{}} & \textbf{1.18 $\pm 0.01$} & \textbf{1.29 $\pm 0.01$} & \textbf{50.25 $\pm 0.03$} & \textbf{92.36 $\pm 0.02$} & 4.43 $\pm 0.04$ & 0.79 $\pm 0.02$ & 42.35 $\pm 0.02$ & 88.66 $\pm 0.02$ \\
\textbf{\adafrozen{}} &\textbf{0.84 $\pm 0.02$} & \underline{\textbf{3.41 $\pm 0.02$}} & 3.80 $\pm 0.00$ &\textbf{91.08 $\pm 0.02$} & 1.87 $\pm 0.05$ & \underline{2.43 $\pm 0.02$} & \textbf{9.68 $\pm 0.01$} & 89.92 $\pm 0.02$ \\
\midrule \rowcolor{lightgray}  \multicolumn{9}{c}{Other Continuous Learning Algorithms }\\ \midrule
\textbf{\ewc{}} & \textbf{1.82 $\pm 0.02$} & \textbf{0.74 $\pm 0.01$} & \textbf{51.13 $\pm 0.03$} & \textbf{91.16 $\pm 0.02$} & 5.9 $\pm 0.04$ & 0.48 $\pm 0.02$ & 44.73 $\pm 0.03$ & 88.28 $\pm 0.02$ \\ \midrule 
\textbf{\er{}} & \textbf{0.71 $\pm 0.01$} & \textbf{0.95 $\pm 0.02$} & \textbf{49.59 $\pm 0.03$} & \textbf{93.51 $\pm 0.01$} & 2.35 $\pm 0.03$ & 0.78 $\pm 0.01$ & 44.87 $\pm 0.03$ & 92.58 $\pm 0.02$ \\ \midrule 
\textbf{\kdlogit{}} & \textbf{1.42 $\pm 0.01$} & \textbf{0.77 $\pm 0.02$} & \textbf{50.79 $\pm 0.03$} & \textbf{91.60 $\pm 0.02$} & 4.07 $\pm 0.04$ & 0.51 $\pm 0.01$ & 44.38 $\pm 0.03$ & 89.65 $\pm 0.02$ \\
\textbf{\kdrep{}} & \textbf{1.49 $\pm 0.01$} & \textbf{0.96 $\pm 0.01$} & \underline{\textbf{51.17 $\pm 0.03$}} & \textbf{91.64 $\pm 0.02$} & 4.00 $\pm 0.04$ & 0.53 $\pm 0.01$ & 45.11 $\pm 0.02$ & 90.17 $\pm 0.02$ \\ \midrule
& \multicolumn{8}{c}{\textbf{Test Slot Filling On}}\\\rowcolor{lightgray} \multicolumn{9}{c}{Shared \{Trans, Task\} Baselines } \\\midrule
\naiveft{} & \textbf{4.15 $\pm 0.18$} & \textbf{0.77 $\pm 0.20$} & \textbf{37.03 $\pm 0.05$} & 67.80 $\pm 0.13$ & 7.06 $\pm 0.23$ & 0.77 $\pm 0.17$ & \underline{33.29 $\pm 0.03$} & \textbf{68.37 $\pm 0.13$} \\
\langspec{} &  & & & 73.90 $\pm 0.83$ &  & & & 73.90 $\pm 0.83$ \\
\langspecadatuned{} &  & & & 72.90 $\pm 0.80$  &  & & & 72.90 $\pm 0.80$ \\
\langspecadafrozen{} &  & & & 67.46 $\pm 0.89$  &  & & & 67.46 $\pm 0.89$ \\
\jointinc{} & \underline{0.78 $\pm 0.11$} & \textbf{0.69 $\pm 0.16$} & \underline{\textbf{37.92 $\pm 0.05$}} & \underline{\textbf{75.14 $\pm 0.13$}} & \underline{\textbf{0.37 $\pm 0.14$}} & -0.47 $\pm 0.19$ & 32.75 $\pm 0.03$ & 75.14 $\pm 0.14$ \\
\multi{} &  & & & 76.34 $\pm 0.82$  &  & & & \underline{76.34 $\pm 0.82$} \\
\rowcolor{lightgray} \multicolumn{9}{c}{Model Expansion Baselines}\\ \midrule
\textbf{\spectrans{}} & \textbf{0.99 $\pm 0.11$} & \textbf{0.92 $\pm 0.18$} & 0.33 $\pm 0.00$ & \textbf{74.88 $\pm 0.13$} & 1.23 $\pm 0.14$ & 0.89 $\pm 0.17$ & \textbf{0.39 $\pm 0.00$} & 74.85 $\pm 0.14$ \\
\textbf{\specenca{}} & 2.35 $\pm 0.15$ & \textbf{1.79 $\pm 0.18$} & 10.57 $\pm 0.01$ & 72.51 $\pm 0.13$ & \textbf{2.03 $\pm 0.15$} & 0.74 $\pm 0.19$ & \textbf{12.63 $\pm 0.01$} & \textbf{74.01 $\pm 0.14$} \\
\textbf{\spechead{}} & \textbf{4.08 $\pm 0.17$} & \textbf{1.91 $\pm 0.16$} & \textbf{0.06 $\pm 0.00$} & \textbf{68.88 $\pm 0.15$} & 7.23 $\pm 0.24$ & -0.67 $\pm 0.19$ & \textbf{0.06 $\pm 0.00$} & 66.28 $\pm 0.13$ \\
\textbf{\adatuned{}} & \textbf{2.46 $\pm 0.14$} &\textbf{2.75 $\pm 0.16$} & \textbf{35.05 $\pm 0.05$} & \textbf{71.79 $\pm 0.15$} & 6.42 $\pm 0.23$ & 0.40 $\pm 0.17$ & 29.89 $\pm 0.03$ & 67.70 $\pm 0.12$ \\
\textbf{\adafrozen{}} & \textbf{2.57 $\pm 0.20$} & \underline{\textbf{4.77 $\pm 0.17$}} & \textbf{3.34 $\pm 0.00$} & \textbf{70.33 $\pm 0.15$} & 5.01 $\pm 0.24$ & \underline{2.70 $\pm 0.20$} & 2.59 $\pm 0.00$ & 67.07 $\pm 0.12$ \\
\midrule \rowcolor{lightgray}  \multicolumn{9}{c}{Other Continuous Learning Algorithms }\\ \midrule
\textbf{\ewc{}} & \textbf{4.22 $\pm 0.20$} & \textbf{1.19 $\pm 0.17$} & \textbf{37.39 $\pm 0.05$} & \textbf{68.33 $\pm 0.13$} &7.53 $\pm 0.25$ & 0.52 $\pm 0.16$ & 33.25 $\pm 0.03$ & 66.91 $\pm 0.14$ \\ \midrule 
\textbf{\er{}} & \textbf{2.32 $\pm 0.15$} & \textbf{1.83 $\pm 0.16$} & \textbf{37.50 $\pm 0.05$} & \textbf{73.31 $\pm 0.14$} &3.48 $\pm 0.20$ & 0.44 $\pm 0.19$ & 32.97 $\pm 0.04$ & 72.00 $\pm 0.15$ \\ \midrule 
\textbf{\kdlogit{}} & \textbf{4.42 $\pm 0.18$} & \textbf{1.79 $\pm 0.15$} & \textbf{37.50 $\pm 0.05$} & \textbf{68.13 $\pm 0.14$} &7.36 $\pm 0.27$ & 0.13 $\pm 0.19$ & 32.86 $\pm 0.04$ & 67.13 $\pm 0.14$ \\
\textbf{\kdrep{}} & \textbf{4.56 $\pm 0.18$} & \textbf{1.61 $\pm 0.15$} & \textbf{37.42 $\pm 0.05$} & 68.28 $\pm 0.13$ & 6.65 $\pm 0.28$ & 1.03 $\pm 0.17$ & 32.57 $\pm 0.03$ & \textbf{69.03 $\pm 0.13$} \\

\bottomrule
\end{tabular}}
    \caption{\label{tab:full-cll-metrics-perm} Per language permutation view: a pairwise comparison between \htol{} (English $\rightarrow$ German $\rightarrow$ French $\rightarrow$ Hindi $\rightarrow$ Spanish $\rightarrow$ Thai) and \ltoh{} (Thai $\rightarrow$ Spanish $\rightarrow$ Hindi $\rightarrow$ French $\rightarrow$ German $\rightarrow$ English). We highlight the best forgetting (lowest), transfer (highest), zero-shot transfer (highest), and final performance (highest) of accuracy and f1 scores among those two orders for each approach in \textbf{bold}, whereas the best scores across approaches for the two orders separately are \underline{underlined}.}
\end{table*}

\begin{table*}[ht!] 
\small
\centering
\scalebox{0.82}{
\begin{tabular}{l|llll|llll} \toprule  
\multirow{3}{*}{\textbf{Model}}  & \multicolumn{4}{c|}{Spanish $\rightarrow$ Hindi $\rightarrow$ English $\rightarrow$ German $\rightarrow$ Thai $\rightarrow$ French} & \multicolumn{4}{c}{French $\rightarrow$ Thai $\rightarrow$ German $\rightarrow$ English $\rightarrow$ Hindi $\rightarrow$ Spanish} \\ 
& \multicolumn{8}{c}{\textbf{Test Intent Accuracy On}} \\
& F $\downarrow$ &  T $\uparrow$ & $T^0$ $\uparrow$ & FP $\uparrow$  &  F $\downarrow$ & T $\uparrow$ & $T^0$ $\uparrow$  & FP $\uparrow$  \\ 
\midrule \rowcolor{lightgray} 

\rowcolor{lightgray} \multicolumn{9}{c}{Shared \{Trans, Task\} Baselines }   \\ \midrule 
\naiveft{} & \textbf{2.62 $\pm 0.03$} & \textbf{0.59 $\pm 0.01$} & 52.07 $\pm 0.03$ & \textbf{91.49 $\pm 0.02$} & 2.63 $\pm 0.03$ & 0.52 $\pm 0.02$ & \textbf{55.0 $\pm 0.02$} & 90.74 $\pm 0.02$ \\
\langspec{} &  & & & 93.40 $\pm 0.08$  &  & & & 93.40 $\pm 0.08$\\
\langspecadatuned{} &  & & & 93.04 $\pm 0.09$  &  & & & 93.04 $\pm 0.09$\\
\langspecadafrozen{} &  & & & 88.79 $\pm 0.13$  &  & & & 88.79 $\pm 0.13$\\
\jointinc{} & \underline{\textbf{0.11 $\pm 0.01$}} & 0.47 $\pm 0.01$ & \textbf{53.86 $\pm 0.02$} & 94.01 $\pm 0.01$ & \underline{0.25 $\pm 0.01$} & \textbf{0.61 $\pm 0.01$} & 50.51 $\pm 0.02$ & \textbf{94.09 $\pm 0.01$} \\
\multi{} &  & & & \underline{94.25 $\pm 0.07$}  &  & & & \underline{94.25 $\pm 0.07$} \\
\rowcolor{lightgray} \multicolumn{9}{c}{Model Expansion Baselines}\\ \midrule
\textbf{\spectrans{}} & \textbf{0.45 $\pm 0.01$} & 0.05 $\pm 0.02$ & \textbf{-0.37 $\pm 0.00$} & 93.43 $\pm 0.01$ & 0.51 $\pm 0.01$ & \textbf{0.39 $\pm 0.02$} & -0.5 $\pm 0.0$ & \textbf{93.63 $\pm 0.01$} \\
\textbf{\specenca{}} & \textbf{0.64 $\pm 0.02$} & 0.54 $\pm 0.01$ & \textbf{26.32 $\pm 0.02$} & \textbf{93.68 $\pm 0.01$} & \textbf{0.81 $\pm 0.02$} & 0.82 $\pm 0.02$ & 25.26 $\pm 0.02$ & 93.59 $\pm 0.01$ \\
\textbf{\spechead{}} & \textbf{2.23 $\pm 0.03$} & 0.46 $\pm 0.02$ & \textbf{0.47 $\pm 0.00$} & \textbf{91.73 $\pm 0.02$} & 3.02 $\pm 0.03$ & \textbf{0.85 $\pm 0.02$} & -0.07 $\pm 0.0$ & 90.91 $\pm 0.02$ \\
\textbf{\adatuned{}} & \textbf{1.36 $\pm 0.02$} & \textbf{1.07 $\pm 0.01$} & 50.06 $\pm 0.02$ & \textbf{92.70 $\pm 0.02$} & 2.33 $\pm 0.03$ & 0.78 $\pm 0.02$ & \textbf{51.96 $\pm 0.02$} & 92.15 $\pm 0.02$ \\
\textbf{\adafrozen{}} & \textbf{0.82 $\pm 0.02$} & \underline{2.61 $\pm 0.02$} & 5.68 $\pm 0.01$ & \textbf{90.34 $\pm 0.02$} & 1.21 $\pm 0.03$ & \underline{\textbf{2.75 $\pm 0.02$}} & \textbf{8.84 $\pm 0.01$} & 90.17 $\pm 0.02$ \\
\midrule \rowcolor{lightgray}  \multicolumn{9}{c}{Other Continuous Learning Algorithms }\\ \midrule
\textbf{\ewc{}} & \textbf{2.55 $\pm 0.02$} & \textbf{0.87 $\pm 0.01$} & 52.29 $\pm 0.03$ & \textbf{92.04 $\pm 0.02$} & 2.57 $\pm 0.03$ & 0.71 $\pm 0.02$ & \textbf{54.84 $\pm 0.02$} & 91.67 $\pm 0.02$ \\ \midrule 
\textbf{\er{}} & \textbf{1.27 $\pm 0.02$} & \textbf{0.70 $\pm 0.02$} & \underline{54.29 $\pm 0.02$} & \textbf{93.08 $\pm 0.01$} & 1.33 $\pm 0.02$ & 0.44 $\pm 0.02$ & \underline{\textbf{55.05 $\pm 0.03$}} & 93.05 $\pm 0.02$ \\ \midrule 
\textbf{\kdlogit{}} & \textbf{2.16 $\pm 0.02$} & \textbf{0.54 $\pm 0.02$} & 52.32 $\pm 0.02$ & \textbf{92.23 $\pm 0.02$} & 2.18 $\pm 0.03$ & 0.45 $\pm 0.02$ & \textbf{53.73 $\pm 0.03$} & 91.84 $\pm 0.02$ \\
\textbf{\kdrep{}} & \textbf{2.04 $\pm 0.03$} & 0.36 $\pm 0.02$ & 52.06 $\pm 0.03$ & \textbf{92.25 $\pm 0.02$} & 2.13 $\pm 0.03$ & \textbf{0.65 $\pm 0.01$} & \textbf{53.55 $\pm 0.03$} & 92.06 $\pm 0.02$ \\ \midrule 
& \multicolumn{8}{c}{\textbf{Test Slot Filling On}}\\\rowcolor{lightgray} \multicolumn{9}{c}{Shared \{Trans, Task\} Baselines } \\\midrule
\naiveft{} & \textbf{5.40 $\pm 0.25$} & \textbf{1.95 $\pm 0.17$} & 36.2 $\pm 0.04$ & \textbf{70.61 $\pm 0.14$} & 5.5 $\pm 0.19$ & 1.81 $\pm 0.16$ & \textbf{38.41 $\pm 0.04$} & 70.30 $\pm 0.15$ \\
\langspec{} &  & & & 73.90 $\pm 0.83$  &  & & & 73.90 $\pm 0.83$ \\
\langspecadatuned{} &  & & & 72.90 $\pm 0.80$  &  & & & 72.90 $\pm 0.80$ \\
\langspecadafrozen{} &  & & & 67.46 $\pm 0.89$  &  & & & 67.46 $\pm 0.89$ \\
\jointinc{} & \underline{\textbf{0.81 $\pm 0.14$}} & 1.57 $\pm 0.16$ & 37.46 $\pm 0.05$ & 74.9 $\pm 0.16$ & \underline{1.03 $\pm 0.15$} & \textbf{1.72 $\pm 0.17$} & \textbf{37.54 $\pm 0.04$} & \textbf{75.34 $\pm 0.15$} \\
\multi{} &  & & & \underline{76.34 $\pm 0.82$}  &  & & & \underline{76.34 $\pm 0.82$} \\
\rowcolor{lightgray} \multicolumn{9}{c}{Model Expansion Baselines}\\ \midrule
\textbf{\spectrans{}} & 1.57 $\pm 0.18$ & \textbf{1.29 $\pm 0.15$} & \textbf{0.49 $\pm 0.00$} & 74.56 $\pm 0.13$ & \textbf{1.29 $\pm 0.13$} & 0.60 $\pm 0.17$ & 0.47 $\pm 0.0$ & \textbf{74.57 $\pm 0.15$} \\
\textbf{\specenca{}} & 1.80 $\pm 0.19$ & \textbf{2.05 $\pm 0.17$} & 13.24 $\pm 0.01$ & \textbf{75.2 $\pm 0.16$} & \textbf{1.25 $\pm 0.17$} & 0.23 $\pm 0.17$ & \textbf{13.57 $\pm 0.01$} & 74.67 $\pm 0.14$ \\
\textbf{\spechead{}} & 4.94 $\pm 0.24$ & \textbf{2.20 $\pm 0.16$} & \textbf{0.11 $\pm 0.00$} & \textbf{71.06 $\pm 0.14$} & \textbf{4.77 $\pm 0.22$} & 1.14 $\pm 0.18$ & 0.05 $\pm 0.0$ & 70.63 $\pm 0.14$ \\
\textbf{\adatuned{}} & \textbf{3.25 $\pm 0.18$} & \textbf{3.26 $\pm 0.16$} & 34.88 $\pm 0.04$ & \textbf{72.38 $\pm 0.16$} & 4.31 $\pm 0.21$ & 1.75 $\pm 0.14$ & \textbf{35.48 $\pm 0.03$} & 70.39 $\pm 0.13$ \\
\textbf{\adafrozen{}} & \textbf{2.52 $\pm 0.2$} & \underline{\textbf{4.03 $\pm 0.18$}} & 3.10 $\pm 0.0$ & 68.22 $\pm 0.14$ & 3.06 $\pm 0.2$ & \underline{\textbf{4.03 $\pm 0.19$}} & \textbf{3.57 $\pm 0.0$} & \textbf{68.67 $\pm 0.14$} \\
\midrule \rowcolor{lightgray}  \multicolumn{9}{c}{Other Continuous Learning Algorithms }\\ \midrule
\textbf{\ewc{}} & 5.54 $\pm 0.24$ & \textbf{1.99 $\pm 0.16$} & 36.34 $\pm 0.04$ & \textbf{70.69 $\pm 0.13$} & \textbf{5.46 $\pm 0.23$} & 1.07 $\pm 0.18$ & \textbf{38.14 $\pm 0.04$} & 70.05 $\pm 0.15$ \\ \midrule 
\textbf{\er{}} & 3.01 $\pm 0.18$ & \textbf{1.98 $\pm 0.16$} & \underline{37.54 $\pm 0.04$} & \textbf{72.92 $\pm 0.13$} &\textbf{2.77 $\pm 0.18$} & 0.81 $\pm 0.17$ & \underline{\textbf{38.66 $\pm 0.04$}} & 72.82 $\pm 0.14$ \\ \midrule 
\textbf{\kdlogit{}} & \textbf{5.00 $\pm 0.25$} & \textbf{2.00 $\pm 0.17$} & 35.67 $\pm 0.04$ & \textbf{71.82 $\pm 0.14$} &5.46 $\pm 0.23$ & 1.52 $\pm 0.17$ & \textbf{37.78 $\pm 0.04$} & 69.76 $\pm 0.14$ \\
\textbf{\kdrep{}} & \textbf{4.84 $\pm 0.22$} & \textbf{1.46 $\pm 0.17$} & 35.96 $\pm 0.04$ & \textbf{70.71 $\pm 0.16$} & 5.01 $\pm 0.22$ & 0.9 $\pm 0.16$ & \textbf{37.25 $\pm 0.04$} & 70.37 $\pm 0.14$ \\ 

\bottomrule
\end{tabular}}

\caption{\label{tab:full-cll-metrics-perm-1} Per language permutation view: a pairwise comparison between Order 3 (Spanish $\rightarrow$ Hindi $\rightarrow$ English $\rightarrow$ German $\rightarrow$ Thai $\rightarrow$ French) and Order 4 (French $\rightarrow$ Thai $\rightarrow$ German $\rightarrow$ English $\rightarrow$ Hindi $\rightarrow$ Spanish).  We highlight the best forgetting (lowest), transfer (highest), zero-shot transfer (highest), and final performance (highest) of accuracy and f1 scores among those two orders for each approach in \textbf{bold},  whereas the best scores across approaches for the two orders separately are \underline{underlined}.}
\end{table*}


\begin{table*}[ht!] 
\small
\centering
\scalebox{0.81}{
\begin{tabular}{l|llll|llll} \toprule  
\multirow{3}{*}{\textbf{Model}}  &
\multicolumn{4}{c|}{Hindi $\rightarrow$ English $\rightarrow$ Spanish $\rightarrow$ Thai $\rightarrow$ French $\rightarrow$ German} & \multicolumn{4}{c}{German $\rightarrow$ French $\rightarrow$ Thai $\rightarrow$ Spanish $\rightarrow$ English $\rightarrow$ Hindi}  \\
&  \multicolumn{8}{c}{\textbf{Test Intent Accuracy On}} \\ 
& F $\downarrow$ &  T $\uparrow$ & $T^0$ $\uparrow$ &  FP $\uparrow$  &  F $\downarrow$ & T $\uparrow$ & $T^0$ $\uparrow$  & FP $\uparrow$  \\ 
\midrule \rowcolor{lightgray} 

\rowcolor{lightgray} \multicolumn{9}{c}{Shared \{Trans, Task\} Baselines }   \\ \midrule 
 
\naiveft{} & 2.97 $\pm 0.03$ & \textbf{0.75 $\pm 0.01$} & 47.04 $\pm 0.03$ & \textbf{91.63 $\pm 0.02$} & \textbf{2.32 $\pm 0.02$} & 0.71 $\pm 0.02$ & \textbf{51.97 $\pm 0.03$} & \textbf{91.63 $\pm 0.02$} \\
\langspec{} &  & & & 93.4 $\pm 0.08$  &  & & & 93.4 $\pm 0.08$\\
\langspecadatuned{} &  & & & 93.04 $\pm 0.09$  &  & & & 93.04 $\pm 0.09$\\
\langspecadafrozen{} &  & & & 88.79 $\pm 0.13$  &  & & & 88.79 $\pm 0.13$\\
\jointinc{} & \underline{0.21 $\pm 0.01$} & \textbf{0.74 $\pm 0.01$} & \underline{48.41 $\pm 0.03$} & \underline{\textbf{94.44 $\pm 0.01$}} & \underline{\textbf{-0.02 $\pm 0.01$}} & 0.54 $\pm 0.02$ & \textbf{51.49 $\pm 0.03$} & 94.23 $\pm 0.01$ \\
\multi{} &  & & & 94.25 $\pm 0.07$  &  & & & \underline{94.25 $\pm 0.07$}\\
\rowcolor{lightgray} \multicolumn{9}{c}{Model Expansion Baselines}\\ \midrule
\textbf{\spectrans{}} & \textbf{0.41 $\pm 0.02$} & 0.03 $\pm 0.02$ & -0.57 $\pm 0.0$ & \textbf{93.39 $\pm 0.01$} & 0.52 $\pm 0.02$ & \textbf{0.29 $\pm 0.02$} & \textbf{-0.11 $\pm 0.0$} & 93.38 $\pm 0.01$ \\
\textbf{\specenca{}} & 0.80 $\pm 0.02$ & 0.74 $\pm 0.01$ & 23.18 $\pm 0.02$ & 93.35 $\pm 0.01$ & \textbf{0.76 $\pm 0.01$} & \textbf{1.05 $\pm 0.01$} & \textbf{26.12 $\pm 0.02$} & \textbf{93.46 $\pm 0.01$} \\
\textbf{\spechead{}} & 2.84 $\pm 0.03$ & 0.67 $\pm 0.01$ & -0.2 $\pm 0.0$ & 91.17 $\pm 0.02$ & \textbf{2.32 $\pm 0.02$} & \textbf{0.76 $\pm 0.01$} & \textbf{0.36 $\pm 0.0$} & \textbf{91.7 $\pm 0.02$} \\
\textbf{\adatuned{}} & 2.49 $\pm 0.03$ & \textbf{1.05 $\pm 0.01$} & 47.67 $\pm 0.03$ & \textbf{92.34 $\pm 0.01$} & \textbf{1.35 $\pm 0.02$} & 0.89 $\pm 0.01$ & \underline{\textbf{53.77 $\pm 0.02$}} & 92.30 $\pm 0.02$ \\
\textbf{\adafrozen{}} & \textbf{1.13 $\pm 0.03$} & \underline{\textbf{3.09 $\pm 0.02$}} & 4.40 $\pm 0.01$ & \textbf{90.50 $\pm 0.02$} & 1.32 $\pm 0.02$ & \underline{2.64 $\pm 0.02$} & \textbf{6.73 $\pm 0.01$} & 90.15 $\pm 0.02$ \\
\midrule \rowcolor{lightgray}  \multicolumn{9}{c}{Other Continuous Learning Algorithms }\\ \midrule
\textbf{\ewc{}} & 3.07 $\pm 0.03$ & 0.79 $\pm 0.01$ & 46.44 $\pm 0.03$ & 91.45 $\pm 0.02$ & \textbf{2.51 $\pm 0.02$} & \textbf{0.81 $\pm 0.01$} & \textbf{51.54 $\pm 0.02$} & \textbf{91.54 $\pm 0.02$} \\ \midrule 
\textbf{\er{}} & 1.11 $\pm 0.02$ & 0.72 $\pm 0.01$ & 48.23 $\pm 0.03$ & 93.00 $\pm 0.02$ & \textbf{0.98 $\pm 0.02$} & \textbf{0.92 $\pm 0.01$} & \textbf{52.23 $\pm 0.03$} & \textbf{93.32 $\pm 0.01$} \\ \midrule 
\textbf{\kdlogit{}} & 2.50 $\pm 0.03$ & \textbf{0.86 $\pm 0.01$} & 47.96 $\pm 0.03$ & 91.27 $\pm 0.02$ & \textbf{1.89 $\pm 0.02$} & 0.59 $\pm 0.02$ & \textbf{51.88 $\pm 0.03$} & \textbf{92.16 $\pm 0.02$} \\
\textbf{\kdrep{}} & 2.24 $\pm 0.03$ & 0.81 $\pm 0.01$ & 48.08 $\pm 0.03$ & 91.89 $\pm 0.02$ & \textbf{1.86 $\pm 0.02$} & \textbf{0.83 $\pm 0.02$} & \textbf{52.51 $\pm 0.03$} & \textbf{92.16 $\pm 0.02$} \\ \midrule 

& \multicolumn{8}{c}{\textbf{Test Slot Filling On}}\\\rowcolor{lightgray} \multicolumn{9}{c}{Shared \{Trans, Task\} Baselines } \\\midrule
\naiveft{} & 6.51 $\pm 0.22$ & \textbf{1.90 $\pm 0.15$} & 34.53 $\pm 0.04$ & 68.93 $\pm 0.13$ & \textbf{5.38 $\pm 0.25$} & 1.00 $\pm 0.18$ & \textbf{38.47 $\pm 0.05$} & \textbf{70.22 $\pm 0.14$} \\
\langspec{} &  & & & 73.9 $\pm 0.83$  &  & & & 73.9 $\pm 0.83$ \\
\langspecadatuned{} &  & & & 72.9 $\pm 0.8$  &  & & & 72.9 $\pm 0.8$ \\
\langspecadafrozen{} &  & & & 67.46 $\pm 0.89$  &  & & & 67.46 $\pm 0.89$ \\
\jointinc{} & \underline{\textbf{0.99 $\pm 0.15$}} & \textbf{1.21 $\pm 0.15$} & 32.99 $\pm 0.03$ & \textbf{74.45 $\pm 0.16$} & 1.52 $\pm 0.15$ & 0.27 $\pm 0.18$ & \underline{\textbf{39.69 $\pm 0.05$}} & 74.31 $\pm 0.14$ \\
\multi{} &  & & & \underline{76.34 $\pm 0.82$}  &  & & & \underline{76.34 $\pm 0.82$} \\
\rowcolor{lightgray} \multicolumn{9}{c}{Model Expansion Baselines}\\ \midrule
\textbf{\spectrans{}} & 1.65 $\pm 0.17$ & \textbf{1.04 $\pm 0.17$} & 0.37 $\pm 0.00$ & 74.51 $\pm 0.14$ & \underline{\textbf{1.17 $\pm 0.14$}} & 0.97 $\pm 0.16$ & \textbf{0.47 $\pm 0.00$} & \textbf{75.04 $\pm 0.14$} \\
\textbf{\specenca{}} & \textbf{1.48 $\pm 0.13$} & \textbf{2.18 $\pm 0.17$} & 10.66 $\pm 0.01$ & \textbf{75.03 $\pm 0.14$} & 2.77 $\pm 0.18$ & 1.67 $\pm 0.18$ & \textbf{13.25 $\pm 0.01$} & 73.73 $\pm 0.14$ \\
\textbf{\spechead{}} & 5.72 $\pm 0.21$ & \textbf{2.4 $\pm 0.17$} & \textbf{0.06 $\pm 0.00$} & \textbf{70.08 $\pm 0.13$} & \textbf{4.80 $\pm 0.24$} & -0.04 $\pm 0.18$ & \textbf{0.06 $\pm 0.00$} & 69.54 $\pm 0.13$ \\
\textbf{\adatuned{}} & 4.96 $\pm 0.25$ & \textbf{2.39 $\pm 0.15$} & 29.17 $\pm 0.03$ & \textbf{72.28 $\pm 0.13$} & \textbf{3.98 $\pm 0.21$} & 1.69 $\pm 0.16$ & \textbf{37.14 $\pm 0.05$} & 72.27 $\pm 0.13$ \\
\textbf{\adafrozen{}} & \textbf{3.15 $\pm 0.21$} & \underline{\textbf{4.51 $\pm 0.18$}} & 1.90 $\pm 0.00$ & \textbf{69.47 $\pm 0.14$} & 3.77 $\pm 0.22$ & \underline{3.54 $\pm 0.16$} & \textbf{2.31 $\pm 0.00$} & 67.57 $\pm 0.14$ \\
\midrule \rowcolor{lightgray}  \multicolumn{9}{c}{Other Continuous Learning Algorithms }\\ \midrule
\textbf{\ewc{}} & 6.38 $\pm 0.23$ & \textbf{2.54 $\pm 0.17$} & 34.29 $\pm 0.04$ & \textbf{71.25 $\pm 0.14$} & \textbf{5.56 $\pm 0.27$} & 1.46 $\pm 0.17$ & \textbf{38.44 $\pm 0.05$} & 70.57 $\pm 0.16$ \\ \midrule 
\textbf{\er{}} & 4.12 $\pm 0.22$ & \textbf{2.90 $\pm 0.16$} & 35.45 $\pm 0.04$ & 73.39 $\pm 0.14$ &\textbf{2.65 $\pm 0.18$} & 0.83 $\pm 0.17$ & \textbf{39.34 $\pm 0.05$} & \textbf{73.56 $\pm 0.15$} \\ \midrule 
\textbf{\kdlogit{}} & 6.03 $\pm 0.27$ & \textbf{2.02 $\pm 0.16$} & 35.2 $\pm 0.04$ & \textbf{70.70 $\pm 0.14$} & \textbf{4.91 $\pm 0.21$} & 0.92 $\pm 0.17$ & \textbf{38.49 $\pm 0.05$} & 70.31 $\pm 0.14$ \\
\textbf{\kdrep{}} & 5.72 $\pm 0.27$ & \textbf{2.6 $\pm 0.15$} & \underline{35.54 $\pm 0.04$} & \textbf{71.61 $\pm 0.15$} & \textbf{5.35 $\pm 0.21$} & 0.97 $\pm 0.15$ & \textbf{38.8 $\pm 0.05$} & 70.15 $\pm 0.13$ \\ 
\bottomrule
\end{tabular}}

\caption{\label{tab:full-cll-metrics-perm-2} Per language permutation view: a pairwise comparison between Order 5(Hindi $\rightarrow$ English $\rightarrow$ Spanish $\rightarrow$ Thai $\rightarrow$ French $\rightarrow$ German) and Order 6 (German $\rightarrow$ French $\rightarrow$ Thai $\rightarrow$ Spanish $\rightarrow$ English $\rightarrow$ Hindi). We highlight the best forgetting (lowest), transfer (highest), zero-shot transfer (highest), and final performance (highest) of accuracy and f1 scores among those two orders for each approach in \textbf{bold},  whereas the best scores across approaches for the two orders separately are \underline{underlined}.}
\end{table*}

\begin{table*}[ht!] 
\small
\centering
\scalebox{1.0}{
\begin{tabular}{l|ll|ll|ll} \toprule  
\multirow{2}{*}{\textbf{Model}}  
& \multicolumn{2}{c|}{F $\downarrow$} &  \multicolumn{2}{c}{T $\uparrow$} & \multicolumn{2}{c}{FP $\uparrow$}  
\\ & Acc & F1 & Acc & F1 & Acc & F1 
\\
\midrule
Order 1 & \underline{1.25 $\pm 0.02$} & \textbf{3.60 $\pm 0.18$} & 0.89 $\pm 0.02$ & 1.76 $\pm 0.17$  & 89.33 $\pm 0.02$ & 65.59 $\pm 0.13$ \\
Order 2 & 5.81 $\pm 0.05$ & 7.89 $\pm 0.28$ & 0.75 $\pm 0.02$ & 0.11 $\pm 0.17$  & 85.81 $\pm 0.02$ & 64.18 $\pm 0.14$ \\
Order 3 & 1.68 $\pm 0.02$ & \underline{4.43 $\pm 0.21$} & 0.77 $\pm 0.02$ & 2.20 $\pm 0.17$  & 89.57 $\pm 0.02$ & 68.88 $\pm 0.14$ \\
Order 4 & 2.70 $\pm 0.04$ & 4.62 $\pm 0.23$ & 0.71 $\pm 0.02$ & 1.22 $\pm 0.17$  & 88.59 $\pm 0.02$ & 68.07 $\pm 0.14$ \\
Order 5 & 1.83 $\pm 0.01$ & 5.74 $\pm 0.24$ & \underline{6.64 $\pm 0.01$} & \textbf{4.89 $\pm 0.15$} & \underline{96.00 $\pm 0.01$} & \underline{71.75 $\pm 0.13$} \\
Order 6 & \textbf{1.08 $\pm 0.01$} & 4.44 $\pm 0.20$ & \textbf{7.09 $\pm 0.01$} & \underline{4.86 $\pm 0.15$} & \textbf{96.40 $\pm 0.01$} & \textbf{71.81 $\pm 0.13$} \\
\bottomrule
\end{tabular}
} \caption{\label{tab:full-cll-equal} Impact of language order across the balanced dataset for \naiveft{}. The best and second best scores for each language for intent classification and slot filling independently across approaches are highlighted in \textbf{bold} and \underline{underlined}, respectively.}
\end{table*}

\subsection{Per Language Analysis}
\label{app:per-lang-analysis}

Tables~\ref{tab:full-cll-forgetting-lang},~\ref{tab:full-cll-fwt-lang}, and~\ref{tab:full-cll-fwt-0-lang} show the full results for forgetting, transfer, and zero-shot transfer respectively, across different languages averaged over different language permutations. We notice that languages like English, German, French, and Spanish have constantly lower forgetting and higher zero-shot transfer than languages like Hindi and Thai for both intent classification and slot filling for \naiveft{} compared to the reference model \jointinc{} for which the forgetting is low and nearly equal between different languages. Approaches like \spectrans{}, \specenca{}, \adafrozen{}, and to a certain degree \er{} also reduce that gap. We also notice that approaches that lower forgetting for a particular language do so uniformly for all languages. The performance in terms of zero-shot transfer is significantly lower in the case of Thai.

\begin{table*}[ht!] 
\small
\centering
\scalebox{1.05}{
\begin{tabular}{l|llllll} \toprule  
\multirow{3}{*}{\textbf{Model}}  & \multicolumn{6}{c}{\textbf{Test Intent Accuracy On}} \\
&  German &  English &  French &  Spanish &  Hindi &  Thai  \\ 
\midrule \rowcolor{lightgray} \multicolumn{7}{c}{Shared \{Trans, Task\} Baselines}   \\ \midrule 
\naiveft{} & 1.52 $\pm 0.12$ & 1.06 $\pm 0.08$ & 1.30 $\pm 0.14$ & 1.49 $\pm 0.13$ & 2.90 $\pm 0.38$ & 5.51 $\pm 1.35$ \\
\jointinc{} & \textbf{0.31 $\pm 0.05$} & \textbf{0.12 $\pm 0.04$} & \textbf{0.19 $\pm 0.05$} & \textbf{0.15 $\pm 0.04$} & \textbf{0.04 $\pm 0.07$} & \textbf{0.28 $\pm 0.08$} \\\midrule
\rowcolor{lightgray} \multicolumn{7}{c}{Model Expansion Baselines}\\ \midrule
\textbf{\spectrans{}} & \underline{0.36 $\pm 0.06$} & \underline{0.33 $\pm 0.04$} & \underline{0.44 $\pm 0.07$} &\underline{0.34 $\pm 0.06$} & \underline{0.42 $\pm 0.08$} & \underline{0.46 $\pm 0.08$} \\
\textbf{\specenca{}} & 0.54 $\pm 0.07$ & 0.45 $\pm 0.05$ & 0.51 $\pm 0.08$ & 0.59 $\pm 0.06$ & 0.66 $\pm 0.10$ & 0.90 $\pm 0.15$ \\
\textbf{\spechead{}} & 1.22 $\pm 0.12$ & 0.95 $\pm 0.09$ & 1.49 $\pm 0.14$ & 1.37 $\pm 0.12$ & 3.20 $\pm 0.40$ & 5.44 $\pm 1.67$ \\
\textbf{\adatuned{}} & 0.88 $\pm 0.08$ & 0.81 $\pm 0.08$ & 1.16 $\pm 0.12$ & 1.00 $\pm 0.09$ & 1.85 $\pm 0.24$ & 4.23 $\pm 1.15$ \\
\textbf{\adafrozen{}} & 0.58 $\pm 0.08$ & 0.61 $\pm 0.08$ & 0.81 $\pm 0.11$ & 0.54 $\pm 0.10$ & 0.86 $\pm 0.11$ & 1.88 $\pm 0.33$ \\\midrule
\rowcolor{lightgray}  \multicolumn{7}{c}{Other Continuous Learning Algorithms }\\ \midrule
\textbf{\ewc{}} & 1.40 $\pm 0.15$ & 1.00 $\pm 0.08$ & 1.74 $\pm 0.15$ & 1.56 $\pm 0.13$ & 3.26 $\pm 0.37$ & 5.62 $\pm 1.75$ \\
\textbf{\er{}} & 0.76 $\pm 0.07$ & 0.53 $\pm 0.05$ & 0.87 $\pm 0.08$ & 0.71 $\pm 0.08$ & 1.13 $\pm 0.12$ & 2.19 $\pm 0.22$ \\
\textbf{\kdlogit{}} & 1.23 $\pm 0.12$ & 0.97 $\pm 0.08$ & 1.47 $\pm 0.12$ & 1.27 $\pm 0.12$ & 2.19 $\pm 0.27$ & 4.41 $\pm 0.75$ \\
\textbf{\kdrep{}} & 1.20 $\pm 0.11$ & 0.80 $\pm 0.07$ & 1.45 $\pm 0.11$ & 1.42 $\pm 0.12$ & 2.29 $\pm 0.27$ & 4.02 $\pm 0.63$ \\\midrule
& \multicolumn{6}{c}{\textbf{Test Slot Filling On}}\\
&  German &  English &  French &  Spanish &  Hindi &  Thai  \\ \rowcolor{lightgray} \multicolumn{7}{c}{Shared \{Trans, Task\} Baselines } \\\midrule
\naiveft{} & 3.64 $\pm 1.31$ & 3.91 $\pm 1.14$ & 2.80 $\pm 0.94$ & 2.94 $\pm 0.94$ & 6.48 $\pm 1.85$ & 8.85 $\pm 3.19$ \\
\jointinc{} & \underline{1.21 $\pm 0.85$} & \underline{1.12 $\pm 0.70$} & \textbf{0.64 $\pm 0.71$} & \textbf{0.96 $\pm 0.62$} & \textbf{1.13 $\pm 0.70$} & \textbf{0.77 $\pm 0.57$} \\\midrule
\rowcolor{lightgray} \multicolumn{7}{c}{Model Expansion Baselines}\\ \midrule
\textbf{\spectrans{}} & \textbf{0.90 $\pm 0.71$} & \textbf{1.02 $\pm 0.62$} & \underline{1.03 $\pm 0.65$} & \underline{1.21 $\pm 0.74$} & \underline{1.28 $\pm 0.75$} & \underline{1.06 $\pm 0.64$} \\
\textbf{\specenca{}} & 2.03 $\pm 0.93$ & 1.83 $\pm 0.81$ & \underline{1.03 $\pm 0.77$} & 1.31 $\pm 0.69$ & 1.76 $\pm 0.81$ & 2.00 $\pm 0.76$ \\
\textbf{\spechead{}} & 3.32 $\pm 1.29$ & 2.96 $\pm 0.97$ & 2.74 $\pm 0.93$ & 2.76 $\pm 0.89$ & 6.89 $\pm 2.01$ & 8.17 $\pm 3.05$ \\
\textbf{\adatuned{}} & 2.96 $\pm 1.12$ & 3.05 $\pm 0.88$ & 1.49 $\pm 0.76$ & 1.52 $\pm 0.82$ & 4.34 $\pm 1.17$ & 6.84 $\pm 2.26$ \\
\textbf{\adafrozen{}} & 1.82 $\pm 0.97$ & 1.85 $\pm 0.88$ & 1.33 $\pm 0.83$ & 1.89 $\pm 0.96$ & 2.72 $\pm 0.99$ & 5.81 $\pm 1.98$ \\\midrule
\rowcolor{lightgray}  \multicolumn{7}{c}{Other Continuous Learning Algorithms }\\ \midrule
\textbf{\ewc{}} & 3.41 $\pm 1.25$ & 3.90 $\pm 1.24$ & 3.08 $\pm 0.95$ & 3.32 $\pm 0.96$ & 6.29 $\pm 1.86$ & 8.74 $\pm 3.22$ \\
\textbf{\er{}} & 1.94 $\pm 0.82$ & 2.01 $\pm 0.96$ & 1.60 $\pm 0.76$ & 1.82 $\pm 0.80$ & 3.65 $\pm 1.04$ & 4.73 $\pm 1.18$ \\
\textbf{\kdlogit{}} & 3.69 $\pm 1.31$ & 3.70 $\pm 1.03$ & 3.10 $\pm 1.01$ & 3.55 $\pm 1.11$ & 5.66 $\pm 1.68$ & 8.05 $\pm 2.68$ \\
\textbf{\kdrep{}} & 3.49 $\pm 1.18$ & 3.85 $\pm 1.09$ & 3.13 $\pm 0.95$ & 2.99 $\pm 0.92$ & 5.81 $\pm 1.66$ & 7.93 $\pm 2.18$ \\
\bottomrule
\end{tabular}
}
\caption{\label{tab:full-cll-forgetting-lang} CCL per language analysis of forgetting. The best and second best scores for each language are highlighted in \textbf{bold} and \underline{underlined}, respectively.}
\end{table*}

\begin{table*}[ht!] 
\small
\centering
\scalebox{1.05}{
\begin{tabular}{l|llllll} \toprule  
\multirow{3}{*}{\textbf{Model}}  & \multicolumn{6}{c}{\textbf{Test Intent Accuracy On}} \\
&  German &  English & French & Hindi &  Spanish &  Thai  \\ 
\midrule \rowcolor{lightgray} \multicolumn{7}{c}{Shared \{Trans, Task\} Baselines}   \\ \midrule 
\naiveft{} & 0.37 $\pm 0.07$ & 0.30 $\pm 0.06$ & 0.77 $\pm 0.08$ & 1.14 $\pm 0.07$ & 0.64 $\pm 0.09$ & 0.85 $\pm 0.11$ \\
\jointinc{} & 0.25 $\pm 0.07$ & 0.04 $\pm 0.06$ & 0.74 $\pm 0.09$ & 1.25 $\pm 0.06$ & 0.27 $\pm 0.12$ & 0.57 $\pm 0.11$ \\\midrule
\rowcolor{lightgray} \multicolumn{7}{c}{Model Expansion Baselines}\\ \midrule
\textbf{\spectrans{}} & -0.36 $\pm 0.08$ & -0.07 $\pm 0.06$ & 0.29 $\pm 0.10$ & 0.93 $\pm 0.08$ & 0.12 $\pm 0.10$ & 0.47 $\pm 0.11$ \\
\textbf{\specenca{}} & 0.39 $\pm 0.07$ & 0.28 $\pm 0.05$ & 0.96 $\pm 0.08$ & 1.09 $\pm 0.07$ & 0.80 $\pm 0.11$ & \underline{1.25 $\pm 0.10$} \\
\textbf{\spechead{}} & 0.22 $\pm 0.07$ & 0.12 $\pm 0.06$ & 0.99 $\pm 0.08$ & 1.11 $\pm 0.07$ & 0.69 $\pm 0.10$ & 0.84 $\pm 0.09$ \\
\textbf{\adatuned{}} & \underline{1.38 $\pm 0.07$} & \underline{0.41 $\pm 0.06$} & \underline{1.30 $\pm 0.09$} & 0.93 $\pm 0.11$ & \underline{1.20 $\pm 0.09$} & 0.65 $\pm 0.10$ \\
\textbf{\adafrozen{}} & \textbf{2.47 $\pm 0.10$} & \textbf{1.43 $\pm 0.08$} & \textbf{3.03 $\pm 0.11$} & \textbf{3.17 $\pm 0.11$} & \textbf{2.00 $\pm 0.15$} & \textbf{4.84 $\pm 0.33$} \\\midrule
\rowcolor{lightgray}  \multicolumn{7}{c}{Other Continuous Learning Algorithms }\\ \midrule
\textbf{\ewc{}} & 0.26 $\pm 0.08$ & 0.12 $\pm 0.05$ & 1.13 $\pm 0.07$ & 1.10 $\pm 0.07$ & 0.85 $\pm 0.09$ & 0.92 $\pm 0.11$ \\
\textbf{\er{}} & 0.27 $\pm 0.08$ & 0.07 $\pm 0.06$ & 1.01 $\pm 0.08$ & 1.16 $\pm 0.07$ & 0.96 $\pm 0.10$ & 1.04 $\pm 0.11$ \\
\textbf{\kdlogit{}} & 0.16 $\pm 0.08$ & 0.13 $\pm 0.06$ & 0.96 $\pm 0.09$ & 0.96 $\pm 0.07$ & 0.68 $\pm 0.10$ & 0.82 $\pm 0.11$ \\
\textbf{\kdrep{}} & 0.12 $\pm 0.08$ & 0.09 $\pm 0.06$ & 0.82 $\pm 0.09$ & \underline{1.30 $\pm 0.07$} & 0.74 $\pm 0.10$ & 1.06 $\pm 0.10$ \\\midrule
& \multicolumn{6}{c}{\textbf{Test Slot Filling On}}\\ &  German &  English &  French &  Spanish &  Hindi &  Thai  \\ \rowcolor{lightgray} \multicolumn{7}{c}{Shared \{Trans, Task\} Baselines } \\\midrule
\naiveft{} & 1.71 $\pm 0.98$ & 1.24 $\pm 0.71$ & 2.01 $\pm 0.90$ & 0.54 $\pm 0.97$ & 0.20 $\pm 0.91$ & 2.50 $\pm 0.75$ \\
\jointinc{} & 1.59 $\pm 0.90$ & -0.17 $\pm 0.89$ & 1.22 $\pm 0.84$ & 1.08 $\pm 0.94$ & -1.10 $\pm 1.04$ & 2.36 $\pm 0.79$ \\\midrule
\rowcolor{lightgray} \multicolumn{7}{c}{Model Expansion Baselines}\\ \midrule
\textbf{\spectrans{}} & 1.75 $\pm 0.95$ & \underline{1.37 $\pm 0.80$} & 1.85 $\pm 0.83$ & -0.25 $\pm 0.91$ & -0.67 $\pm 0.93$ & 1.67 $\pm 0.74$ \\
\textbf{\specenca{}} & 1.80 $\pm 0.92$ & 0.45 $\pm 1.05$ & \underline{2.11 $\pm 0.86$} & 0.67 $\pm 0.98$ & 0.51 $\pm 0.88$ & 3.12 $\pm 0.88$ \\
\textbf{\spechead{}} & 2.28 $\pm 1.07$ & -0.27 $\pm 0.86$ & 1.55 $\pm 1.07$ & 0.56 $\pm 1.26$ & 0.44 $\pm 0.94$ & 2.36 $\pm 0.86$ \\
\textbf{\adatuned{}} & \underline{3.24 $\pm 0.94$} & -0.54 $\pm 0.72$ & 1.04 $\pm 0.95$ & \underline{1.59 $\pm 0.94$} & \textbf{3.37 $\pm 0.98$} & \underline{3.53 $\pm 0.82$} \\
\textbf{\adafrozen{}} & \textbf{3.48 $\pm 1.00$} & \textbf{3.38 $\pm 0.87$} & 1.46 $\pm 1.00$ & \textbf{4.68 $\pm 1.04$} & \underline{2.11 $\pm 1.06$} & \textbf{8.48 $\pm 1.27$} \\\midrule
\rowcolor{lightgray}  \multicolumn{7}{c}{Other Continuous Learning Algorithms }\\ \midrule
\textbf{\ewc{}} & 1.58 $\pm 1.02$ & 0.39 $\pm 0.82$ & \underline{2.11 $\pm 0.87$ }& 1.58 $\pm 1.05$ & -0.09 $\pm 0.93$ & 3.19 $\pm 0.73$ \\
\textbf{\er{}} & 1.97 $\pm 0.93$ & 0.29 $\pm 0.89$ & 2.05 $\pm 0.94$ & 1.38 $\pm 1.04$ & 0.23 $\pm 0.87$ & 2.87 $\pm 0.93$ \\
\textbf{\kdlogit{}} & 2.20 $\pm 0.98$ & 0.50 $\pm 0.83$ & 2.00 $\pm 0.84$ & 1.35 $\pm 1.00$ & -0.64 $\pm 0.94$ & 2.97 $\pm 0.76$ \\
\textbf{\kdrep{}} & 1.90 $\pm 0.88$ & 0.90 $\pm 0.75$ & \textbf{2.54 $\pm 0.88$} & 1.01 $\pm 0.91$ & -0.23 $\pm 0.96$ & 2.45 $\pm 0.75$ \\
\bottomrule
\end{tabular}
}
\caption{\label{tab:full-cll-fwt-lang} CCL per language analysis of transfer. The best and second best scores for each language are highlighted in \textbf{bold} and \underline{underlined}, respectively.}
\end{table*}

\begin{table*}[ht!] 
\small
\centering
\scalebox{1}{
\begin{tabular}{l|llllll} \toprule  
\multirow{3}{*}{\textbf{Model}}  & \multicolumn{6}{c}{\textbf{Test Intent Accuracy On}} \\
&  German &  English  & French & Hindi &  Spanish &  Thai  \\ 
\midrule \rowcolor{lightgray} \multicolumn{7}{c}{Shared \{Trans, Task\} Baselines}   \\ \midrule 
\naiveft{} & 58.53 $\pm 1.49$ & 69.09 $\pm 12.56$ & 60.83 $\pm 3.24$ & 59.42 $\pm 24.92$ & 33.38 $\pm 1.35$ & \underline{20.17 $\pm 1.10$} \\
\jointinc{} & 58.48 $\pm 2.13$ & 70.13 $\pm 12.56$ & 61.17 $\pm 2.62$ & 61.18 $\pm 19.86$ & 32.28 $\pm 2.56$ & 17.20 $\pm 0.19$ \\\midrule
\rowcolor{lightgray} \multicolumn{7}{c}{Model Expansion Baselines}\\ \midrule
\textbf{\spectrans{}} & -1.42 $\pm 0.00$ & 0.44 $\pm 0.01$ & -0.01 $\pm 0.01$ & -0.95 $\pm 0.01$ & -0.15 $\pm 0.00$ & -0.47 $\pm 0.00$ \\
\textbf{\specenca{}} & 26.17 $\pm 7.44$ & 33.16 $\pm 10.88$ & 25.56 $\pm 7.00$ & 27.21 $\pm 18.32$ & 21.79 $\pm 2.33$ & 11.51 $\pm 0.77$ \\
\textbf{\spechead{}} & -0.25 $\pm 0.12$ & 0.38 $\pm 0.01$ & 0.63 $\pm 0.06$ & -0.66 $\pm 0.02$ & 0.60 $\pm 0.03$ & -0.09 $\pm 0.01$ \\
\textbf{\adatuned{}} & 55.95 $\pm 0.91$ & 67.93 $\pm 14.89$ & 60.21 $\pm 4.16$ & 58.14 $\pm 33.89$ & \textbf{36.44 $\pm 4.20$} & 17.40 $\pm 1.10$ \\
\textbf{\adafrozen{}} & 5.08 $\pm 0.51$ & 14.37 $\pm 1.06$ & 7.61 $\pm 0.49$ & 6.87 $\pm 1.00$ & 5.50 $\pm 0.90$ & -0.30 $\pm 0.04$ \\\midrule
\rowcolor{lightgray}  \multicolumn{7}{c}{Other Continuous Learning Algorithms }\\ \midrule
\textbf{\ewc{}} & \underline{58.57 $\pm 1.77$} & \underline{69.39 $\pm 12.59$} & \underline{60.71 $\pm 3.48$} & 58.99 $\pm 24.22$ & 33.59 $\pm 1.40$ & 19.71 $\pm 1.30$ \\
\textbf{\er{}} & \textbf{59.70 $\pm 1.68$} & \textbf{70.20 $\pm 13.83$} & \textbf{61.32 $\pm 4.05$} & \textbf{60.09 $\pm 24.40$} & 33.38 $\pm 1.24$ & 19.57 $\pm 1.48$ \\
\textbf{\kdlogit{}} & 58.12 $\pm 1.32$ & 68.87 $\pm 12.38$ & 60.85 $\pm 3.45$ & \underline{59.69 $\pm 24.27$} & 33.55 $\pm 1.46$ & 19.99 $\pm 1.19$ \\
\textbf{\kdrep{}} & 58.47 $\pm 1.20$ & 68.64 $\pm 12.23$ & 60.96 $\pm 3.56$ & \underline{59.69 $\pm 24.54$} & \underline{34.22 $\pm 1.07$} & \textbf{20.49 $\pm 1.00$} \\\midrule
& \multicolumn{6}{c}{\textbf{Test Slot Filling On}}\\ &  German &  English &  French &  Spanish &  Hindi &  Thai \\ \rowcolor{lightgray} \multicolumn{7}{c}{Shared \{Trans, Task\} Baselines } \\\midrule
\naiveft{} & 44.25 $\pm 1.16$ & 48.42 $\pm 8.10$ & 47.58 $\pm 1.63$ & 46.60 $\pm 15.31$ & \underline{18.97 $\pm 0.44$} & 12.09 $\pm 0.33$ \\
\jointinc{} & \textbf{44.73 $\pm 1.68$} & \underline{48.74 $\pm 10.90$} & \underline{47.67 $\pm 2.19$} & \underline{46.98 $\pm 18.10$} & 18.05 $\pm 0.31$ & 12.20 $\pm 0.22$ \\\midrule
\rowcolor{lightgray} \multicolumn{7}{c}{Model Expansion Baselines}\\ \midrule
\textbf{\spectrans{}} & 0.45 $\pm 0.00$ & 0.76 $\pm 0.01$ & 0.33 $\pm 0.00$ & 0.83 $\pm 0.01$ & 0.00 $\pm 0.00$ & 0.15 $\pm 0.00$ \\
\textbf{\specenca{}} & 14.81 $\pm 3.81$ & 15.50 $\pm 6.12$ & 16.09 $\pm 4.03$ & 16.11 $\pm 8.84$ & 6.62 $\pm 1.29$ & 4.80 $\pm 0.35$ \\
\textbf{\spechead{}} & 0.07 $\pm 0.00$ & 0.15 $\pm 0.00$ & 0.08 $\pm 0.00$ & 0.04 $\pm 0.00$ & -0.02 $\pm 0.00$ & 0.09 $\pm 0.00$ \\
\textbf{\adatuned{}} & 41.08 $\pm 1.24$ & 44.36 $\pm 18.19$ & 45.26 $\pm 2.44$ & 42.56 $\pm 21.09$ & 17.62 $\pm 1.27$ & 10.72 $\pm 0.13$ \\
\textbf{\adafrozen{}} & 4.42 $\pm 0.10$ & 1.12 $\pm 0.04$ & 4.51 $\pm 0.32$ & 4.86 $\pm 0.93$ & 1.80 $\pm 0.03$ & 0.09 $\pm 0.00$ \\\midrule
\rowcolor{lightgray}  \multicolumn{7}{c}{Other Continuous Learning Algorithms }\\ \midrule
\textbf{\ewc{}} & \underline{44.17 $\pm 1.16$} & 48.52 $\pm 8.21$ & 47.51 $\pm 1.62$ & 46.38 $\pm 15.32$ & 18.94 $\pm 0.42$ & 12.32 $\pm 0.30$ \\
\textbf{\er{}} & \textbf{44.73 $\pm 1.45$} & \textbf{49.60 $\pm 9.35$} & \textbf{48.17 $\pm 2.22$} & \textbf{47.26 $\pm 15.85$} & \textbf{19.06 $\pm 0.44$} & 12.62 $\pm 0.24$ \\
\textbf{\kdlogit{}} & 43.79 $\pm 1.04$ & 48.30 $\pm 8.21$ & 47.31 $\pm 2.05$ & 46.77 $\pm 15.51$ & 18.85 $\pm 0.37$ & \underline{12.49 $\pm 0.22$} \\
\textbf{\kdrep{}} & 43.81 $\pm 1.35$ & 48.10 $\pm 7.99$ & 47.38 $\pm 1.85$ & 46.60 $\pm 15.21$ & 18.83 $\pm 0.45$ & \textbf{12.82 $\pm 0.26$} \\
\bottomrule
\end{tabular}
}
\caption{\label{tab:full-cll-fwt-0-lang} CCL per language zero-shot forward transfer. The best and second best scores for each language for intent classification and slot filling independently across approaches are highlighted in \textbf{bold} and \underline{underlined}, respectively.}
\end{table*}


\subsection{More Analysis}
\label{app:more-analysis}


\begin{figure*}[ht!]
 \centering
\includegraphics[width=1\textwidth]{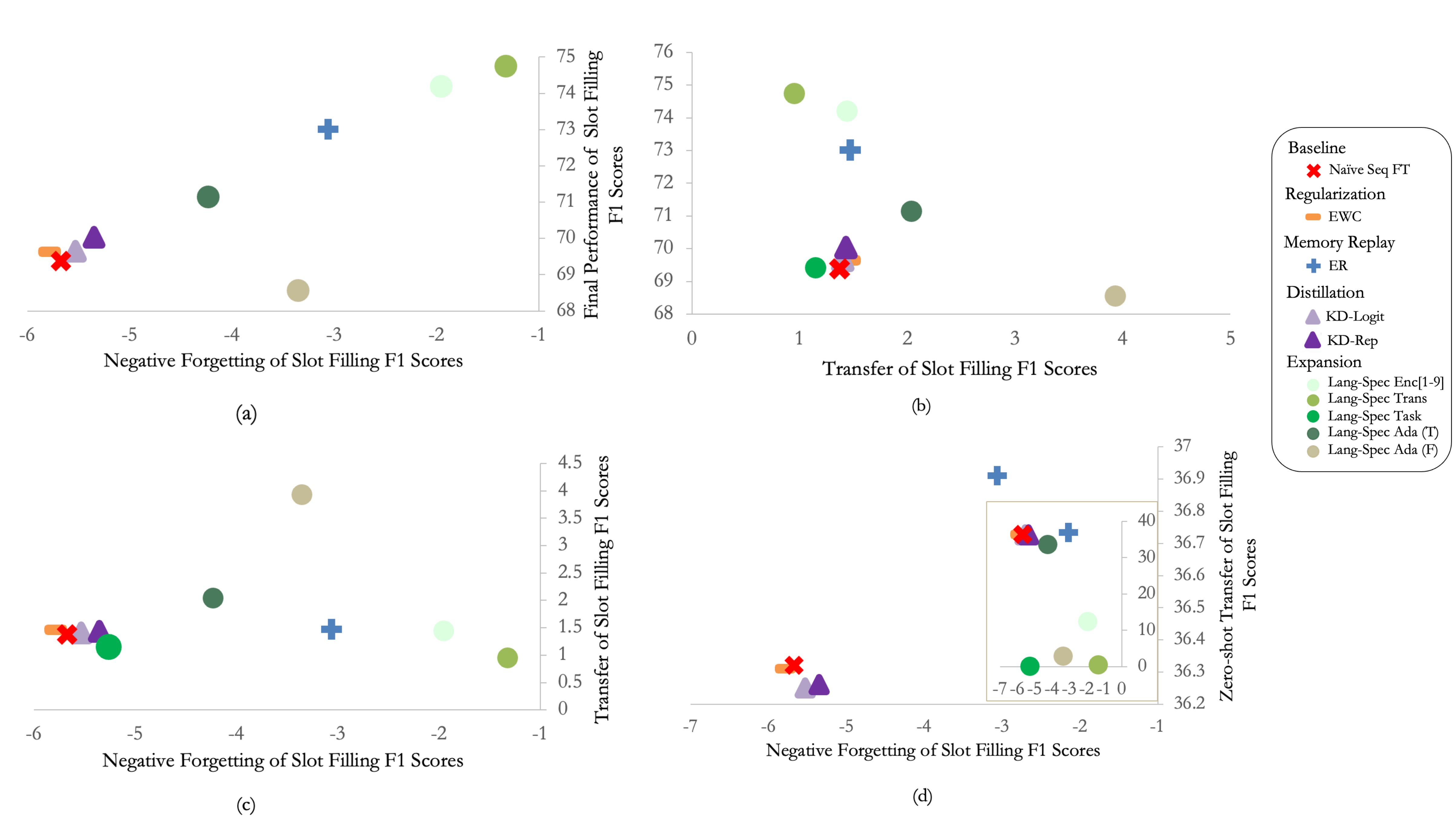}
 \caption{Correlations between different pairs of metrics: (a) Final performance versus negative forgetting for the task of slot filling. The lower the forgetting, the higher the final performance. (b) Final performance versus transfer for the task of slot filling. (c) Transfer versus negative forgetting for slot filling task. (d) Zero-shot generalization versus negative forgetting for slot filling. Model expansion approaches are highlighted in shades of green. We zoom over the rest of the models in the main graph and show an overview of all approaches in the lower right corner subplot. The same trends observed for intent classification in Figure~\ref{fig:all-correlation-intents} can be observed here.}
 \label{fig:all-correlation-slots}
 \vspace{-0.5cm}
\end{figure*}

\begin{figure*}[ht!]
    \begin{subfigure}[b]{0.5\textwidth}
        \centering
        \includegraphics[width=1\textwidth]{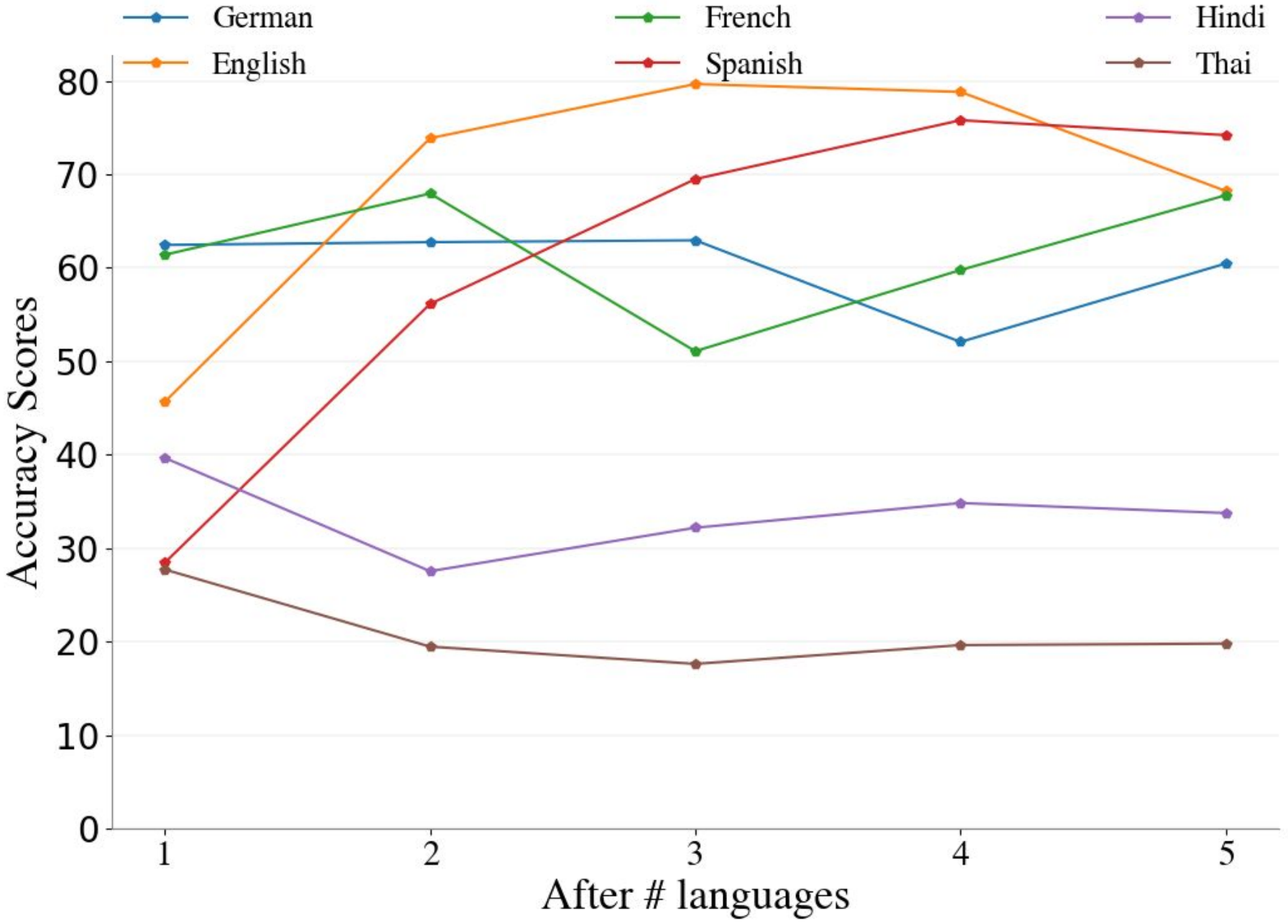}
        \caption{Accuracy for intent classification.}
        \label{fig:generalization-hops-lang-intent}
    \end{subfigure}
    \begin{subfigure}[b]{0.5\textwidth}
        \includegraphics[width=1\textwidth]{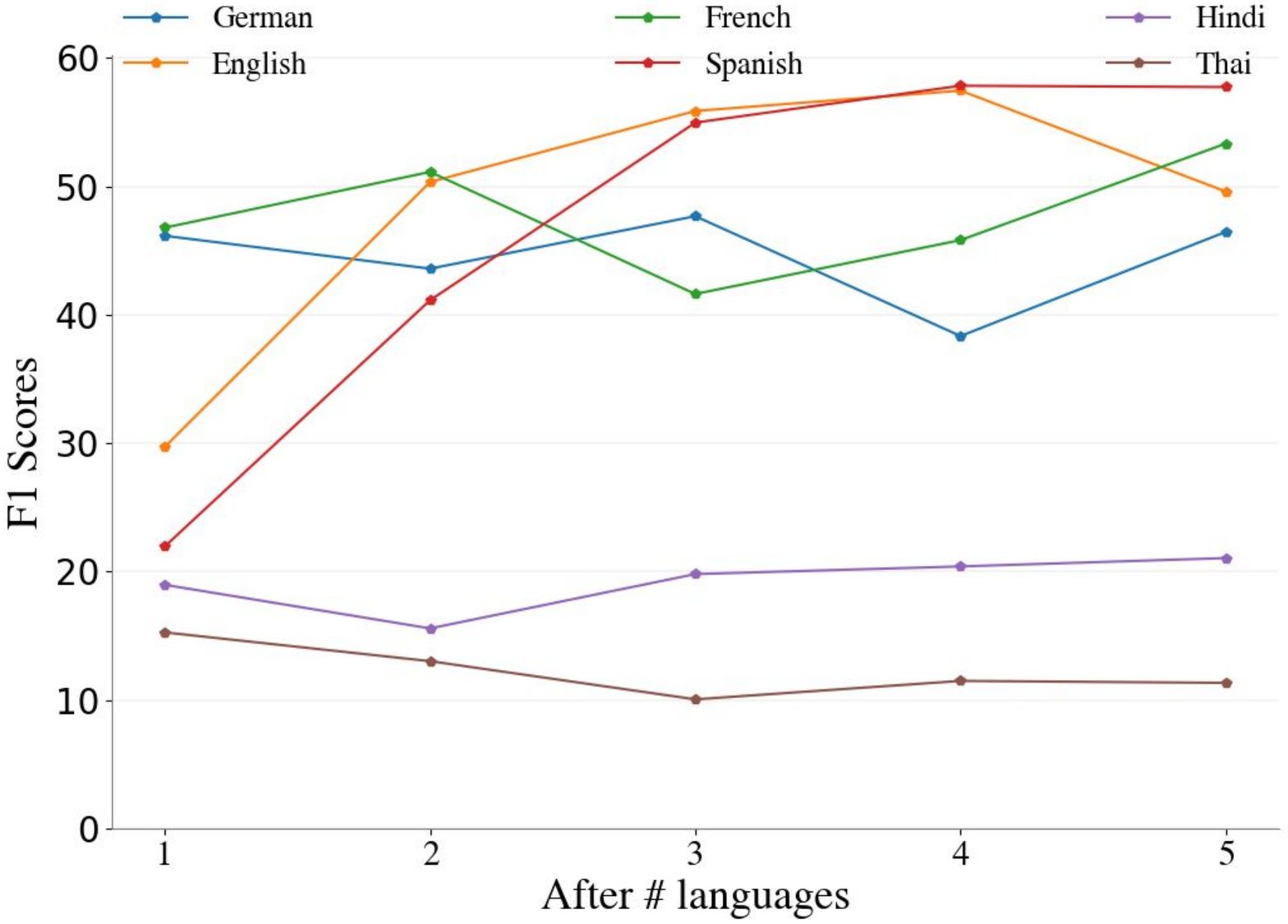}
         \caption{F1 score for slot filling.}
        \label{fig:generalization-hops-lang-slots}
     \end{subfigure}
 \caption{Comparing cross-lingual generalization of \naiveft{} across many hops and different languages for intent classification and slot filling.}
 \label{fig:generalization-naiveft-lang-hops}
\end{figure*}

Figure~\ref{fig:all-correlation-slots} plots final performance versus negative forgetting, final performance versus transfer, transfer versus negative forgetting, and zero-shot transfer versus negative forgetting for the subtask of slot filling. The same trends observed for intent classification can also be observed for slot filling. Figures~\ref{fig:generalization-hops-lang-intent} and~\ref{fig:generalization-hops-lang-slots} show how \naiveft{} intent classification accuracy score and slot filling F1 score, respectively, change for each language separately after different hops of training. We can see that although the performance increases as more hops are seen for high-resource Latin-script languages like English, Spanish, and to some degree French, the same cannot be said for low-resource languages Thai and Hindi, which also suffer from being script-isolated.

\begin{figure*}[ht!]
    \begin{subfigure}[b]{0.5\textwidth}
        \centering
        \includegraphics[width=1\textwidth]{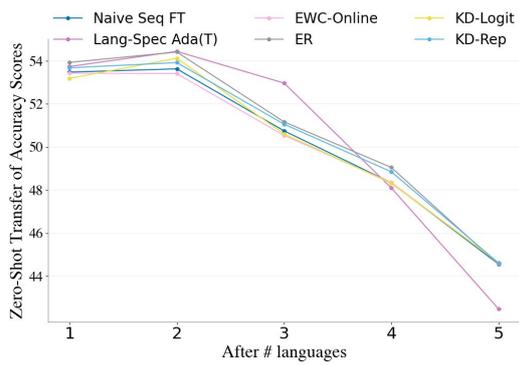}
        \caption{Zero-shot transfer of accuracy for intent classification.}
        \label{fig:generalization-hops-intents-zero-shot}
    \end{subfigure}
    \begin{subfigure}[b]{0.5\textwidth}
        \centering
        \includegraphics[width=1\textwidth]{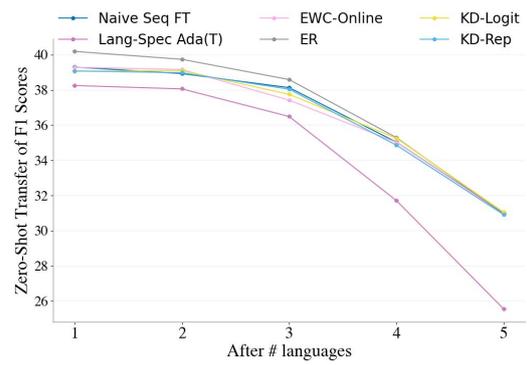}
        \caption{Zero-shot transfer of f1 score for slot filling.}
        \label{fig:generalization-hops-slots-zero-shot}
    \end{subfigure}
    \begin{subfigure}[b]{0.5\textwidth}
        \centering
        \includegraphics[width=1\textwidth]{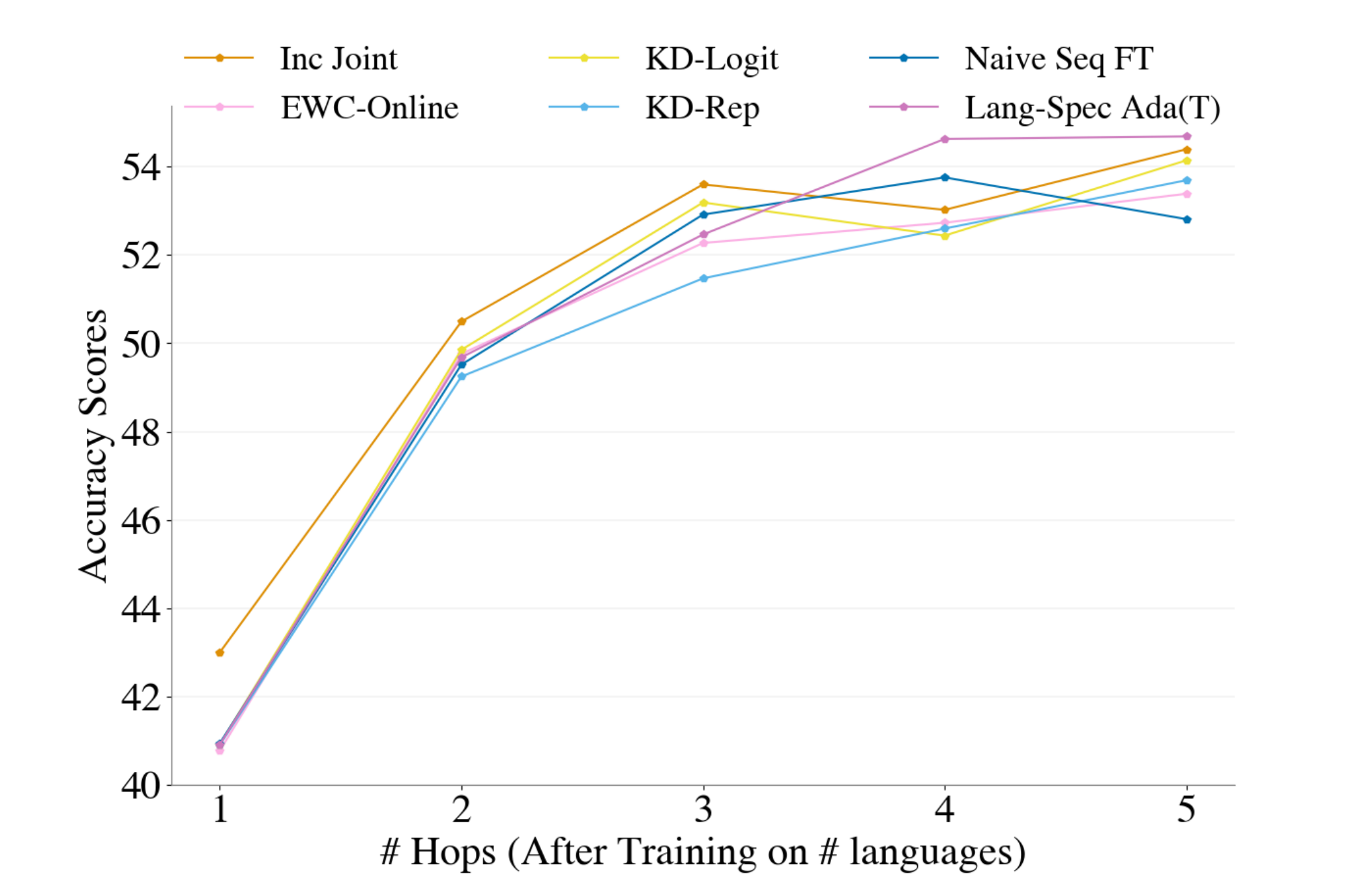}
        \caption{Accuracy for intent classification.}
        \label{fig:generalization-hops-intent}
    \end{subfigure}
    \begin{subfigure}[b]{0.5\textwidth}
        \includegraphics[width=1\textwidth]{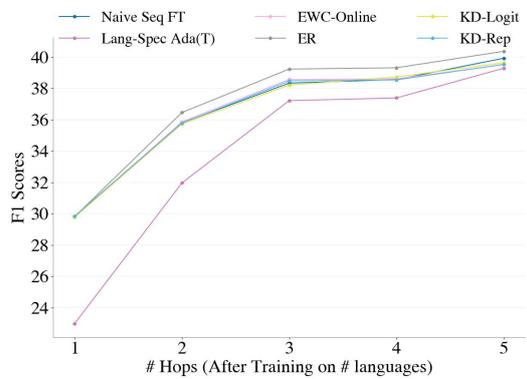}
    \caption{F1 score for slot filling.}
        \label{fig:generalization-hops-slots}
     \end{subfigure}
 \caption{Measuring cross-lingual generalization to new languages across many hops for intent classification and slot filling. This is both in terms of zero-shot transfer metric and plain accuracy and f1 scores.}
 \label{fig:generalization-hops}
\end{figure*}

\begin{figure*}[ht!]
\begin{subfigure}[b]{0.45\textwidth}
 \centering
\includegraphics[width=1\textwidth]{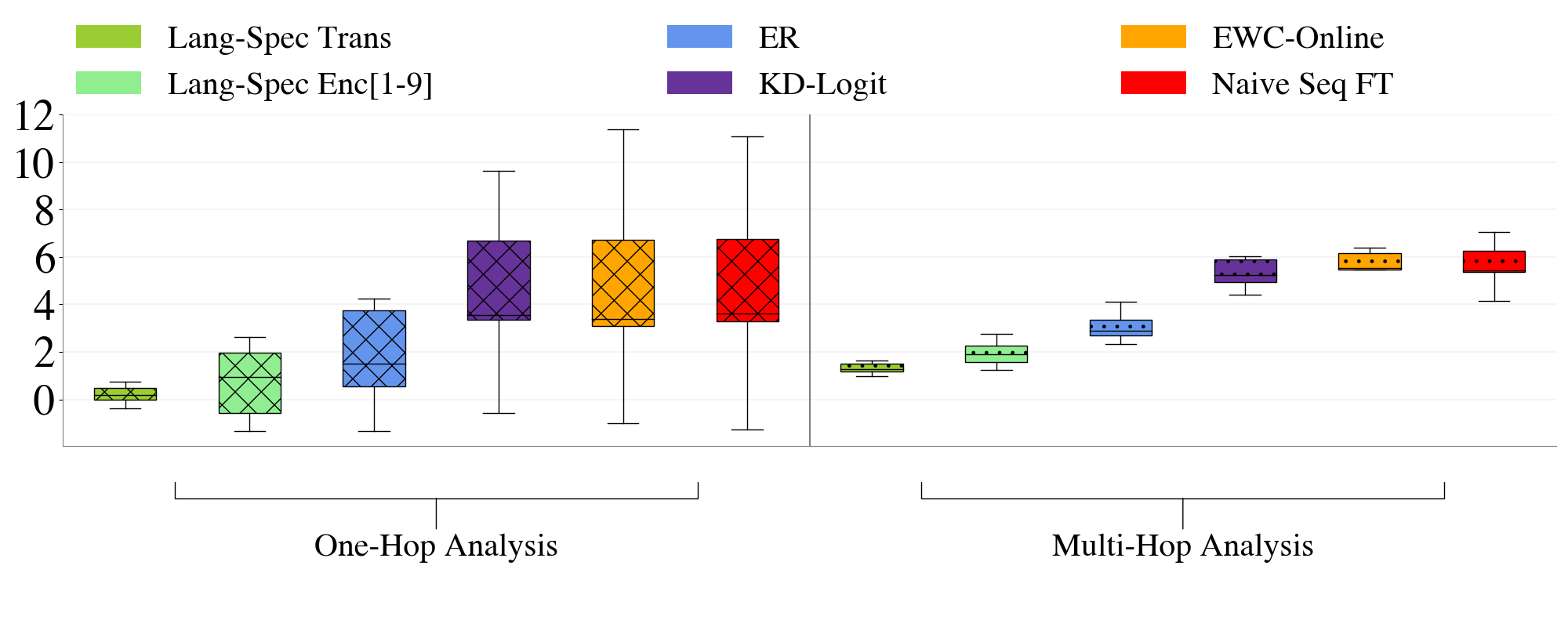}
 \caption{Forgetting for slot filling.}
 \label{fig:forg-steps-slots}
\end{subfigure}
\begin{subfigure}[b]{0.45\textwidth}
 \centering
\includegraphics[width=1\textwidth]{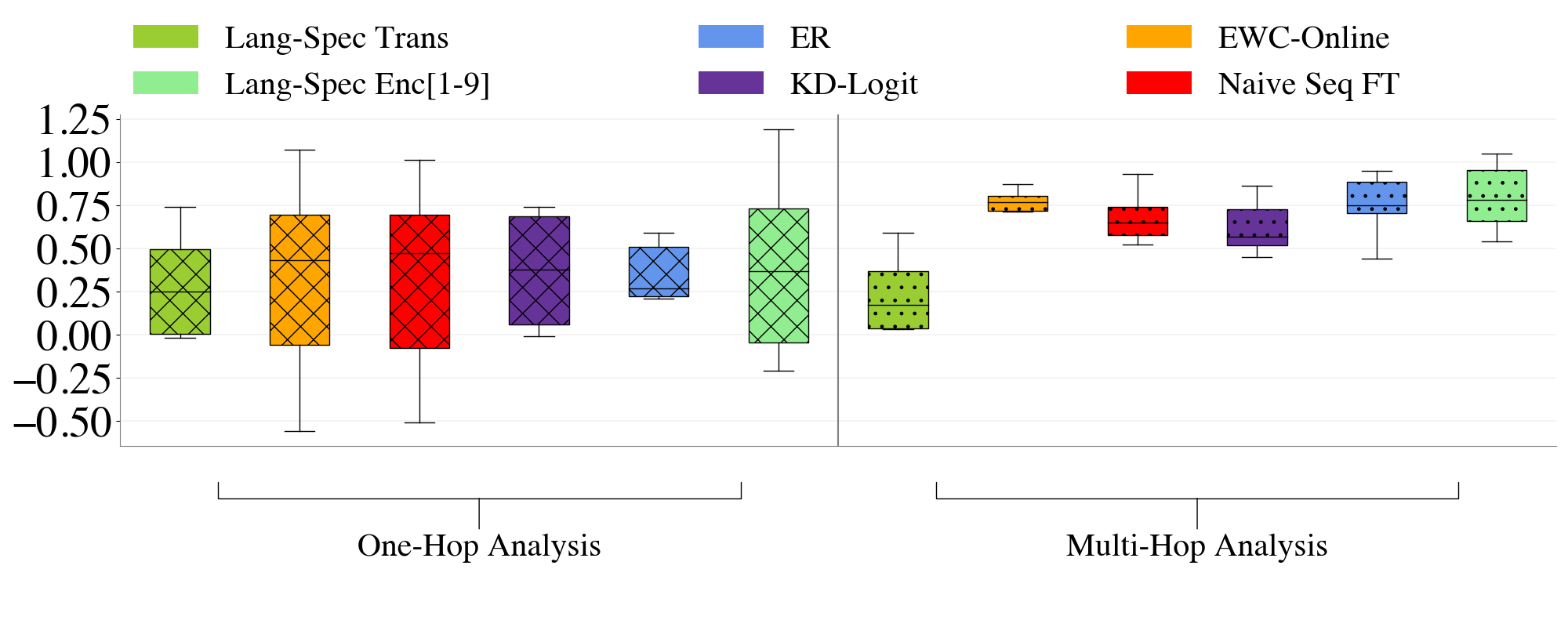}
 \caption{Transfer for intent classification.}
 \label{fig:trans-steps-intents}
\end{subfigure}
\begin{subfigure}[b]{0.45\textwidth}
 \centering
\includegraphics[width=1\textwidth]{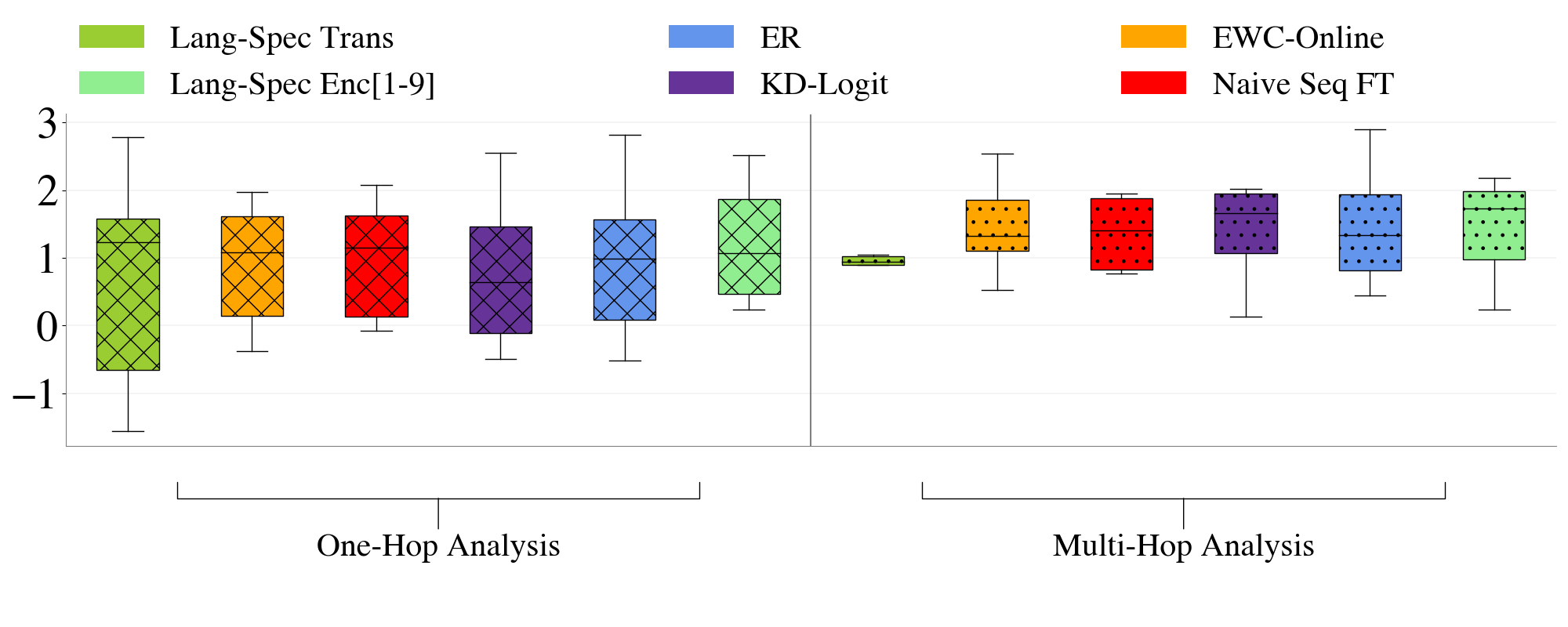}
 \caption{Transfer for slot filling.}
 \label{fig:trans-steps-slots}
\end{subfigure}
\begin{subfigure}[b]{0.45\textwidth}
 \centering
\includegraphics[width=1\textwidth]{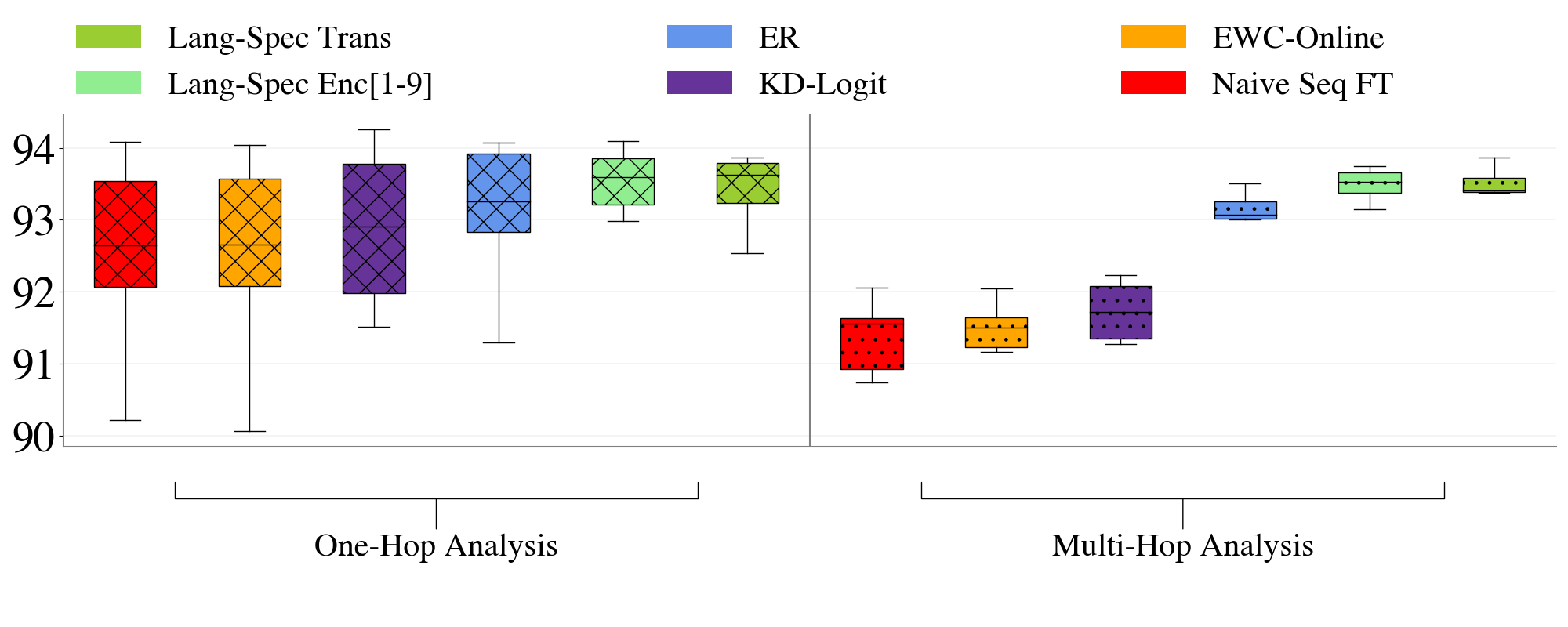}
 \caption{Final performance for intent classification.}
 \label{fig:fp-steps-intents}
\end{subfigure}
\begin{subfigure}[b]{\linewidth}
 \centering
\includegraphics[width=0.5\textwidth]{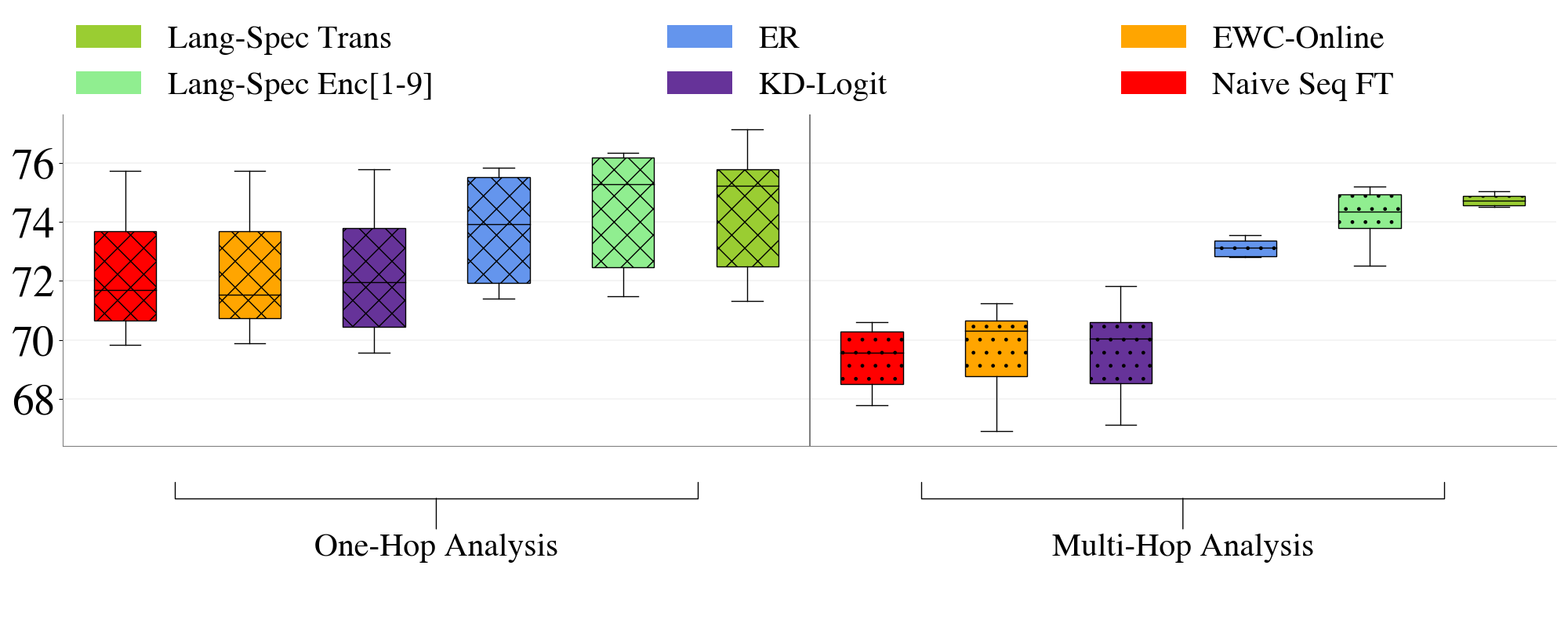}
 \caption{Final performance for slot filling.}
 \label{fig:fp-steps-slots}
\end{subfigure}

\caption{\label{fig:two-multi-steps-more} Comparison between different metrics using one-hop (crossed boxplots) and multi-hop analysis (dotted boxplots), on the left and right, respectively, for each approach.}
\end{figure*}

To analyze the zero-shot generalization to unseen languages, we analyze the performance of each model across different hops. In other words, we consider the average performance after seeing from 1 to 5 languages, enabled by the balanced datastreams we carefully curated~\ref{sec:downstream-task-datastreams}. We can check the performance after training on each $x$ language(s) from exactly one datastream. Figures~\ref{fig:generalization-hops-intents-zero-shot} and~\ref{fig:generalization-hops-slots-zero-shot} show a comparison between different approaches across different hops of training using zero-shot transfer metrics for intent classification and slot filling, respectively. In general, we can observe that the average performance of the zero-shot transfer decreases after seeing $n$ languages, where $n \in [1\ldots 5]$. In this case, after seeing one language, the performance is equivalent to conventional transfer learning involving two hops, whereas the performance after seeing $n >= 2$ is for multi-hop continual learning. We notice that as we increase the number of hops, the transfer capabilities decrease nearly uniformly across most approaches, making the problem more challenging and different from conventional transfer learning. 
Figures~\ref{fig:generalization-hops-intent} and ~\ref{fig:generalization-hops-slots} show the generalization trends for different continual learning approaches compared to the baselines for intent classification and slot filling, respectively. We can see that most continual learning approaches improve in terms of both intent accuracy and slot filling F1 scores over \naiveft{}, and the gap increases mainly as more languages are seen (except at $hop_4$). After 5 hops, there is a clear gap between \naiveft{} and continual learning approaches on top of them \adatuned{} and \kdlogit{}. Figure~\ref{fig:two-multi-steps-more} shows more results for multi-hop versus one-hop analysis for more metrics and tasks. In general, we can observe the same trend, whereby multi-hop dotted boxplots analysis has smaller confidence intervals than one-hop crossed boxplots.

\subsection{Experience Replay Ablation Studies}
\label{app:er-ablation-studies}
Table \ref{tab:er-ablations} shows a comparison between the performance of experience replay variants with different memory sizes ranging from 750 to 6000 instances which accounts for 5\% to 60\% of the training data for each language. Although we notice that forgetting is the lowest and the final performance is the highest when a memory of 6000 instances is used, the gap is not that significant as the memory is scaled down. Moreover, differences in transfer are not correlated with the size of the memory. We notice that ER achieves a performance that surpasses \naiveft{} even when using the lowest memory size. This suggests that even tiny bits of memory are helpful. 
\begin{table*}[ht] 
\centering
\scalebox{0.75}{
\begin{tabular}{l|ll|ll|ll|ll} \toprule  
\multirow{2}{*}{\textbf{Model}}  
& \multicolumn{2}{c|}{F $\downarrow$} &  \multicolumn{2}{c|}{T $\uparrow$} &  \multicolumn{2}{c|}{$T^0$ $\uparrow$} &  \multicolumn{2}{c}{FP $\uparrow$}  \\ & Acc & F1 & Acc & F1 & Acc & F1 & Acc & F1 \\
\midrule
\naiveft{} & 2.93 $\pm 1.24$ & 5.67 $\pm 0.93$ & 0.68 $\pm 0.14$ & 1.37 $\pm 0.53$ & 50.24 $\pm 3.43$ & 36.32 $\pm 1.91$ & 91.06 $\pm 1.08$ & 69.37 $\pm 1.06$ \\ \midrule
\er{}-750 & 1.97 $\pm 0.73$ & 4.28 $\pm 0.63$ & 0.65 $\pm 0.19$ & 1.46 $\pm 0.59$ & 50.41 $\pm 3.19$ & 36.53 $\pm 1.91$ & 92.10 $\pm 0.68$ & 71.65 $\pm 1.02$ \\
\er{}-1500 & 1.55 $\pm 0.44$ & 3.88 $\pm 0.42$ & 0.68 $\pm 0.26$ & 1.55 $\pm 0.69$ & \underline{50.83 $\pm 3.38$} & 36.59 $\pm 1.93$ & 92.65 $\pm 0.35$ & 71.68 $\pm 0.71$ \\
\er{}-3000 & \underline{1.40 $\pm 0.44$} & \underline{3.36 $\pm 0.47$} & \underline{0.70 $\pm 0.25$} & \textbf{1.48 $\pm 0.71$} & \textbf{51.03 $\pm 3.60$} & 36.77 $\pm 2.06$ & \underline{92.93 $\pm 0.37$} & \underline{72.71 $\pm 0.56$} \\
\er{}-4500 & 1.43 $\pm 0.58$ & 3.39 $\pm 0.75$ & 0.59 $\pm 0.11$ & 1.44 $\pm 0.38$ & 50.46 $\pm 3.68$ & \underline{36.91 $\pm 2.19$} & 92.73 $\pm 0.72$ & 72.46 $\pm 1.05$ \\
\er{}-6000 & \textbf{1.29 $\pm 0.51$} & \textbf{3.06 $\pm 0.59$} & \textbf{0.75 $\pm 0.17$} & \underline{1.47 $\pm 0.85$} & 50.71 $\pm 3.55$ & \textbf{36.91 $\pm 2.14$} & \textbf{93.09 $\pm 0.29$} & \textbf{73.00 $\pm 0.52$} \\
\bottomrule
\end{tabular}
}
\caption{\label{tab:er-ablations} Ablation Studies of Experience Replay where we experiment with different memory sizes per language. For each metric and score, we highlight the best score in \textbf{bold} and \underline{underline} the second best score.}
\end{table*}

\section{More Results using Multiple Seeds}
\label{app:seeds-experiments}
In this section, we show the results using different seeds for key experiments in the main paper. We show in Table~\ref{tab:cll-metrics-average-fp-seeds} and~\ref{tab:cll-metrics-average-seeds} the average final performance, forgetting, and transfer averaged across different language permutations for the baseline model compared to reference models. We also show in Table \ref{tab:short-metrics-perm-seeds} the performance on intent classification comparison between the baseline and different continual learning algorithms across \htol{} and \ltoh{}. Overall, we notice the same trends and findings observed earlier in Tables~\ref{tab:cll-metrics-average-fp},~\ref{tab:cll-metrics-average}, and~\ref{tab:short-metrics-perm}.

\begin{table}[ht] 
\centering
\scalebox{0.70}{
\begin{tabular}{l|ll} \toprule  
\textbf{Model} & Acc & F1 \\
\toprule
\naiveft{} & 90.40 $\pm 1.53$ & 65.01 $\pm 1.25$ \\
\langspec{} & 93.28 $\pm 0.31$ & 68.93 $\pm 1.17$ \\
\jointinc{} & \underline{94.14 $\pm 0.08$} & \underline{71.70 $\pm 0.43$} \\
\multi{} & \textbf{94.20 $\pm 0.21$} & \textbf{72.23 $\pm 0.99$} \\

\bottomrule
\end{tabular}
}
\caption{\label{tab:cll-metrics-average-fp-seeds} The average final performance across different language permutations for the baseline compared to reference models using multiple seeds. We highlight the best scores in \textbf{bold} and \underline{underline} the second best across models. We notice the same findings as when using bootstrap sampling but with tighter confidence intervals, as shown in Table~\ref{tab:cll-metrics-average-fp}.
}
\end{table}

\begin{table}[ht!] 
\centering
\scalebox{0.68}{
\begin{tabular}{l|ll|ll} \toprule  
\multirow{2}{*}{\textbf{Model}}  
& \multicolumn{2}{c|}{F $\downarrow$} &  \multicolumn{2}{c}{T $\uparrow$}  \\ & Acc & F1 & Acc & F1  \\
\toprule
\naiveft{} & 3.2 $\pm 1.66$ & 5.47 $\pm 0.87$ & \textbf{0.73 $\pm 0.16$} & \textbf{2.75 $\pm 0.63$} \\
\jointinc{} & \textbf{-0.1 $\pm 0.01$} & \textbf{-0.38 $\pm 0.45$} & 0.57 $\pm 0.14$ & 1.73 $\pm 1.05$\\

\bottomrule
\end{tabular}
}
\caption{\label{tab:cll-metrics-average-seeds} Forgetting (F) and transfer (T) performance averaged across different language permutations for \emph{sequential baseline and reference models} using different seeds. We highlight the best models in \textbf{bold}. We notice exactly the same trends as when using bootstrap sampling for our analysis in Table~\ref{tab:cll-metrics-average}.}
\end{table}

\begin{table*}[ht!]
\centering
\scalebox{0.7}{
\begin{tabular}{l|ll||ll||ll} \toprule  
 \multirow{2}{*}{\textbf{Model}} & \multicolumn{2}{c||}{F $\downarrow$} &  \multicolumn{2}{c||}{T $\uparrow$}  &  \multicolumn{2}{c}{FP $\uparrow$}          \\ 
 & \htol{} & \ltoh{} & \htol{} & \ltoh{} & \htol{} & \ltoh{} \\
 \toprule
\naiveft{} & \textbf{1.37 $\pm 0.14$} & 5.38 $\pm 0.34$ & \textbf{0.95 $\pm 0.03$} & 0.56 $\pm 0.07$ & \textbf{91.83 $\pm 0.55$} & 88.28 $\pm 0.55$ \\
\spectrans{} & \underline{\textbf{0.01 $\pm 0.01$}} & \underline{0.17 $\pm 0.08$} & \textbf{0.57 $\pm 0.06$} & 0.09 $\pm 0.01$ & \underline{\textbf{93.81 $\pm 0.06$}} & \underline{93.27 $\pm 0.10$} \\
\spechead{} & \textbf{1.29 $\pm 0.08$} & 5.52 $\pm 0.87$ & \textbf{0.88 $\pm 0.12$} & 0.43 $\pm 0.19$ & \textbf{92.12 $\pm 0.18$} & 87.20 $\pm 1.76$ \\
\adatuned{} & \textbf{0.81 $\pm 0.08$} & 4.17 $\pm 0.30$ & \textbf{1.16 $\pm 0.09$} & 0.65 $\pm 0.06$ & \textbf{92.53 $\pm 0.22$} & 88.61 $\pm 0.44$ \\
\adafrozen{} & \textbf{0.38 $\pm 0.09$} & 1.04 $\pm 0.61$ & \textbf{\underline{3.54 $\pm 0.15$}} & \underline{2.34 $\pm 0.11$} & \textbf{91.15 $\pm 0.04$} & 90.0 $\pm 0.39$ \\
\ewc{} & \textbf{1.35 $\pm 0.24$} & 5.42 $\pm 0.60$ & \textbf{0.87 $\pm 0.11$} & 0.71 $\pm 0.12$ & \textbf{91.86 $\pm 0.52$} & 88.09 $\pm 0.20$ \\
\er{}-6000 & \textbf{0.69 $\pm 0.14$} & 1.93 $\pm 0.28$ & \textbf{0.93 $\pm 0.07$} & 0.72 $\pm 0.14$ & \textbf{93.43 $\pm 0.08$} & 92.50 $\pm 0.25$ \\
\kdlogit{} & \textbf{1.33 $\pm 0.11$} & 3.82 $\pm 0.23$ & \textbf{0.81 $\pm 0.11$} & 0.54 $\pm 0.07$ & \textbf{91.86 $\pm 0.31$} & 89.85 $\pm 0.4$ \\
\kdrep{} & \textbf{1.37 $\pm 0.1$} & 3.7 $\pm 0.25$ & \textbf{0.85 $\pm 0.23$} & 0.52 $\pm 0.13$ & \textbf{91.64 $\pm 0.49$} & 89.73 $\pm 0.8$ \\
 \bottomrule
\end{tabular}
}
\caption{\label{tab:short-metrics-perm-seeds} Performance on intent classification comparison between the baseline and continual learning algorithms across two language permutations using multiple seeds. We highlight in \textbf{bold} the lowest forgetting (F), highest transfer (T), and final performance (FP) of accuracy scores among \htol{} and \ltoh{}, whereas the best scores across approaches for \htol{} and \ltoh{} separately are \underline{underlined}. We notice the same trends and findings as Table~\ref{tab:short-metrics-perm}, where only bootstrap sampling is used to compute the confidence intervals.}
\vspace{-0.3cm}
\end{table*}

\section{Statistical Significance}
\label{sec:statistical-significance}
\begin{figure*}[ht]
 \centering
 \begin{subfigure}[b]{0.48\textwidth}
 \centering
\includegraphics[width=1\textwidth]{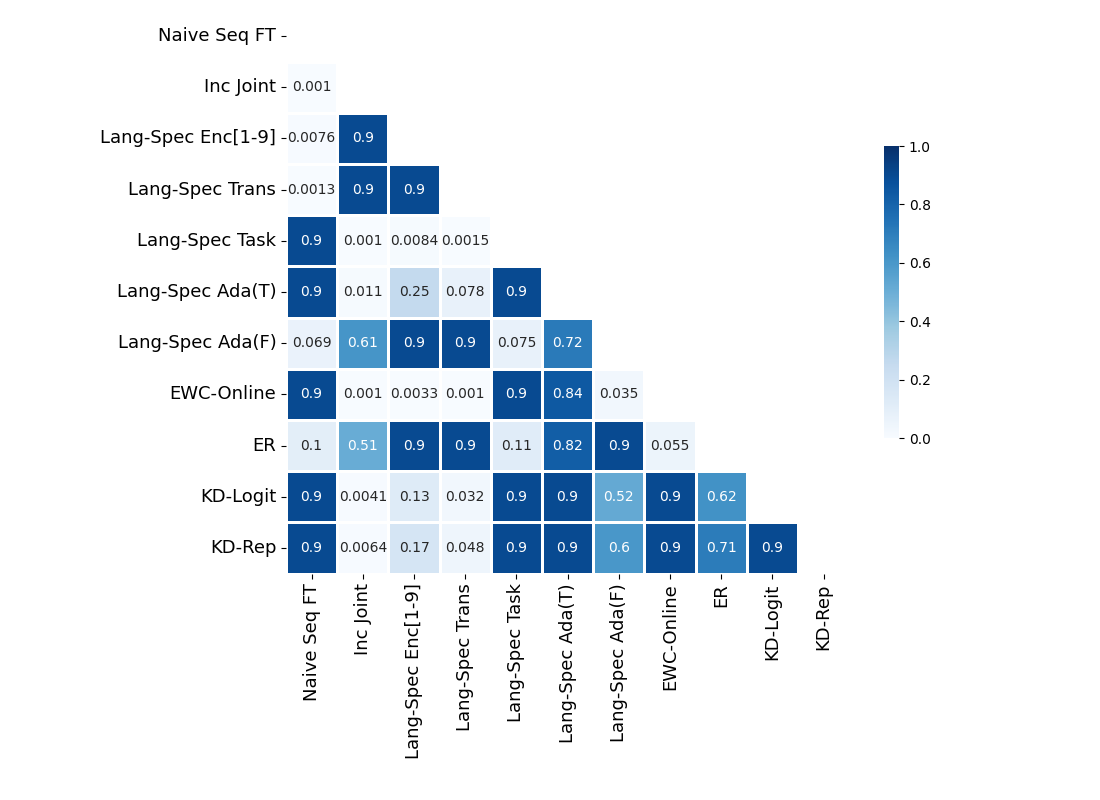}
 \caption{Forgetting of intent accuracy.}
 \label{fig:sg-intent-forget}
\end{subfigure}
\begin{subfigure}[b]{0.48\textwidth}
 \centering
 \includegraphics[width=1\textwidth]{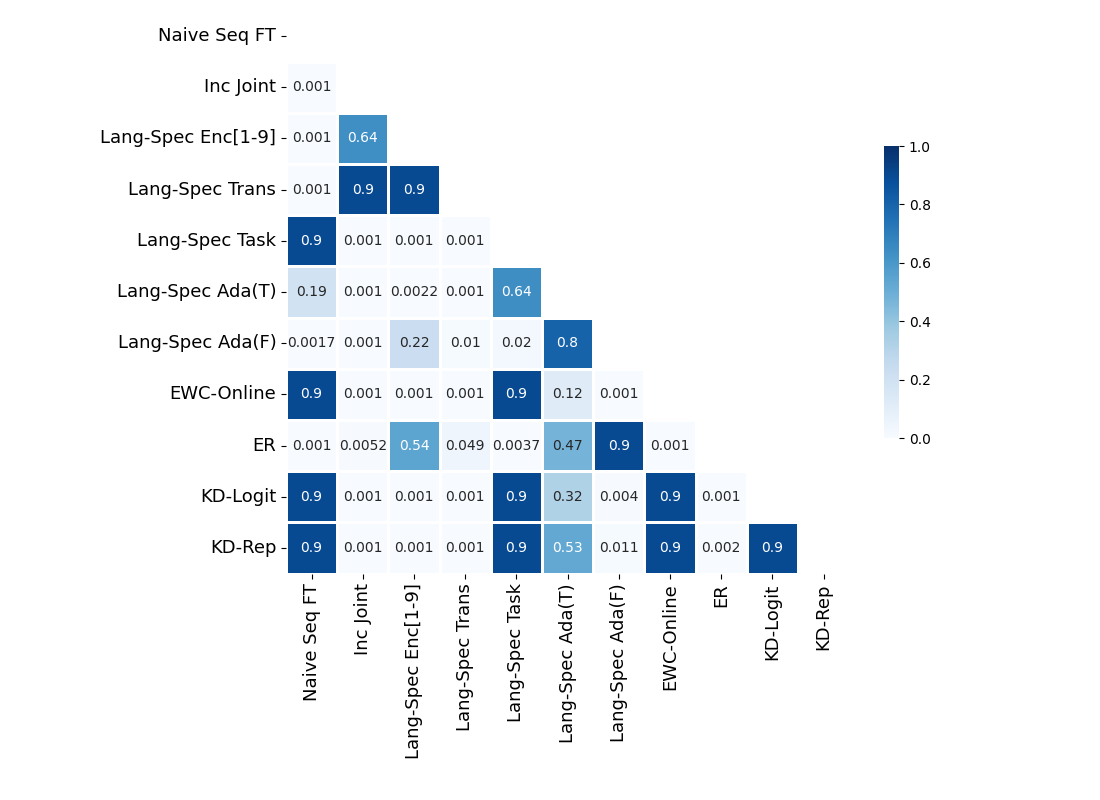}
 \caption{Forgetting of slot filling.}
 \label{fig:sg-slot-forget}
\end{subfigure}
\begin{subfigure}[b]{0.48\textwidth}
 \centering
 \includegraphics[width=1\textwidth]{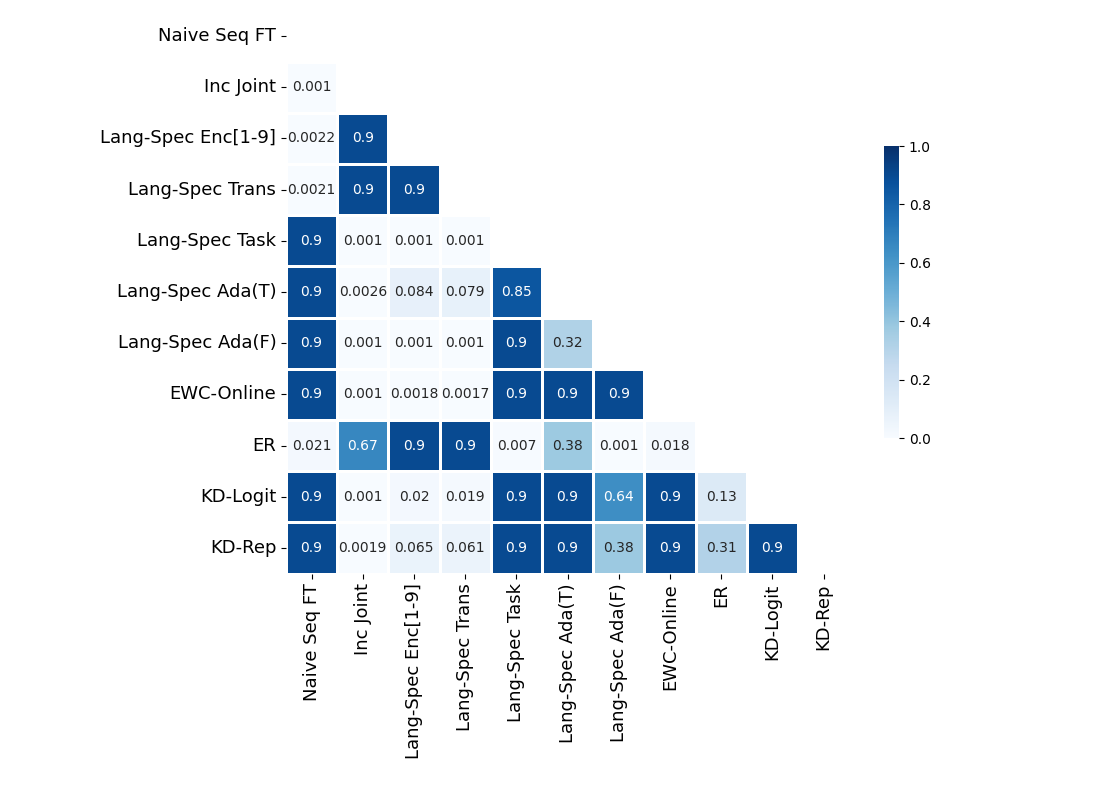}
 \caption{Final performance of intent accuracy.}
 \label{fig:sg-intent-fp}
\end{subfigure}
\begin{subfigure}[b]{0.48\textwidth}
 \centering
 \includegraphics[width=1\textwidth]{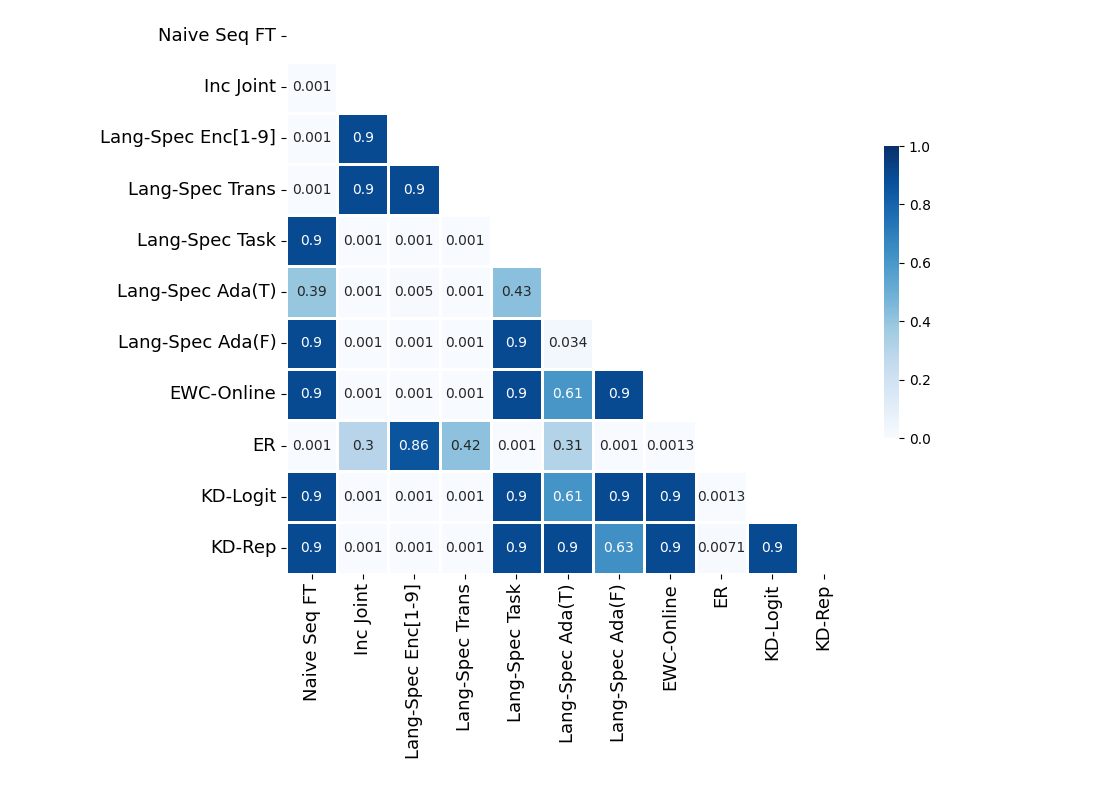}
 \caption{Final performance of slot filling.}
 \label{fig:sg-slot-fp}
\end{subfigure}
\begin{subfigure}[b]{0.48\textwidth}
 \centering
 \includegraphics[width=1\textwidth]{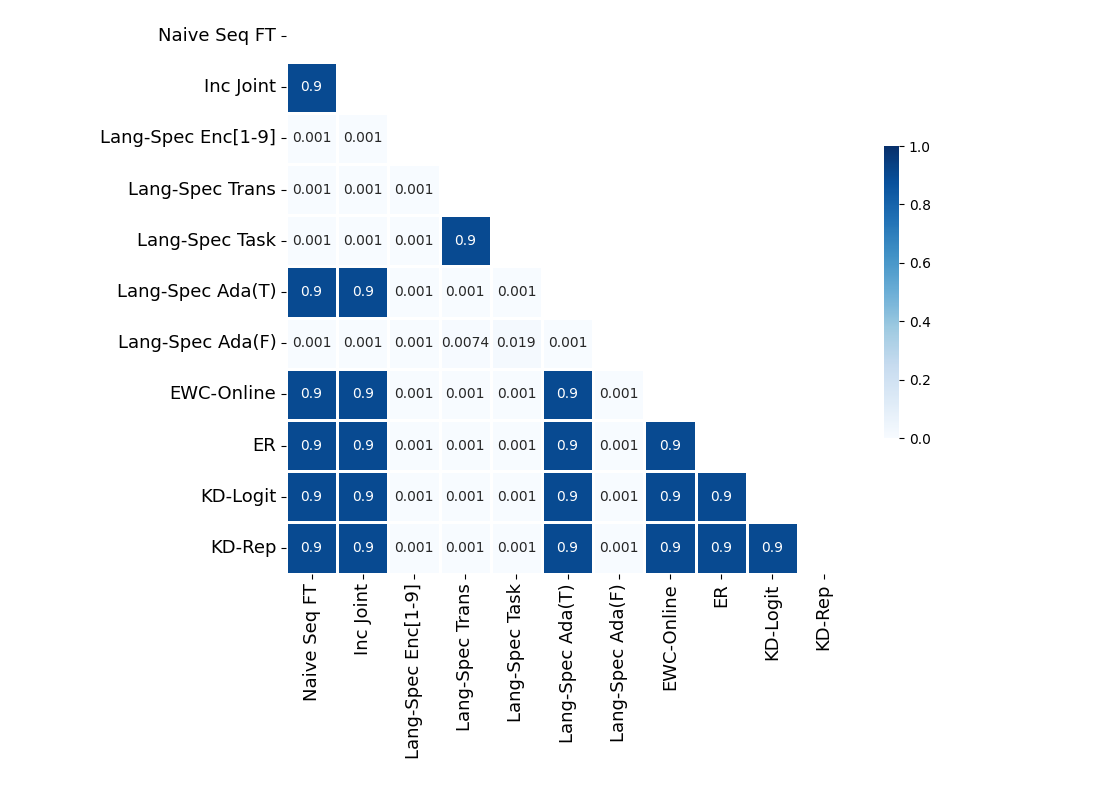}
 \caption{Zero-shot transfer of intent accuracy.}
 \label{fig:sg-intent-fwt0}
\end{subfigure}
\begin{subfigure}[b]{0.48\textwidth}
 \centering
 \includegraphics[width=1\textwidth]{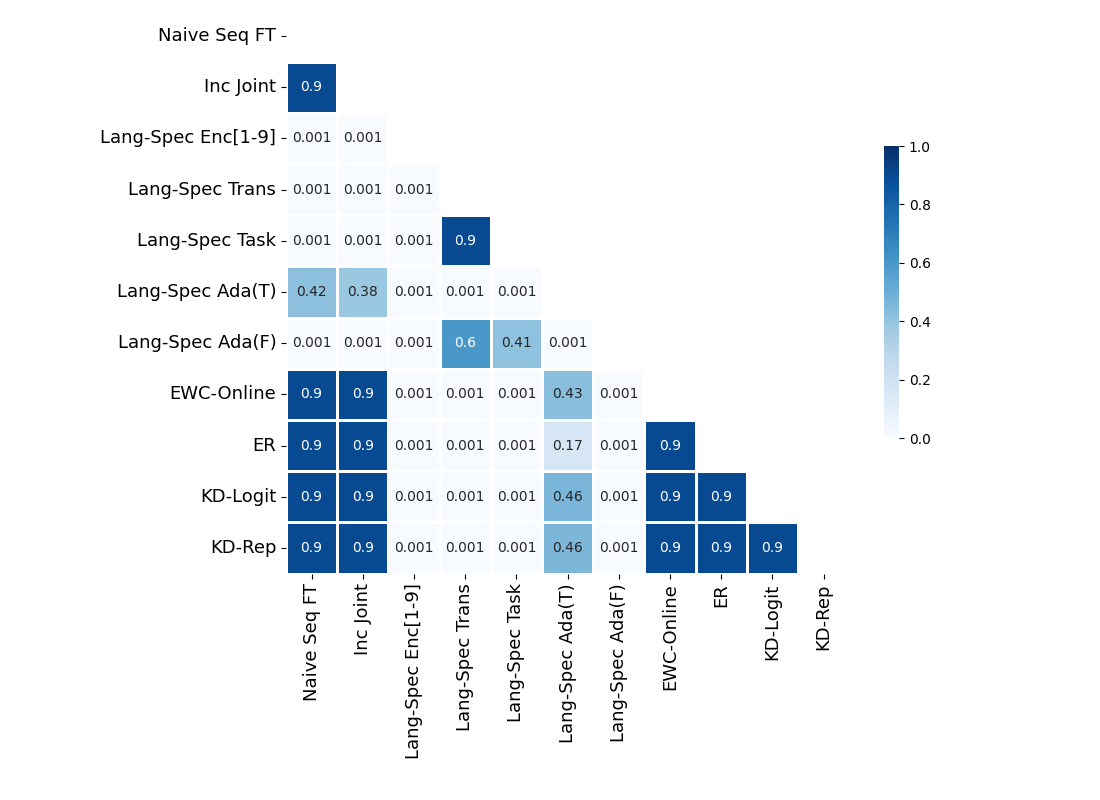}
 \caption{Zero-shot transfer of slot filling.}
 \label{fig:sg-slot-fwt0}
\end{subfigure}
\caption{\label{fig:hsd} P-values for different pairwise comparisons of different continual learning approaches using Tukey's honestly significant difference (HSD) test using bootstrap sampling.}
\end{figure*}

\begin{figure*}
\begin{subfigure}[b]{0.48\textwidth}
 \centering
 \includegraphics[width=1\textwidth]{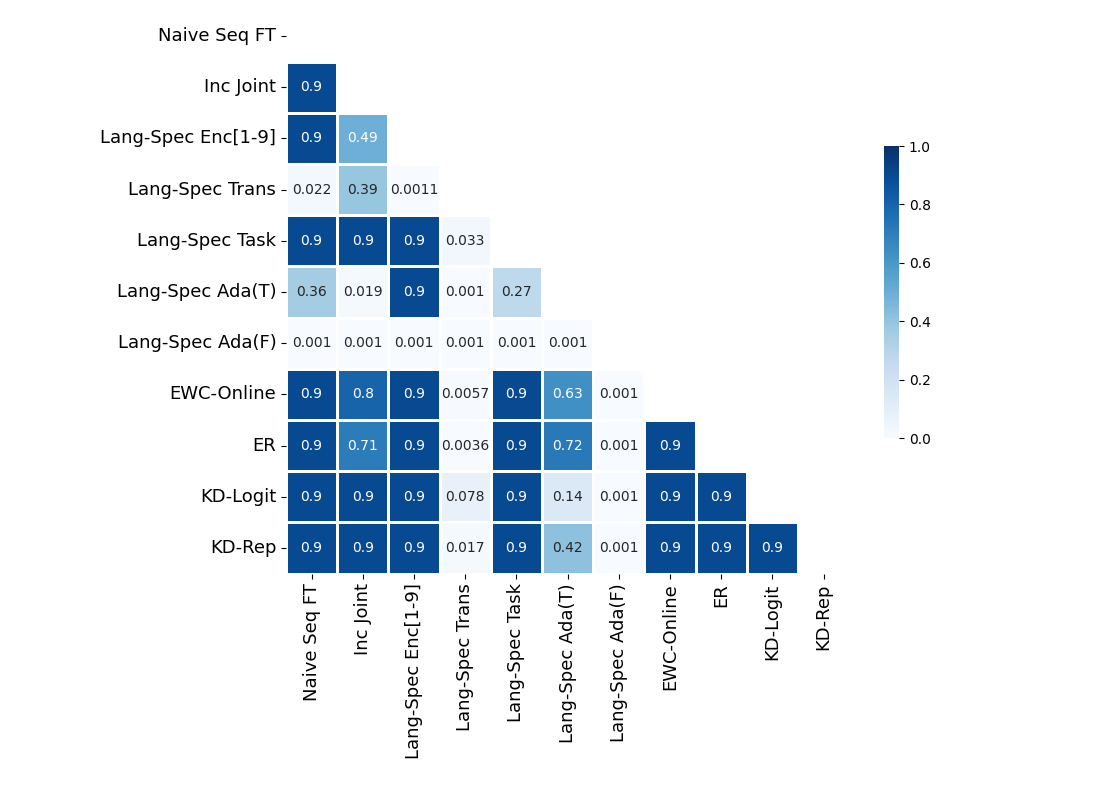}
 \caption{Transfer of intent accuracy.}
 \label{fig:sg-intent-fwt}
\end{subfigure}
\begin{subfigure}[b]{0.48\textwidth}
 \centering
 \includegraphics[width=1\textwidth]{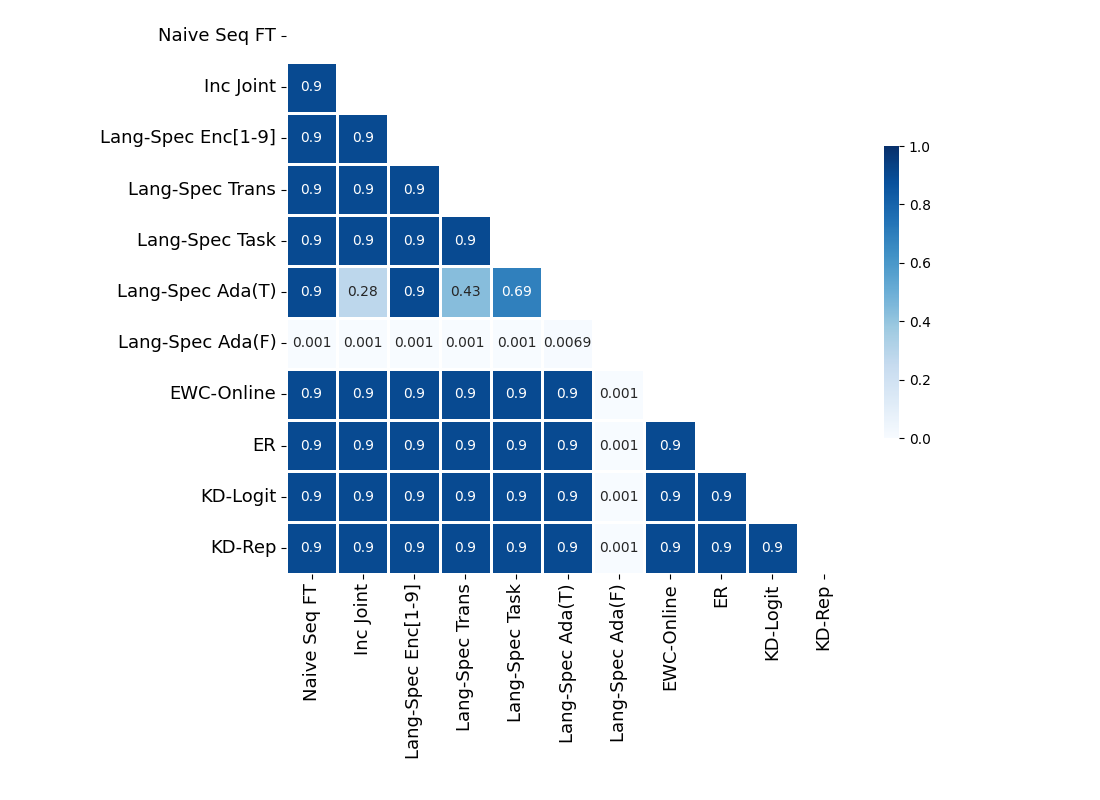}
 \caption{Transfer of slot filling.}
 \label{fig:sg-slot-fwt}
\end{subfigure}
\caption{\label{fig:hsdcont} P-values for different pairwise comparisons of different continual learning approaches using Tukey's honestly significant difference (HSD) test using bootstrap sampling (Cont.).}
\end{figure*}

\begin{figure*}[ht]
 \centering
 \begin{subfigure}[b]{0.48\textwidth}
 \centering
\includegraphics[width=1\textwidth]{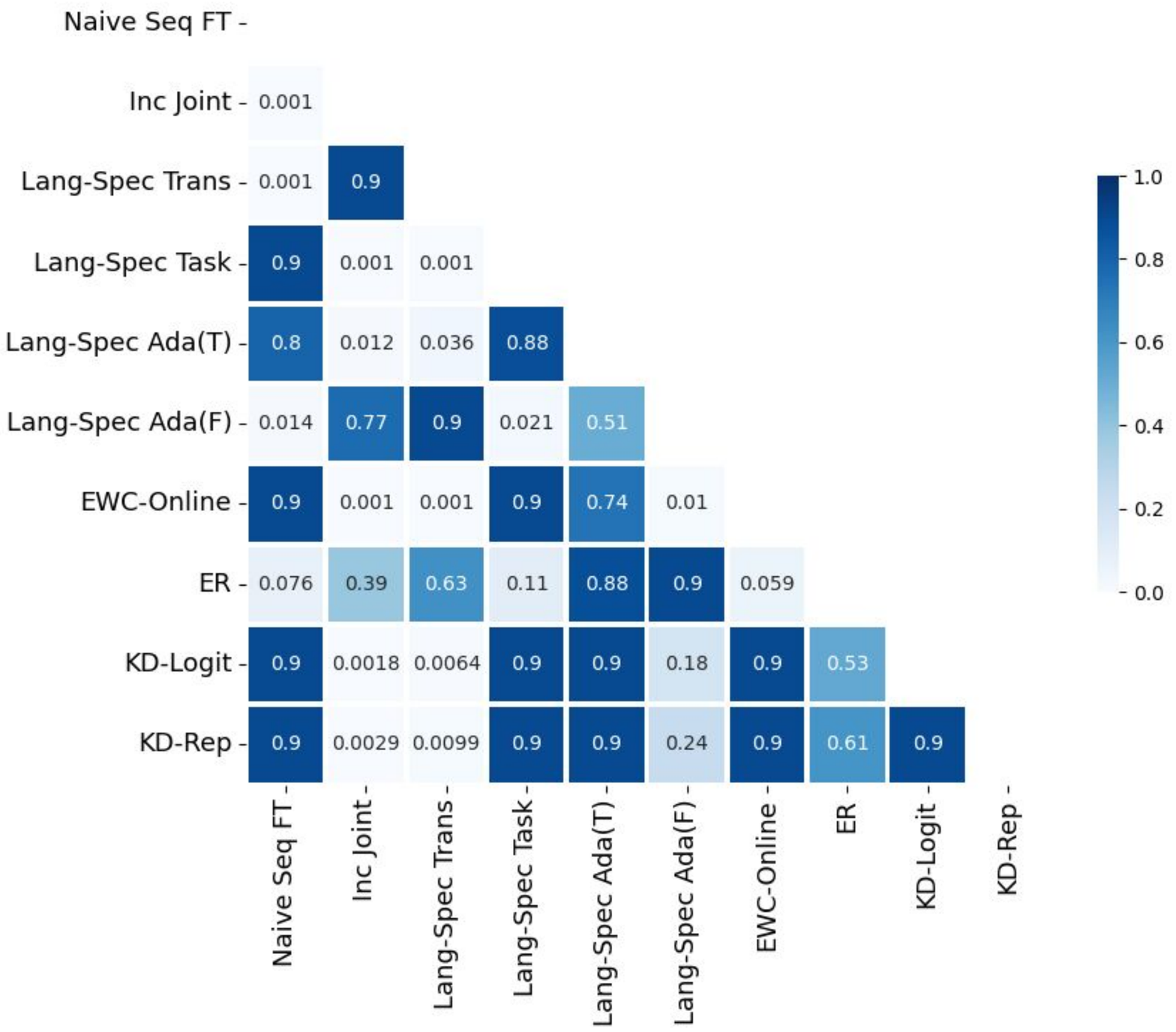}
 \caption{Forgetting of intent accuracy.}
 \label{fig:sg-intent-forget-seed}
\end{subfigure}
\begin{subfigure}[b]{0.48\textwidth}
 \centering
 \includegraphics[width=1\textwidth]{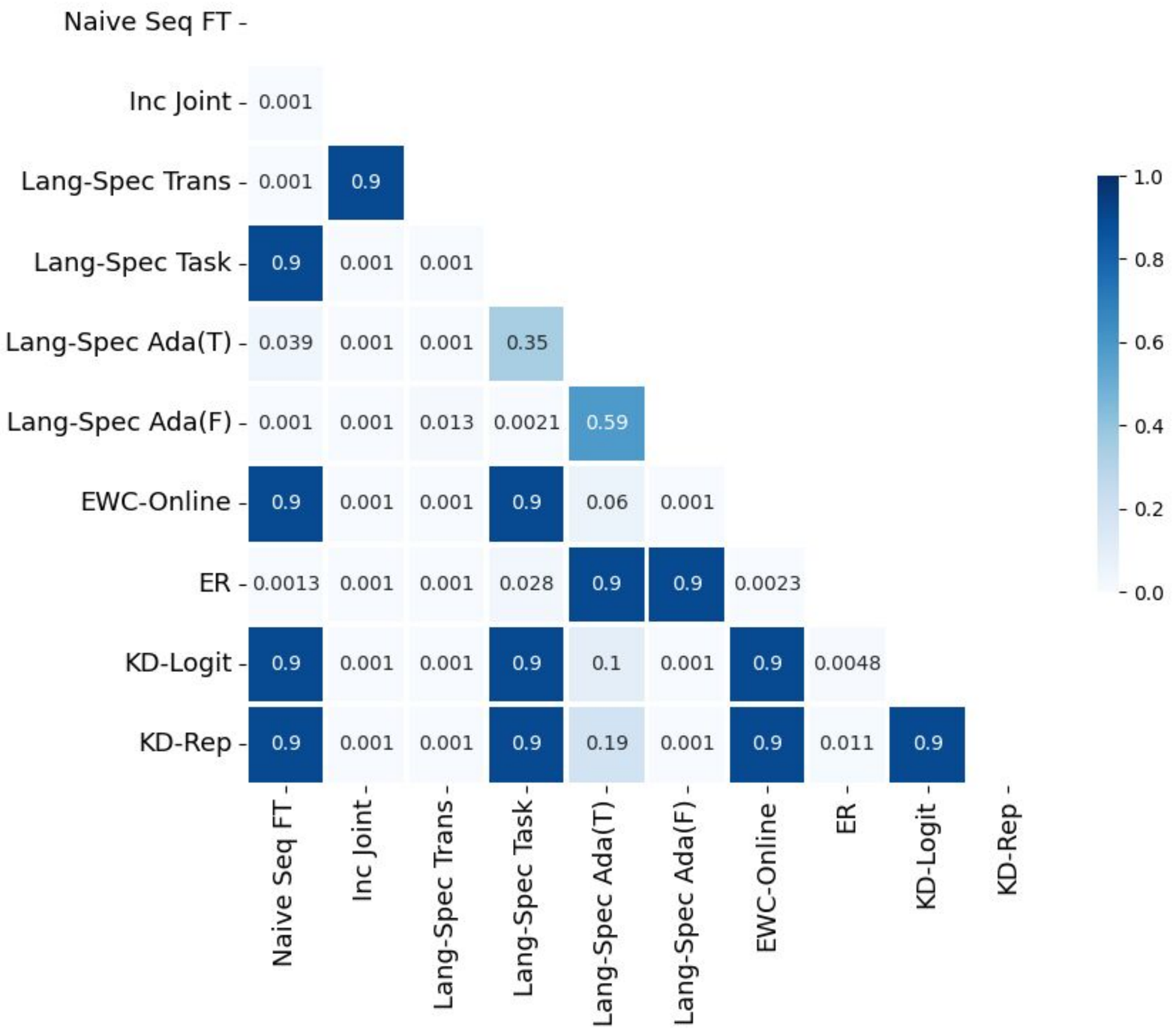}
 \caption{Forgetting of slot filling.}
 \label{fig:sg-slot-forget-seed}
\end{subfigure}
\begin{subfigure}[b]{0.48\textwidth}
 \centering
 \includegraphics[width=1\textwidth]{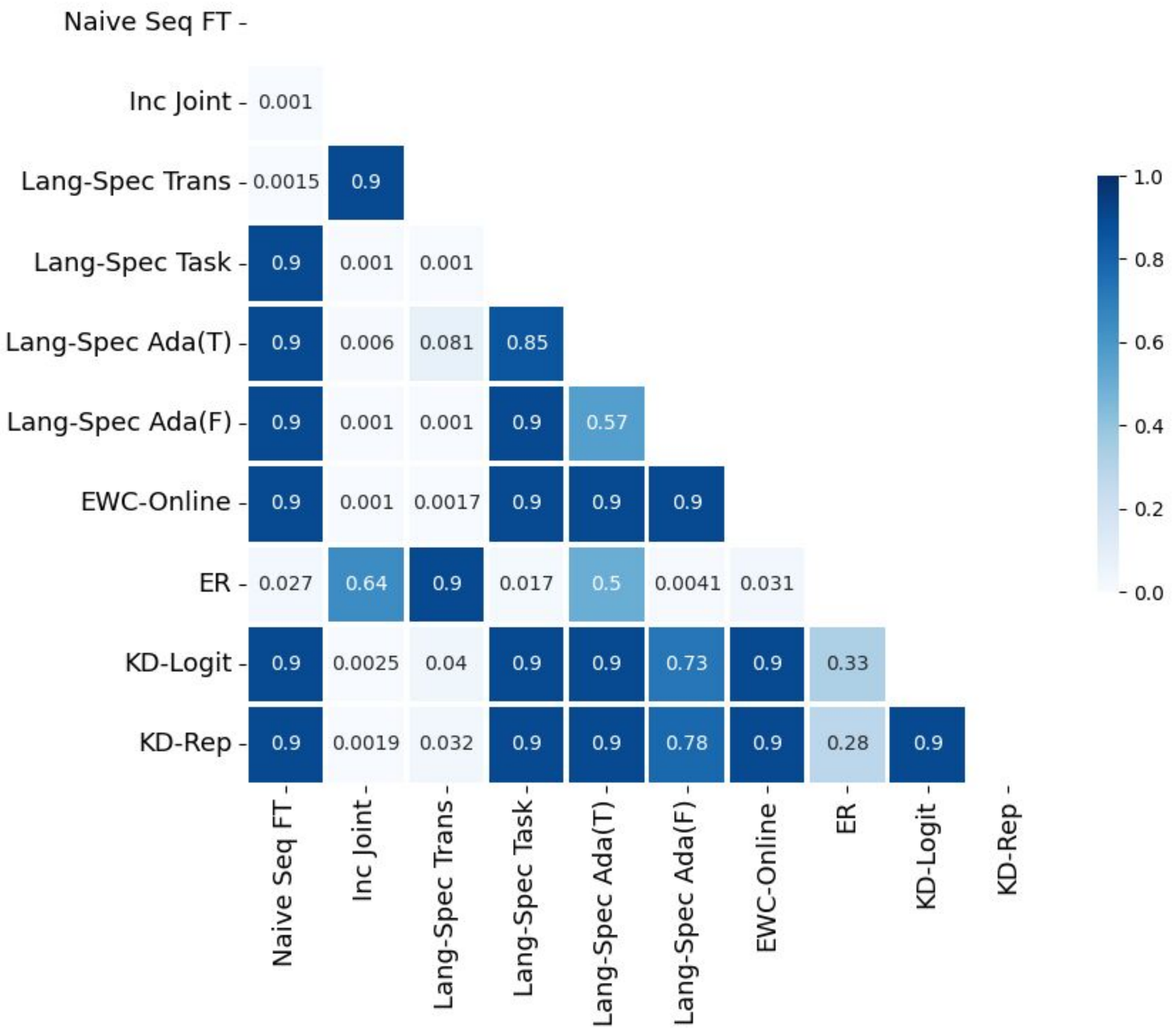}
 \caption{Final performance of intent accuracy.}
 \label{fig:sg-intent-fp-seed}
\end{subfigure}
\begin{subfigure}[b]{0.48\textwidth}
 \centering
 \includegraphics[width=1\textwidth]{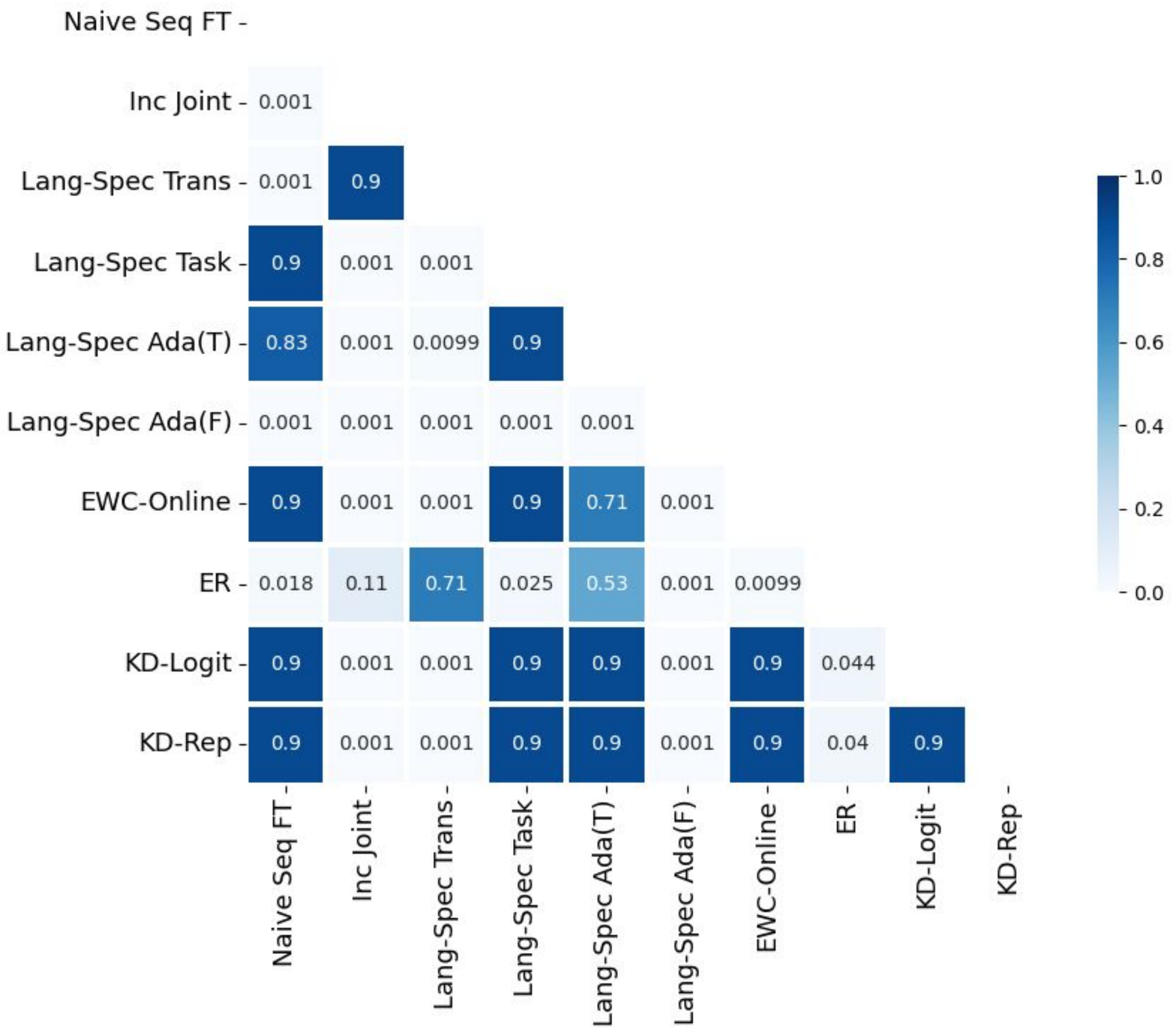}
 \caption{Final performance of slot filling.}
 \label{fig:sg-slot-fp-seed}
\end{subfigure}
\begin{subfigure}[b]{0.48\textwidth}
 \centering
 \includegraphics[width=1\textwidth]{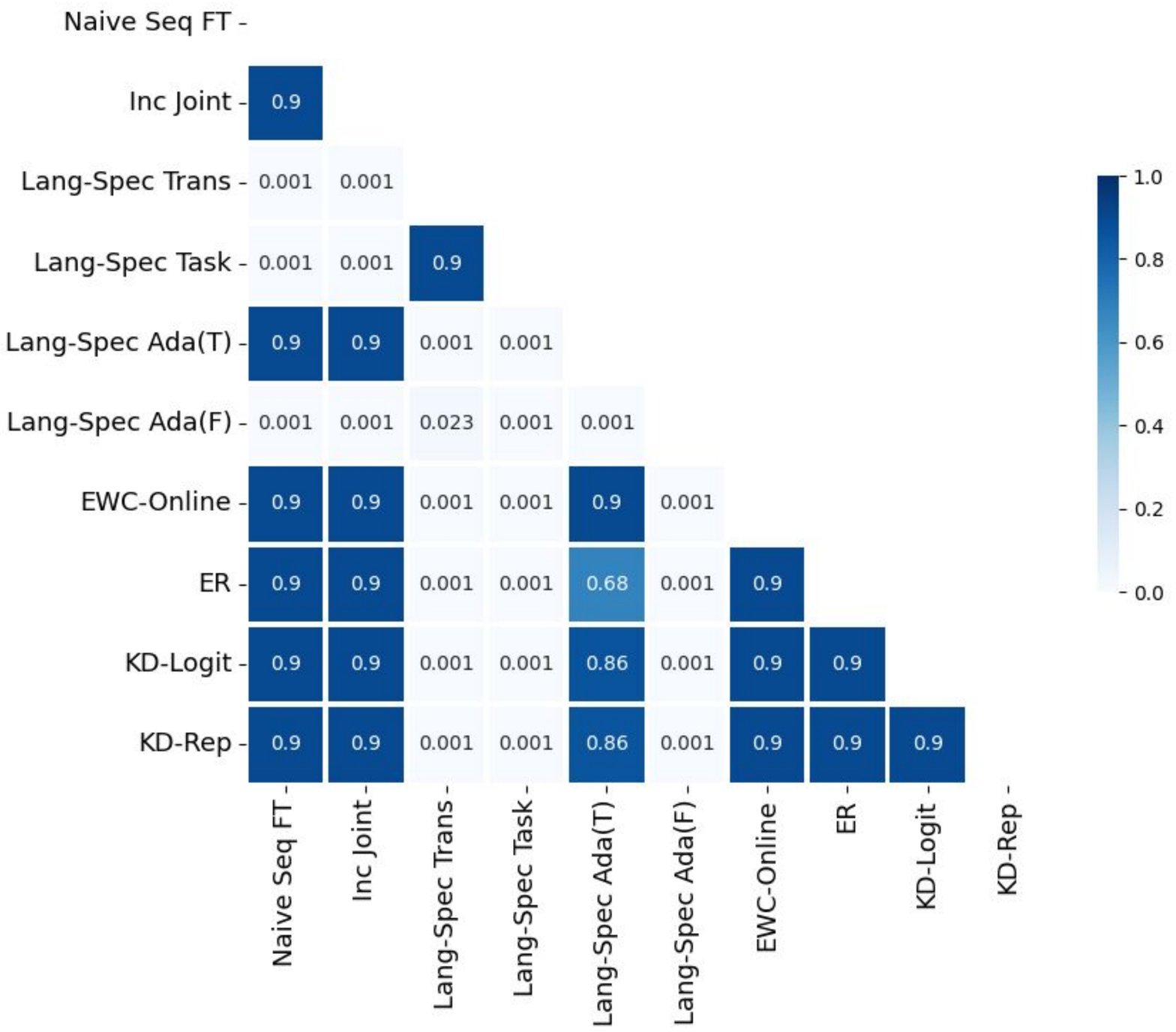}
 \caption{Zero-shot transfer of intent accuracy.}
 \label{fig:sg-intent-fwt0-seed}
\end{subfigure}
\begin{subfigure}[b]{0.48\textwidth}
 \centering
 \includegraphics[width=1\textwidth]{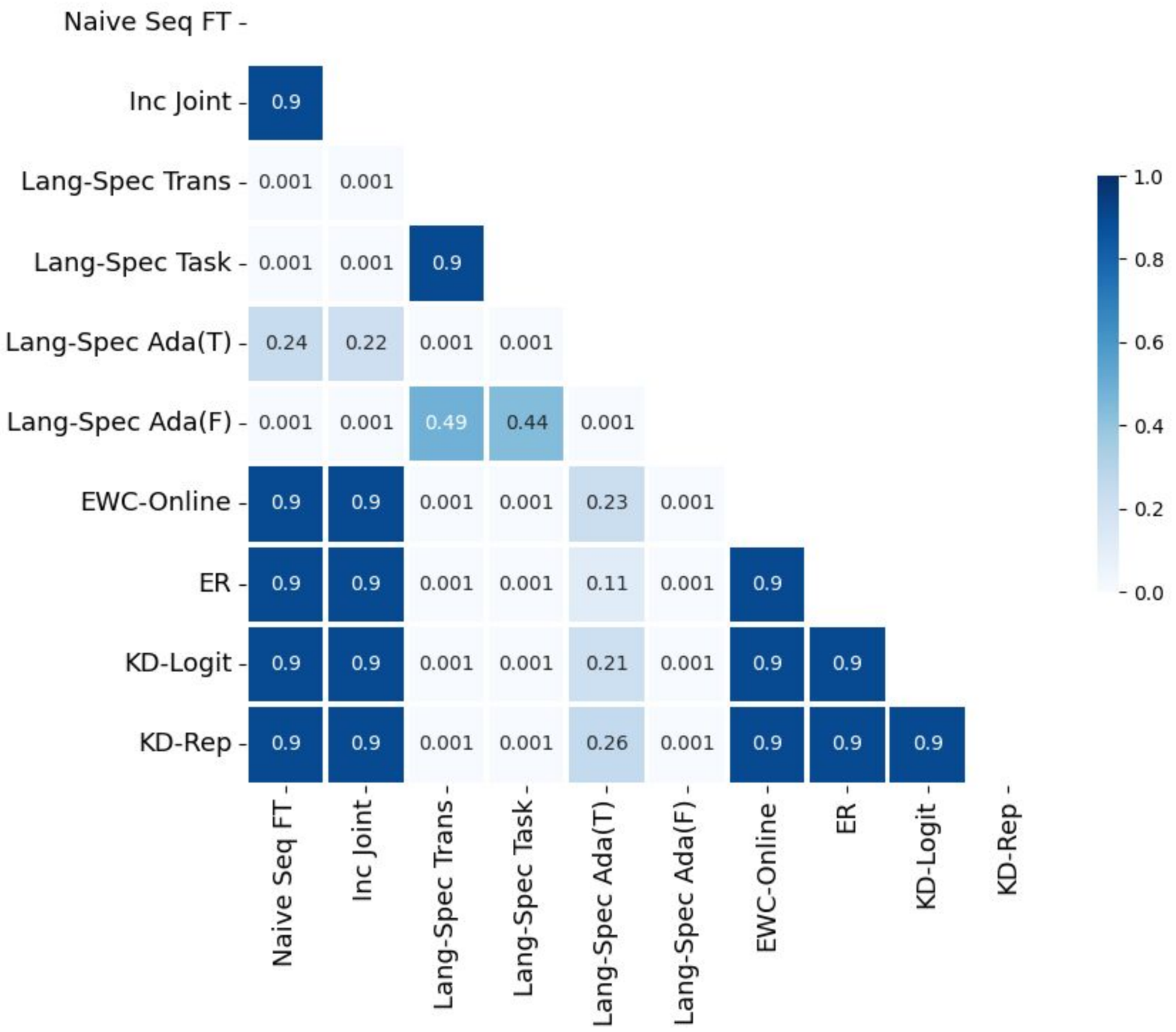}
 \caption{Zero-shot transfer of slot filling.}
 \label{fig:sg-slot-fwt0-seed}
\end{subfigure}
\caption{\label{fig:hsd-seeds} P-values for different pairwise comparisons of different continual learning approaches using Tukey's honestly significant difference (HSD) test using different seeds.}
\end{figure*}

\begin{figure*}
\begin{subfigure}[b]{0.48\textwidth}
 \centering
 \includegraphics[width=1\textwidth]{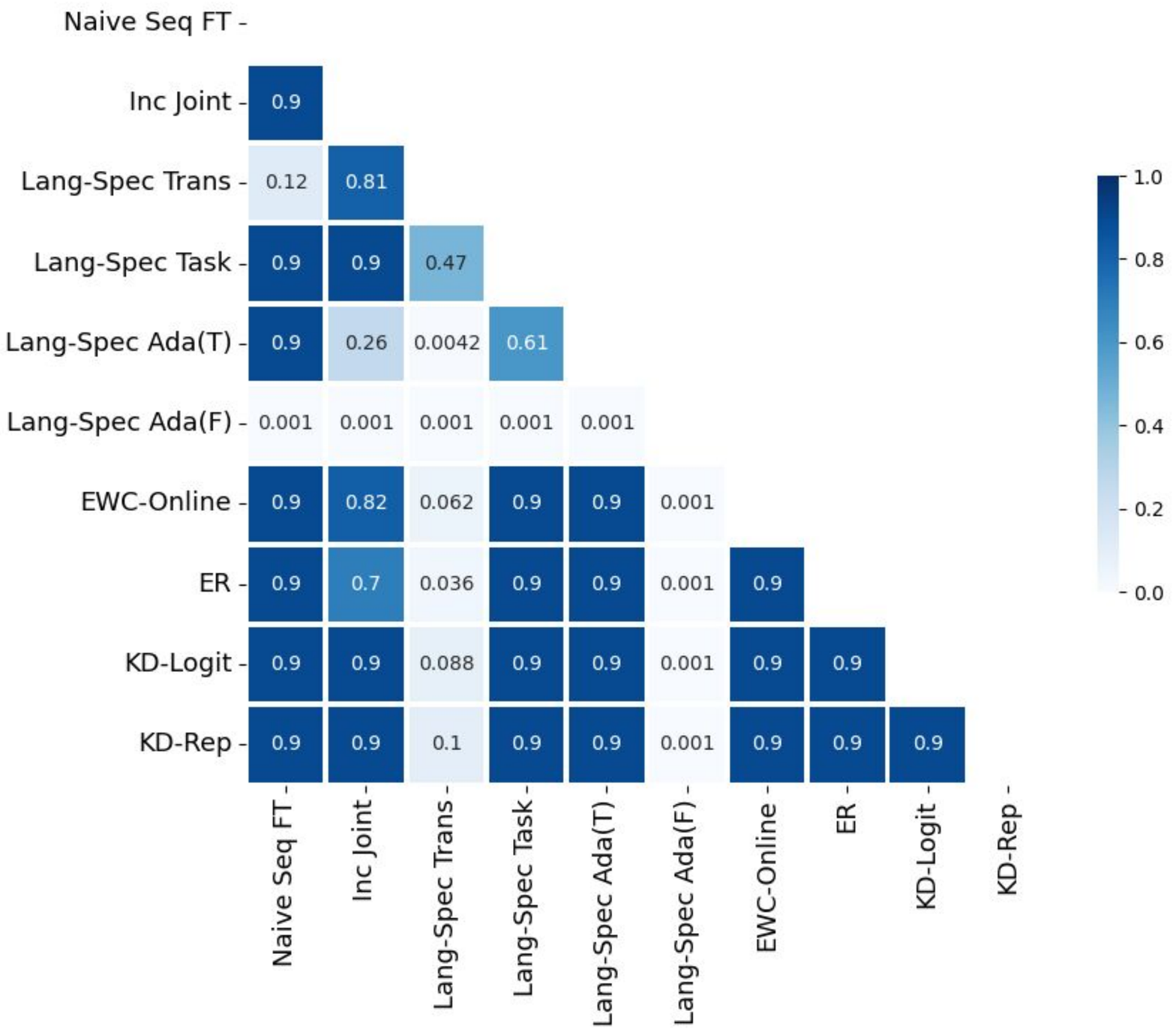}
 \caption{Transfer of intent accuracy.}
 \label{fig:sg-intent-fwt-seed}
\end{subfigure}
\begin{subfigure}[b]{0.48\textwidth}
 \centering
 \includegraphics[width=1\textwidth]{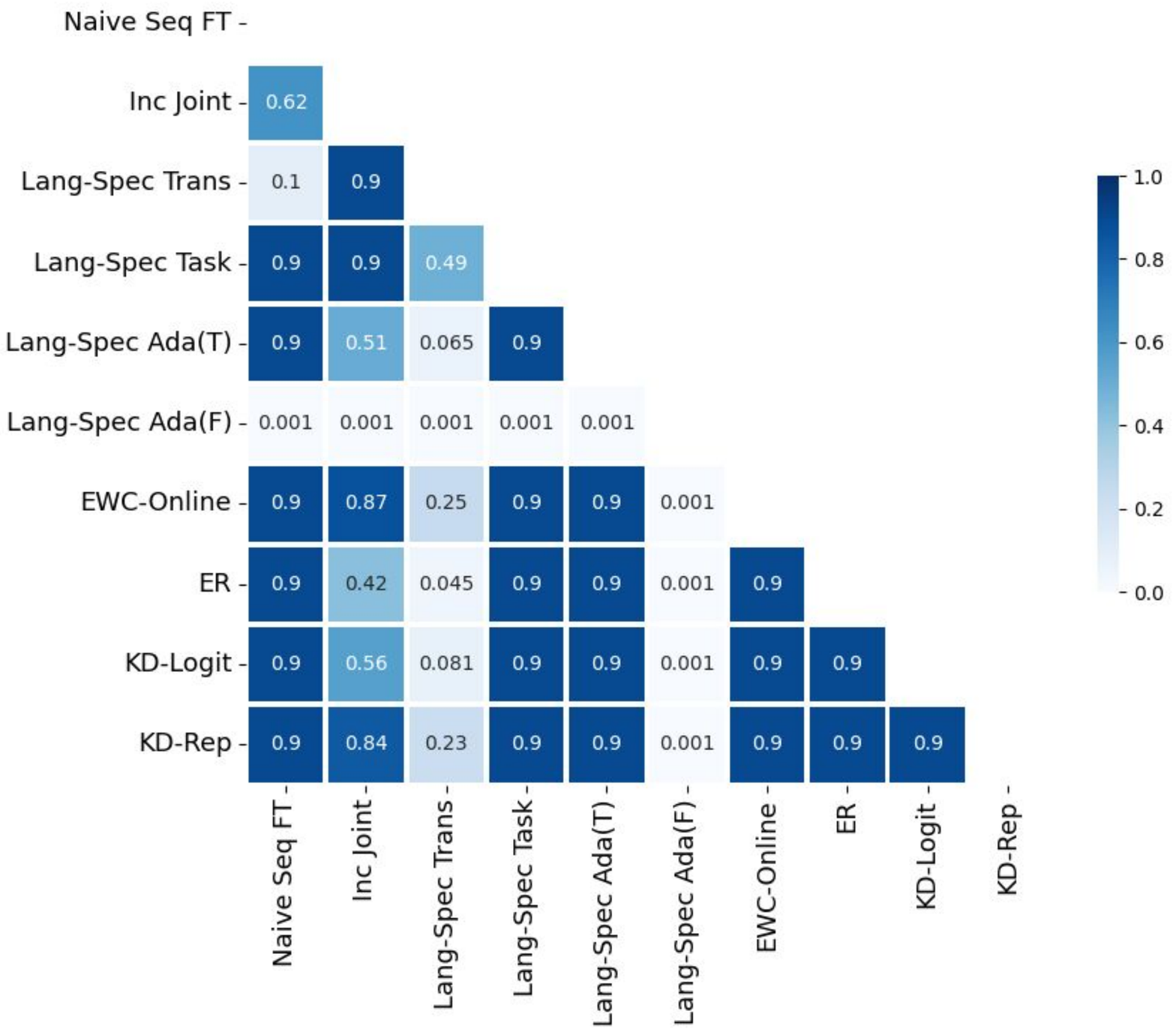}
 \caption{Transfer of slot filling.}
 \label{fig:sg-slot-fwt-seed}
\end{subfigure}
\caption{\label{fig:hsdcont-seeds} P-values for different pairwise comparisons of different continual learning approaches using Tukey's honestly significant difference (HSD) test using different seeds (Cont.).}
\end{figure*}

We show in Figures~\ref{fig:hsd} and~\ref{fig:hsdcont} the results for different approaches with a p-value lower than 0.05 for confidence intervals of 95\%, thus rejecting the null hypothesis that they are drawn from the same distribution. Figures~\ref{fig:sg-intent-forget},~\ref{fig:sg-intent-fwt},~\ref{fig:sg-intent-fp},~\ref{fig:sg-slot-forget},~\ref{fig:sg-intent-fwt},~\ref{fig:sg-slot-fp},~\ref{fig:sg-intent-fwt0}, and~\ref{fig:sg-slot-fwt0} show confusion plots of statistical significance p-values for different metrics (forgetting, transfer, and final performance) for intent classification and slot filling, respectively. For example, for forgetting, we notice that improvements or losses from approaches are statistically significant with $95\%$ confidence more than $49\%$ and $61\%$ of the time for intent classification and slot filling. For zero-shot transfer, we notice $60\%$ and $56\%$ of pairwise comparisons are statistically significant for intent classification and slot filling. For the final performance, we notice $47\%$ and $49\%$ of pairwise comparisons are statistically significant for intent classification and slot filling. For transfer, we notice that improvements or degradation over the transfer of intent classification are not statistically significant with the exceptions of \spectrans{}, which is the lowest in terms of transfer, and \adafrozen{}, which exhibits high transfer. The same can be said for \adafrozen{} in slot filling. Overall, model expansion approaches exhibit the highest statistical significance, whereas \ewconline{} and knowledge distillation are among the lowest.
Figures~\ref{fig:hsd-seeds} and~\ref{fig:hsdcont-seeds} show the corresponding statistical significance p-value confusion plots using multiple seeds. With a few exceptions like \langspecadatuned{} and \langspecadafrozen{}, most pairwise p-values that indicate statistical significance between two models using bootstrap sampling analysis are compliant with statistical significance computed using multiple seeds.  

\end{document}